%% file: neurips_data_2024.tex
\documentclass{article}

\PassOptionsToPackage{numbers, compress}{natbib}
\usepackage[preprint]{neurips_data_2024}

\usepackage[utf8]{inputenc} 
\usepackage[T1]{fontenc}    
\usepackage{hyperref}       
\usepackage{url}            
\usepackage{booktabs}       
\usepackage{amsfonts}       
\usepackage{nicefrac}       
\usepackage{microtype}      
\usepackage[dvipsnames]{xcolor}         

\usepackage{amsmath}
\usepackage{amssymb}
\usepackage{mathtools}
\usepackage{amsthm}

\usepackage[capitalize,noabbrev]{cleveref}

\theoremstyle{plain}

\theoremstyle{definition}

\theoremstyle{remark}

\usepackage{longtable}
\usepackage{multirow}
\usepackage{wrapfig}
\usepackage{adjustbox}
\usepackage{placeins}
\usepackage{listings}
\usepackage[labelfont=it]{caption}
\usepackage{graphicx}
\usepackage{geometry}
\usepackage{subfigure}
\usepackage{pifont}

\newcommand{\xmark}{\ding{55}}

\definecolor{codegreen}{rgb}{0,0.6,0}
\definecolor{codegray}{rgb}{0.5,0.5,0.5}
\definecolor{codepurple}{rgb}{0.58,0,0.82}
\definecolor{backcolour}{rgb}{0.95,0.95,0.92}

\lstdefinestyle{mystyle}{
    language=Python,
    backgroundcolor=\color{backcolour},   
    commentstyle=\color{codegreen},
    keywordstyle=\color{magenta},
    numberstyle=\tiny\color{codegray},
    stringstyle=\color{codepurple},
    basicstyle=\ttfamily\footnotesize,
    breakatwhitespace=false,         
    breaklines=true,                 
    captionpos=b,                    
    keepspaces=true,                 
    numbers=left,                    
    numbersep=5pt,                  
    showspaces=false,                
    showstringspaces=false,
    showtabs=false,                  
    tabsize=2,
    basewidth=0.58em,
}

\usepackage{tikz}
\usetikzlibrary{positioning,calc,fit}
\definecolor{myblue}{RGB}{124,156,205}
\definecolor{mypurple}{RGB}{208,167,203}
\definecolor{mypink}{RGB}{233,198,235}

\newlength\myframesep
\newlength\horizframesep
\pgfdeclarelayer{proglanguages}
\pgfdeclarelayer{rlp}
\pgfdeclarelayer{puzzle}
\pgfsetlayers{proglanguages,rlp,puzzle,main}

\pgfkeys{
    /tikz/node distance/.append code={
        \pgfkeyssetvalue{/tikz/node distance value}{#1}
    }
}

\crefname{lstlisting}{Listing}{Listings}

\def\rlb{\texttt{PUZZLES}}

\title{PUZZLES: A Benchmark for Neural\\Algorithmic Reasoning}

\author{%
Benjamin Estermann, Luca A. Lanzendörfer, Yannick Niedermayr, Roger Wattenhofer\\
ETH Zürich\\
\texttt{\{estermann, lanzendoerfer, yannickn, wattenhofer\}@ethz.ch}
}

\begin{document}

\maketitle

\begin{abstract}
    Algorithmic reasoning is a fundamental cognitive ability that plays a pivotal role in problem-solving and decision-making processes.
    Reinforcement Learning (RL) has demonstrated remarkable proficiency in tasks such as motor control, handling perceptual input, and managing stochastic environments.
    These advancements have been enabled in part by the availability of benchmarks.
    In this work we introduce \rlb{}, a benchmark based on Simon Tatham's Portable Puzzle Collection, aimed at fostering progress in algorithmic and logical reasoning in RL.
    \rlb{} contains 40 diverse logic puzzles of adjustable sizes and varying levels of complexity; many puzzles also feature a diverse set of additional configuration parameters.
    The 40 puzzles
    provide detailed information on the strengths and generalization capabilities of RL agents.
    Furthermore, we evaluate various RL algorithms on \rlb{}, providing baseline comparisons and demonstrating the potential for future research.
    All the software, including the environment, is available at \url{https://github.com/ETH-DISCO/rlp}.
\end{abstract}

    Human intelligence relies heavily on logical and algorithmic reasoning as integral components for solving complex tasks.
    While Machine Learning (ML) has achieved remarkable success in addressing many real-world challenges, logical and algorithmic reasoning remains an open research question \cite{serafini2016logic,dai2019bridging,li2020strong,velivckovic2021neural,masry2022chartqa,jiao2022merit,bardin2023machine}. 
    This research question is supported by the availability of benchmarks, which allow for a standardized and broad evaluation framework to measure and encourage progress~\citep{li2021isarstep, velikovic2022algorithmicreasoning,srivastava2022beyond}. 

    Reinforcement Learning (RL) has made remarkable progress in various domains, showcasing its capabilities in tasks such as game playing \citep{mnih2013atari,tang2017exploration,silver2018general,badia2020agent57,wurman2022outracing} , robotics \citep{kalashnikov2018scalable, kiran2021deep,rudin2022learning, rana2023residual} and control systems \citep{wang2020reinforcement,wu2022deep, brunke2022safe}.
    Various benchmarks have been proposed to enable progress in these areas~\citep{todorov2012mujoco,bellemare2013arcade, brockman2016openai,duan2016benchmarking,tassa2018deepmind, cote2018textworld, Lanctot2019OpenSpiel}.
    More recently, advances have also been made in the direction of logical and algorithmic reasoning within RL~\citep{jiang2019neural,fawzi2022discovering,mankowitz2023faster}. 
    Popular examples also include the games of Chess, Shogi, and Go~\citep{lai2015giraffe, silver2016alphago}.
    Given the importance of logical and algorithmic reasoning, we propose a benchmark to guide future developments in RL and more broadly machine learning.

    Logic puzzles have long been a playful challenge for humans, and they are an ideal testing ground for evaluating the algorithmic and logical reasoning capabilities of RL agents.
    A diverse range of puzzles, similar to the Atari benchmark~\cite{bellemare2013arcade}, favors methods that are broadly applicable.
    Unlike tasks with a fixed input size, logic puzzles can be solved iteratively once an algorithmic solution is found. This allows us to measure how well a solution attempt can adapt and generalize to larger inputs. 
    Furthermore, in contrast to games such as Chess and Go, logic puzzles have a known solution, making reward design easier and enabling tracking progress and guidance with intermediate rewards.

    \begin{figure*}[t]
      \centering
        \includegraphics[width=\linewidth]{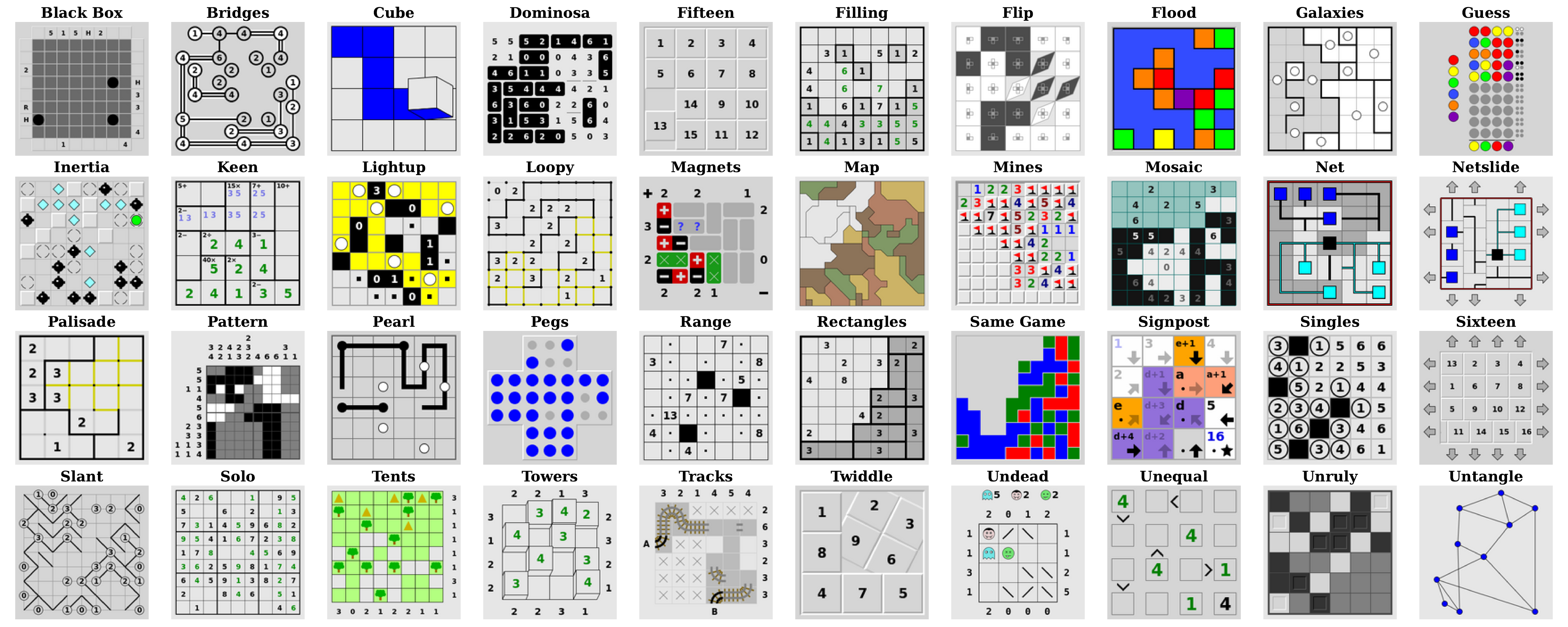}
        \caption{All puzzle classes of Simon Tatham's Portable Puzzle Collection.
        }
        \label{fig:puzzle-collection}
    \end{figure*}

    In this paper, we introduce \rlb{}, a comprehensive RL benchmark specifically designed to evaluate RL agents' algorithmic reasoning and problem-solving abilities in the realm of logical and algorithmic reasoning.
    Simon Tatham's Puzzle Collection~\citep{site:sgt-puzzles}, curated by the renowned computer programmer and puzzle enthusiast Simon Tatham, serves as the foundation of \rlb{}.
    This collection includes a set of 40 logic puzzles, shown in \cref{fig:puzzle-collection}, each of which presents distinct challenges with various dimensions of adjustable complexity.
    They range from more well-known puzzles, such as \textit{Solo} or \textit{Mines} (commonly known as \textit{Sudoku} and \textit{Minesweeper}, respectively) to lesser-known puzzles such as \textit{Cube} or \textit{Slant}.
    \rlb{} includes all 40 puzzles in a standardized environment, each playable with a visual or discrete input and a discrete action space.

    \paragraph{Contributions.} 

    We propose \rlb{}, an RL environment based on Simon Tatham's Puzzle Collection, comprising a collection of 40 diverse logic puzzles.
    To ensure compatibility, we have extended the original C source code to adhere to the standards of the Pygame library.
    Subsequently, we have integrated \rlb{} into the Gymnasium framework API, providing a straightforward, standardized, and widely-used interface for RL applications.
    \rlb{} allows the user to arbitrarily scale the size and difficulty of logic puzzles, providing detailed information on the strengths and generalization capabilities of RL agents.
    Furthermore, we have evaluated various RL algorithms on \rlb{}, providing baseline comparisons and demonstrating the potential for future research.

\section{Related Work}
    \paragraph{RL benchmarks.}
    Various benchmarks have been proposed in RL.
    \citet{bellemare2013arcade} introduced the influential Atari-2600 benchmark, on which \citet{mnih2013atari} trained RL agents to play the games directly from pixel inputs.
    This benchmark demonstrated the potential of RL in complex, high-dimensional environments.
    \rlb{} allows the use of a similar approach where only pixel inputs are provided to the agent.
    \citet{todorov2012mujoco} presented MuJoCo which provides a diverse set of continuous control tasks based on a physics engine for robotic systems.
    Another control benchmark is the DeepMind Control Suite by \citet{duan2016benchmarking}, featuring continuous actions spaces and complex control problems.
    The work by \citet{cote2018textworld} emphasized the importance of natural language understanding in RL and proposed a benchmark for evaluating RL methods in text-based domains.
    \citet{Lanctot2019OpenSpiel} introduced OpenSpiel, encompassing a wide range of games, enabling researchers to evaluate and compare RL algorithms' performance in game-playing scenarios.
    These benchmarks and frameworks have contributed significantly to the development and evaluation of RL algorithms.
    OpenAI Gym by \citet{brockman2016openai}, and its successor Gymnasium by the Farama Foundation~\citep{site:farama-gymnasium} helped by providing a standardized interface for many benchmarks.
    As such, Gym and Gymnasium have played an important role in facilitating reproducibility and benchmarking in reinforcement learning research.
    Therefore, we provide \rlb{} as a Gymnasium environment to enable ease of use.

    \paragraph{Logical and algorithmic reasoning within RL.}
    Notable research in RL on logical reasoning includes automated theorem proving using deep RL~\citep{kalashnikov2018scalable} or RL-based logic synthesis~\citep{wang2022rethinking}.
    \citet{dasgupta2019causal} find that RL agents can perform a certain degree of causal reasoning in a meta-reinforcement learning setting.
    The work by \citet{jiang2019neural} introduces Neural Logic RL, which improves interpretability and generalization of learned policies.
    \citet{eppe2022intelligent} provide steps to advance problem-solving as part of hierarchical RL.
    \citet{fawzi2022discovering} and \citet{mankowitz2023faster} demonstrate that RL can be used to discover novel and more efficient algorithms for well-known problems such as matrix multiplication and sorting.
    Neural algorithmic reasoning has also been used as a method to improve low-data performance in classical RL control environments~\citep{deac2021neural,he2022continuous}.
    Logical reasoning might be required to compete in certain types of games such as chess, shogi and Go~\citep{lai2015giraffe,silver2016alphago,silver2017mastering, silver2018general}, Poker \citep{dahl2001reinforcement,heinrich2016deep,steinberger2019pokerrl, zhao2022alphaholdem} or board games \citep{ghory2004reinforcement, szita2012reinforcement, xenou2019deep, perolat2022mastering}.
    However, these are usually multi-agent games, with some also featuring imperfect information and stochasticity.

    \paragraph{Reasoning benchmarks.}
    Various benchmarks have been introduced to assess different types of reasoning capabilities, although only in the realm of classical ML.
    IsarStep, proposed by \citet{li2021isarstep}, specifically designed to evaluate high-level mathematical reasoning necessary for proof-writing tasks.
    Another significant benchmark in the field of reasoning is the CLRS Algorithmic Reasoning Benchmark, introduced by \citet{velikovic2022algorithmicreasoning}. 
    This benchmark emphasizes the importance of algorithmic reasoning in machine learning research.
    It consists of 30 different types of algorithms sourced from the renowned textbook ``Introduction to Algorithms'' by~\citet{cormen2022clrs}. 
    The CLRS benchmark serves as a means to evaluate models' understanding and proficiency in learning various algorithms.
    In the domain of large language models (LLMs), BIG-bench has been introduced by \citet{srivastava2022beyond}.
    BIG-bench incorporates tasks that assess the reasoning capabilities of LLMs, including logical reasoning.
    
    Despite these valuable contributions, a suitable and unified benchmark for evaluating logical and algorithmic reasoning abilities in single-agent perfect-information RL has yet to be established.
    Recognizing this gap, we propose \rlb{} as a relevant and necessary benchmark with the potential to drive advancements and provide a standardized evaluation platform for RL methods that enable agents to acquire algorithmic and logical reasoning abilities.

\section{The \rlb{} Environment}

    In the following section we give an overview of the \rlb{} environment.\footnotemark
    The environment is written in both Python and C.
    For a detailed explanation of all features of the environment as well as their implementation, please see \cref{app:env_features,app:environment-implementation}.
    
    \footnotetext{ The puzzles are available to play online at \url{https://www.chiark.greenend.org.uk/~sgtatham/puzzles/}; excellent standalone apps for Android and iOS exist as well.}

        \begin{figure*}[t]
        \centering
        \begin{tikzpicture}[node distance=3pt,outer sep=0pt,
        blueb/.style={
          draw=black,
          fill=myblue,
          rounded corners,
          text width=2.39cm,
          font={\sffamily\bfseries\color{black}},
          align=center,
          text height=12pt,
          text depth=16pt},
        purpleb/.style={blueb,fill=magenta},
        pinkb/.style={blueb,fill=mypink},
        greenb/.style={blueb,fill=Green!60},
        ]
        \node[greenb, fill=Maroon!40] (A) {Gymnasium RL Code};
        \node[purpleb,right=of A, fill=Green!25] (B) {\small\verb+puzzle_env.py+};
        \node[purpleb,right=of B, fill=Green!20] (C) {\small\verb+puzzle.py+};
        \node[purpleb,right=of C, fill=Green!20] (D) {\small\verb+pygame.c+};
        \node[greenb,right=of D, fill=Orange!40] (E) {Puzzle C Sources};
        \node[greenb, below=of C, yshift=-2.5\myframesep, fill=WildStrawberry!30] (J) {Pygame Library};
        \draw[line width=5pt] (A)  -- (B);
        \draw[line width=5pt] (B)  -- (C);
        \draw[line width=5pt] (C)  -- (D);
        \draw[line width=5pt] (D)  -- (E);
        \begin{pgfonlayer}{puzzle}
        \draw[blueb,draw=black,fill=Green!30,opacity=.7] 
          ([xshift=-\horizframesep,yshift=3\myframesep]C.north west) 
            rectangle 
          ([xshift=\horizframesep,yshift=-\myframesep]D.south east);
        \draw[line width=5pt] (B.center)  -- (J.center);
        \draw[line width=5pt] (C.center)  -- (J.center);
        \draw[line width=5pt] (D.center)  -- (J.center);
        \end{pgfonlayer}
        \begin{pgfonlayer}{rlp}
        \draw[blueb,draw=black,fill=Green!40,opacity=.7] 
          ([xshift=-\horizframesep,yshift=6\myframesep]B.north west) 
            rectangle 
          ([xshift=\horizframesep+1pt,yshift=-2\myframesep]D.south east);
        \end{pgfonlayer}
        \begin{pgfonlayer}{proglanguages}
        \draw[blueb,draw=black,fill=myblue!40] 
          ([xshift=-\horizframesep,yshift=9\myframesep]A.north west) 
            rectangle 
          ([xshift=\horizframesep,yshift=-\myframesep]J.south east);
        \draw[blueb,draw=black,fill=myblue!40] 
          ([xshift=\horizframesep,yshift=9\myframesep]E.north east) 
            rectangle 
          ([xshift=\horizframesep+1pt,yshift=-\myframesep]J.south east);
        \end{pgfonlayer}
        \node[font=\sffamily\itshape\color{black},above=of C, yshift=2pt] (F) {\verb+puzzle+ Module};
        \node[font=\sffamily\itshape\color{black},above=of B, xshift=-2pt, yshift=3\myframesep] (G) {\verb+rlp+ Package};
        \node[font=\sffamily\itshape\color{black},above=of A, xshift=-14pt, yshift=6\myframesep] (H) {Python};
        \node[font=\sffamily\itshape\color{black},above=of D, xshift=-26pt, yshift=6\myframesep+2pt] (I) {C};
        \end{tikzpicture}
        
        \caption{Code and library landscape around the \color{ForestGreen} \rlb{} Environment\color{Black}, made up of the \color{ForestGreen} \texttt{rlp} Package \color{Black} and the \color{ForestGreen} \texttt{puzzle} Module \color{Black}. The figure shows how the \color{ForestGreen}\texttt{puzzle} Module \color{black}presented in this paper fits within \color{BurntOrange}Tathams's Puzzle Collection\protect\footnotemark \color{black}  code, the \color{OrangeRed}\texttt{Pygame} package\color{black}, and a user's \color{Maroon}\texttt{Gymnasium} reinforcement learning code\color{black} . The different parts are also categorized as \color{blue}Python language and C language\color{black}.}
        \label{fig:environment-landscape}
    \end{figure*}
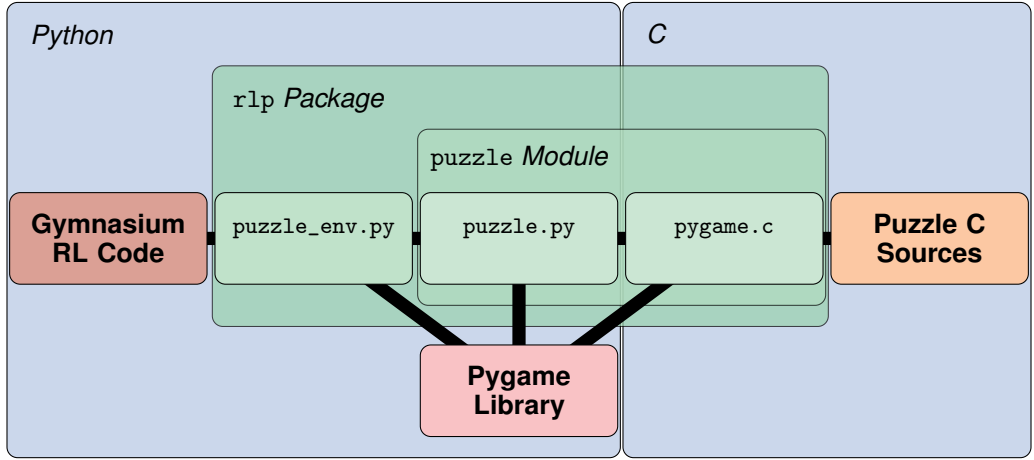
    
\subsection{Environment Overview}    
    Within the \rlb{} environment, we encapsulate the tasks presented by each logic puzzle by defining consistent state, action, and observation spaces.
    It is also important to note that the large majority of the logic puzzles are designed so that they can be solved without requiring any guesswork.
    By default, we provide the option of two observation spaces, one is a representation of the discrete internal game state of the puzzle, the other is a visual representation of the game interface.
    These observation spaces can easily be wrapped in order to enable \rlb{} to be used with more advanced neural architectures such as graph neural networks (GNNs) or Transformers.
    All puzzles provide a discrete action space which only differs in cardinality.
    To accommodate the inherent difficulty and the need for proper algorithmic reasoning in solving these puzzles, the environment allows users to implement their own reward structures, facilitating the training of successful RL agents.
    All puzzles are played in a two-dimensional play area with deterministic state transitions, where a transition only occurs after a valid user input.
    Most of the puzzles in PUZZLES do not have an upper bound on the number of steps, they can only be completed by successfully solving the puzzle.
    An agent with a bad policy is likely never going to reach a terminal state.
    For this reason, we provide the option for early episode termination based on state repetitions.
    As we show in \cref{experiments:ep_length}, this is an effective method to facilitate learning.

\subsection{Difficulty Progression and Generalization}
    The \rlb{} environment places a strong emphasis on giving users control over the difficulty exhibited by the environment.
    For each puzzle, the problem size and difficulty can be adjusted individually.
    The difficulty affects the complexity of strategies that an agent needs to learn to solve a puzzle.
    As an example, \textit{Sudoku} has tangible difficulty options: harder difficulties may require the use of new strategies such as \textit{forcing chains}\footnote{\textit{Forcing chains} works by following linked cells to evaluate possible candidates, usually starting with a two-candidate cell.} to find a solution, whereas easy difficulties only need the \textit{single position} strategy.\footnote{The \textit{single position} strategy involves identifying cells which have only a single possible value.}

    The scalability of the puzzles in our environment offers a unique opportunity to design increasingly complex puzzle configurations, presenting a challenging landscape for RL agents to navigate.
    This dynamic nature of the benchmark serves two important purposes.
    Firstly, the scalability of the puzzles facilitates the evaluation of an agent's generalization capabilities.
    In the \rlb{} environment, it is possible to train an agent in an easy puzzle setting and subsequently evaluate its performance in progressively harder puzzle configurations.
    For most puzzles, the cardinality of the action space is independent of puzzle size.
    It is therefore also possible to train an agent only on small instances of a puzzle and then evaluate it on larger sizes.
    This approach allows us to assess whether an agent has learned the correct underlying algorithm and generalizes to out-of-distribution scenarios.
    Secondly, it enables the benchmark to remain adaptable to the continuous advancements in RL methodologies.
    As RL algorithms evolve and become more capable, the puzzle configurations can be adjusted accordingly to maintain the desired level of difficulty.
    This ensures that the benchmark continues to effectively assess the capabilities of the latest RL methods.

\section{Empirical Evaluation} \label{sec:baselines}
    We evaluate the baseline performance of numerous commonly used RL algorithms on our \rlb{} environment.
    Additionally, we also analyze the impact of certain design decisions of the environment and the training setup.
    Our metric of interest is the average number of steps required by a policy to successfully complete a puzzle, where lower is better.
    We refer to the term \textit{successful episode} to denote the successful completion of a single puzzle instance.
    We also look at the success rate, i.e. what percentage of the puzzles was completed successfully.

    To provide an understanding of the puzzle's complexity and to contextualize the agents' performance, we include an upper-bound estimate of the optimal number of steps required to solve the puzzle correctly.
    This estimate is a combination of both the steps required to solve the puzzle using an optimal strategy, and an upper bound on the environment steps required to achieve this solution, such as moving the cursor to the correct position.
    The upper bound is denoted as \textit{Optimal}.
    Please refer to \cref{tab:parameters} for details on how this upper bound is calculated for each puzzle.

    We run experiments based on all the RL algorithms presented in \cref{table:rl-algorithms}.
    We include both popular traditional algorithms such as PPO, as well as algorithms designed more specifically for the kinds of tasks presented in \rlb{}.
    Where possible, we used the implementations available in the RL library Stable Baselines~3~\citep{raffin2021baselines}, using the default hyper-parameters. 
    For MuZero and DreamerV3, we used the code available at \citep{muzero-general} and \citep{dreamerv3-code}, respectively.
    We provide a summary of all algorithms in Appendix \cref{table:rl-algorithms}.
    In total, our experiments required approximately 10'000 GPU hours.

    All selected algorithms are compatible with the discrete action space required by our environment.
    This circumstance prohibits the use of certain other common RL algorithms such as Soft-Actor Critic (SAC)~\citep{haarnoja2018soft} or Twin Delayed Deep Deterministic Policy Gradients (TD3)~\citep{fujimoto2018addressing}.

\subsection{Baseline Experiments}\label{sec:experiment-baseline}
    For the general baseline experiments, we trained all agents on all puzzles and evaluate their performance.
    Due to the challenging nature of our puzzles, we have selected an easy difficulty and small size of the puzzle where possible.
    Every agent was trained on the discrete internal state observation using five different random seeds.
    We trained all agents by providing rewards only at the end of each episode upon successful completion or failure. 
    For computational reasons, we truncated all episodes during training and testing at 10,000 steps.
    For such a termination, reward was kept at 0.
    We evaluate the effect of this episode truncation in \cref{experiments:ep_length}
    We provide all experimental parameters, including the exact parameters supplied for each puzzle in \cref{sec:experiment-parameters}.

    \begin{figure}
    \centering
    \includegraphics[width=0.5 \columnwidth]{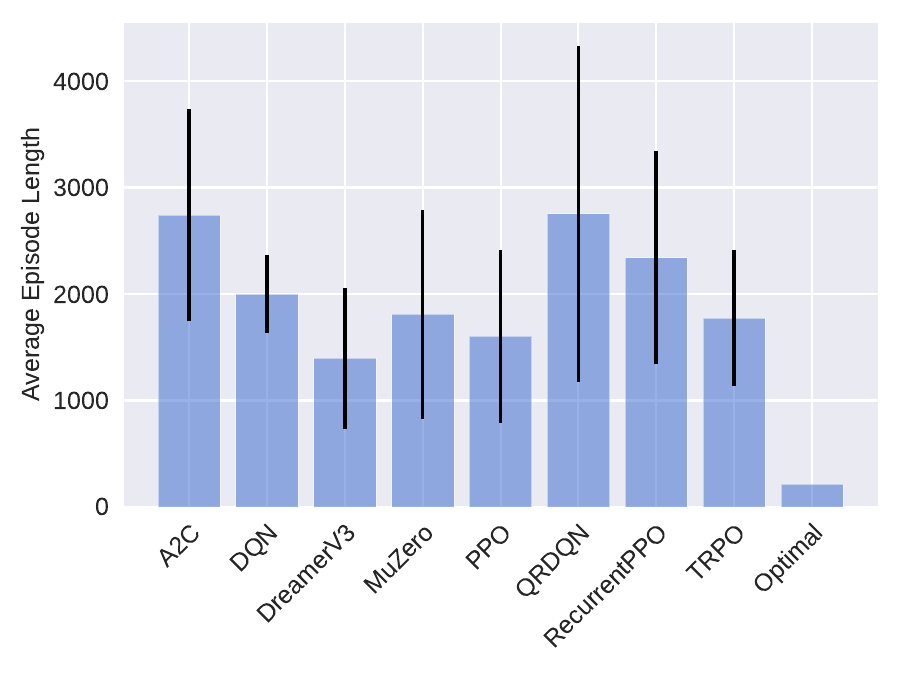}
    \caption{
        Average episode length of successful episodes for all evaluated algorithms on all puzzles in the easiest setting (lower is better).
        Some puzzles, namely Loopy, Pearl, Pegs, Solo, and Unruly, were intractable for all algorithms and were therefore excluded in this aggregation.
        The standard deviation is computed with respect to the performance over all evaluated instances for all trained seeds, aggregated for the total number of puzzles.
        Optimal refers the upper bound of the performance of an optimal policy, it therefore does not include a standard deviation.
        We see that DreamerV3 performs the best with an average episode length of 1334.
        However, this is still worse than the optimal upper bound at an average of 217 steps.
    }
    \label{fig:baseline-summary}
\end{figure}
    
    To track an agent's progress, we use episode lengths, i.e., how many actions an agent needs to solve a puzzle.
    A lower number of actions indicates a stronger policy that is closer to the optimal solution.
    To obtain the final evaluation, we run each policy on 1000 random episodes of the respective puzzle, again with a maximum step size of 10,000 steps.
    All experiments were conducted on NVIDIA 3090 GPUs.
    The training time for a single agent with 2 million PPO steps varied depending on the puzzle and ranged from approximately 1.75 to 3 hours.
    The training for DreamerV3 and MuZero was more demanding and training time ranged from approximately 10 to 20 hours.

    \cref{fig:baseline-summary} shows the average successful episode length for all algorithms.
    It can be seen that DreamerV3 performs best while PPO also achieves good performance, closely followed by TRPO and MuZero.
    This is especially interesting since PPO and TRPO follow much simpler training routines than DreamerV3 and MuZero.
    It seems that the implicit world models learned by DreamerV3 struggle to appropriately capture some puzzles.
    The high variance of MuZero may indicate some instability during training or the need for puzzle-specific hyperparamater tuning.
    Upon closer inspection of the detailed results, presented in Appendix \cref{tab:phat_table_state} and \ref{tab:phat_table_state_cont}, DreamerV3 manages to solve 62.7\% of all puzzle instances.
    In 14 out of the 40 puzzles, it has found a policy that solves the puzzles within the \textit{Optimal} upper bound.
    PPO and TRPO managed to solve an average of 61.6\% and 70.8\% of the puzzle instances, however only 8 and 11 of the puzzles have consistently solved within the \textit{Optimal} upper bound.
    The algorithms A2C, RecurrentPPO, DQN and QRDQN perform worse than a pure random policy.
    Overall, it seems that some of the environments in \rlb{} are quite challenging and well suited to show the difference in performance between algorithms.
    It is also important to note that all the logic puzzles are designed so that they can be solved without requiring any guesswork.

\subsection{Difficulty}
    
    We further evaluate the performance of a subset of the puzzles on the easiest preset difficulty level for humans. 
    We selected all puzzles where a random policy was able to solve them with a probability of at least 10\%, which are Netslide, Same Game and Untangle.
    By using this selection, we estimate that the reward density should be relatively high, ideally allowing the agent to learn a good policy.
    Again, we train all algorithms listed in \cref{table:rl-algorithms}.
    We provide results for the two strongest algorithms, PPO and DreamerV3 in \cref{tab:human-easy-comparison}, with complete results available in Appendix \cref{tab:phat_table_state}.
    Note that as part of \cref{experiments:ep_length}, we also perform ablations using DreamerV3 on more puzzles on the easiest preset difficulty level for humans.

    \begin{table*}[h]
        \caption{
            {
            Comparison of how many steps agents trained with PPO and DreamerV3 need on average to solve puzzles of two difficulty levels. In brackets, the percentage of successful episodes is reported. The difficulty levels correspond to the overall easiest and the easiest-for-humans settings. We also give the upper bound of optimal steps needed for each configuration.
            }
        }
        \label{tab:human-easy-comparison}
            \centering
            {
            \small
            \begin{tabular}{lllll}
                \toprule
                \textbf{Puzzle} & \textbf{Parameters} & \textbf{PPO} & \textbf{DreamerV3} & \textbf{\# Optimal Steps} \\
                \midrule
                \multirow{2}{*}{Netslide} & \texttt{2x3b1} & $35.3\pm0.7$ \hfill (100.0\%) & $12.0\pm0.4$ \hfill(100.0\%) & 48\\*
                & \texttt{3x3b1} & $4742.1\pm2960.1$ \hfill (9.2\%) & $3586.5\pm676.9$ \hfill (22.4\%) & 90 \\
                \midrule
                \multirow{2}{*}{Same Game} & \texttt{2x3c3s2} & $11.5\pm0.1$ \hfill (100.0\%) & $7.3\pm0.2$ \hfill (100.0\%) & 42\\*
                & \texttt{5x5c3s2} & $1009.3\pm1089.4$ \hfill (30.5\%) & $527.0\pm162.0$ \hfill (30.2\%) & 300 \\
                \midrule
                \multirow{2}{*}{Untangle} & \texttt{4} & $34.9\pm10.8$ \hfill (100.0\%) & $6.3\pm0.4$ \hfill (100.0\%) & 80\\*
                & \texttt{6} & $2294.7\pm2121.2$ \hfill (96.2\%) & $1683.3\pm73.7$ \hfill (82.0\%) & 150 \\
                \bottomrule\\
            \end{tabular}
            }
    \end{table*} 
    
    We can see that for both PPO and DreamerV3, the percentage of successful episodes decreases, with a large increase in steps required.
    DreamerV3 performs clearly stronger than PPO, requiring consistently fewer steps, but still more than the optimal policy.
    Our results indicate that puzzles with relatively high reward density at human difficulty levels remain challenging.
    We propose to use the easiest human difficulty level as a first measure to evaluate future algorithms. The details of the easiest human difficulty setting can be found in Appendix \cref{tab:experiment-parameters}.
    If this level is achieved, difficulty can be further scaled up by increasing the size of the puzzles.
    Some puzzles also allow for an increase in difficulty with fixed size.

\subsection{Effect of Action Masking and Observation Representation} \label{sec:experiments_am_obs}
\begin{figure}[htb]
        \centering
        \includegraphics[height=5cm]{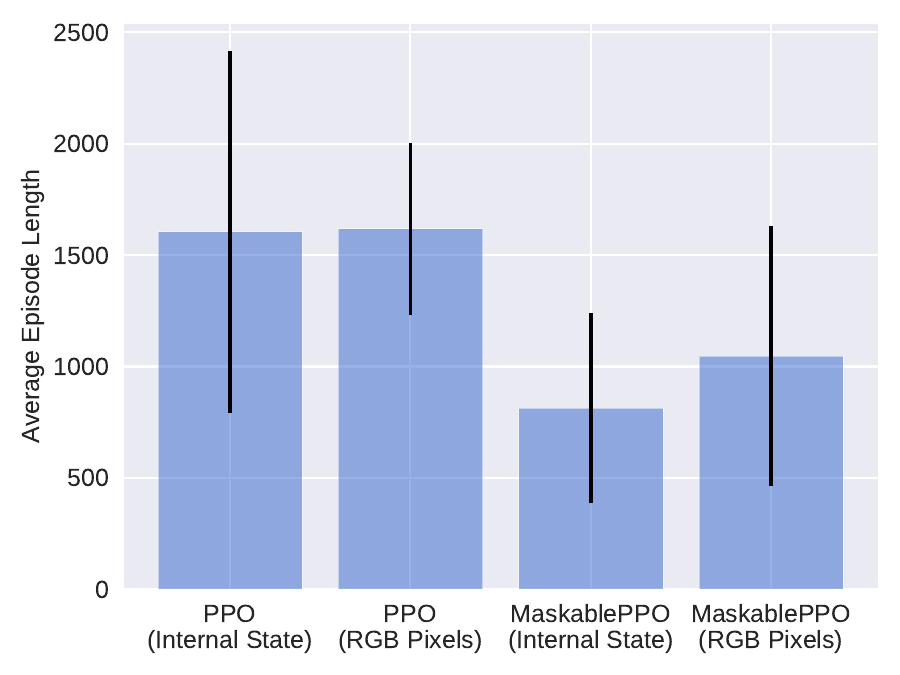}
        \hfill
        \includegraphics[height=5cm]{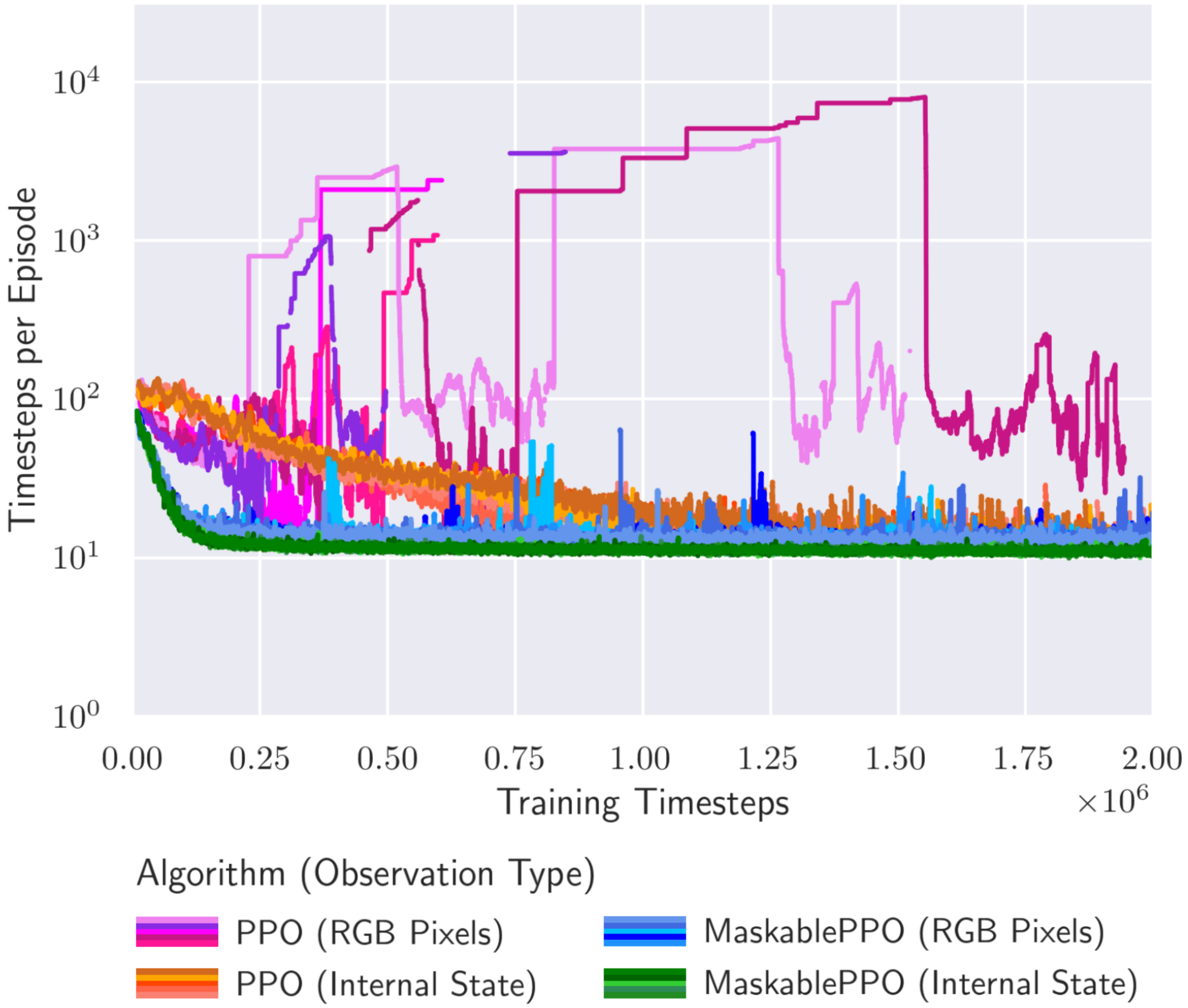}

        \caption{(left) We demonstrate the effect of action masking in both RGB observation and internal game state. By masking moves that do not change the current state, the agent requires fewer actions to explore, and therefore, on average solves a puzzle using fewer steps. (right) Moving average episode length during training for the \textit{Flood} puzzle.
        Lower episode length is better, as the episode gets terminated as soon as the agent has solved a puzzle.
        Different colors describe different algorithms, where different shades of a color indicate different random seeds.
        Sparse dots indicate that an agent only occasionally managed to find a policy that solves a puzzle.
        It can be seen that both the use of discrete internal state observations and action masking have a positive effect on the training, leading to faster convergence and a stronger overall performance.}
        \label{fig:training_performance_pixel_action_masking}
    \end{figure}

    We evaluate the effect of action masking, as well as observation type, on training performance.
    Firstly, we analyze whether action masking, as described in paragraph ``Action Masking'' in \cref{par:action_masking}, can positively affect training performance.
    Secondly, we want to see if agents are still capable of solving puzzles while relying on pixel observations.
    Pixel observations allow for the exact same input representation to be used for all puzzles, thus achieving a setting that is very similar to the Atari benchmark.
    We compare MaskablePPO to the default PPO without action masking on both types of observations.
    We summarize the results in \cref{fig:training_performance_pixel_action_masking}.
    Detailed results for masked RL agents on the pixel observations are provided in Appendix \cref{tab:phat_table_masked}.

    As we can observe in \cref{fig:training_performance_pixel_action_masking}, action masking has a strongly positive effect on training performance.
    This benefit is observed both in the discrete internal game state observations and on the pixel observations.
    We hypothesize that this is due to the more efficient exploration, as actions without effect are not allowed.
    As a result, the reward density during training is increased, and agents are able to learn a better policy.
    Particularly noteworthy are the outcomes related to \textit{Pegs}.
    They show that an agent with action masking can effectively learn a successful policy, while a random policy without action masking consistently fails to solve any instance.
    As expected, training RL agents on pixel observations increases the difficulty of the task at hand.
    The agent must first understand how the pixel observation relates to the internal state of the game before it is able to solve the puzzle.
    Nevertheless, in combination with action masking, the agents manage to solve a large percentage of all puzzle instances, with 10 of the puzzles consistently solved within the optimal upper bound.
    
    Furthermore, \cref{fig:training_performance_pixel_action_masking} shows the individual training performance on the puzzle \textit{Flood}.
    It can be seen that RL agents using action masking and the discrete internal game state observation converge significantly faster and to better policies compared to the baselines. 
    The agents using pixel observations and no action masking  struggle to converge to any reasonable policy.

\subsection{Effect of Episode Length and Early Termination}
    \label{experiments:ep_length}
    We evaluate whether the cutoff episode length or early termination have an effect on training performance of the agents.
    For computational reasons, we perform these experiments on a selected subset of the puzzles on human level difficulty and only for DreamerV3 (see \cref{app:ep_length_params} for details).
    As we can see in \cref{tab:ep_length_termination_ablation}, increasing the maximum episode length during training from 10,000 to 100,000 does not improve performance. 
    Only when episodes get terminated after visiting the exact same state more than 10 times, the agent is able to solve more puzzle instances on average (31.5\% vs. 25.2\%).
    Given the sparse reward structure, terminating episodes early seems to provide a better trade-off between allowing long trajectories to successfully complete and avoiding wasting resources on unsuccessful trajectories.
    \begin{table}[h]
        \caption{
            {
            Comparison of the effect of the maximum episode length (\# Steps) and early termination (ET) on final performance.
            For each setting, we report average success episode length with standard deviation with respect to the random seed, all averaged over all selected puzzles.
            In brackets, the percentage of successful episodes is reported.
            }
        }
        \label{tab:ep_length_termination_ablation}
            \centering
            {
            \small
            \begin{tabular}{lll}
                \toprule
                \textbf{\#Steps} & \textbf{ET} & \textbf{DreamerV3} \\
                \midrule
                \multirow{2}{*}{$1e5$} & 10 & $2950.9 \pm 1260.2$ \hfill (31.6\%) \\
                & - & $2975.4 \pm 1503.5$ \hfill (25.2\%) \\
                \midrule
                \multirow{2}{*}{$1e4$} & 10 & $3193.9 \pm 1044.2$ \hfill (26.1\%) \\
                 & - & $2892.4 \pm 908.3$ \hfill (26.8\%) \\
                \bottomrule
            \end{tabular}
            }
    \end{table} 

\subsection{Generalization}
\rlb{} is explicitly designed to facilitate the testing of generalization capabilities of agents with respect to different puzzle sizes or puzzle difficulties.
For our experiments, we select puzzles with the highest reward density.
We utilize a a custom observation wrapper and transformer-based encoder in order for the agent to be able to work with different input sizes, see \cref{app:custom_observation,app:generalization_example} for details. We call this approach PPO (Transformer)

    \begin{table*}[b]
        \caption{
            {
            We test generalization capabilities of agents by evaluating them on puzzle sizes larger than their training environment.
            We report the average number of steps an agent needs to solve a puzzle, and the percentage of successful episodes in brackets.
            The difficulty levels correspond to the overall easiest and the easiest-for-humans settings.
            For PPO (Transformer), we selected the best checkpoint during training according to the performance in the training environment.
            For PPO (Transformer)$^\dagger$, we selected the best checkpoint during training according to the performance in the generalization environment.
            }
        }
        \label{tab:generalization}
            \centering
            {
            \small
            \begin{tabular}{llllll}
                \toprule
                \textbf{Puzzle} & \textbf{Parameters} & \textbf{Trained on} & \textbf{PPO (Transformer)} &  \textbf{PPO (Transformer)$^\dagger$} \\
                \midrule
                \multirow{2}{*}{Netslide} & \texttt{2x3b1} & \checkmark & $244.1\pm313.7$ \hfill (100.0\%) & $242.0\pm379.3$ \hfill(100.0\%)\\*
                & \texttt{3x3b1} & \xmark & $9014.6\pm2410.6$ \hfill (18.6\%) & $9002.8\pm2454.9$ \hfill (18.0\%) \\
                \midrule
                \multirow{2}{*}{Same Game} & \texttt{2x3c3s2} & \checkmark & $9.3\pm10.9$ \hfill (99.8\%) & $26.2\pm52.9$ \hfill (99.7\%)\\*
                & \texttt{5x5c3s2} &  \xmark & $379.0\pm261.6$ \hfill (9.4\%) & $880.1\pm675.4$ \hfill (18.1\%) \\
                \midrule
                \multirow{2}{*}{Untangle} & \texttt{4} & \checkmark & $38.6\pm58.2$ \hfill (99.8\%) & $69.8\pm66.4$ \hfill (100.0\%)\\*
                & \texttt{6} & \xmark & $3340.0\pm3101.2$ \hfill (87.3\%) & $2985.8\pm2774.7$ \hfill (93.7\%) \\
                \bottomrule\\
            \end{tabular}
            }
    \end{table*}

The results presented in \cref{tab:generalization} indicate that while it is possible to learn a policy that generalizes it remains a challenging problem.
Furthermore, it can be observed that selecting the best model during training according to the performance on the generalization environment yields a performance benefit in that setting.
This suggests that agents may learn a policy that generalizes better during the training process, but then overfit on the environment they are training on.
It is also evident that generalization performance varies substantially across different random seeds.
For Netslide, the best agent is capable of solving 23.3\% of the puzzles in the generalization environment whereas the worst agent is only able to solve 11.2\% of the puzzles, similar to a random policy.
Our findings suggest that agents are generally capable of generalizing to more complex puzzles.
However, further research is necessary to identify the appropriate inductive biases that allow for consistent generalization without a significant decline in performance.

\section{Discussion}
    The experimental evaluation demonstrates varying degrees of success among different algorithms.
    For instance, puzzles such as \textit{Tracks}, \textit{Map} or \textit{Flip} were not solvable by any of the evaluated RL agents, or only with performance similar to a random policy.
    This points towards the potential of intermediate rewards, better game rule-specific action masking, or model-based approaches.
    To encourage exploration in the state space, a mechanism that explicitly promotes it may be beneficial.
    On the other hand, the fact that some algorithms managed to solve a substantial amount of puzzles with presumably optimal performance demonstrates the advances in the field of RL.
    In light of the promising results of DreamerV3, the improvement of agents that have certain reasoning capabilities and an implicit world model by design stay an important direction for future research.

    \paragraph{Experimental Results.} The experimental results presented in \cref{sec:experiment-baseline} and \cref{sec:experiments_am_obs} underscore the positive impact of action masking and the correct observation type on performance.
    While a pixel representation would lead to a uniform observation for all puzzles, it currently increases complexity too much compared the discrete internal game state.
    Our findings indicate that incorporating action masking significantly improves the training efficiency of reinforcement learning algorithms. 
    This enhancement was observed in both discrete internal game state observations and pixel observations.
    The mechanism for this improvement can be attributed to enhanced exploration, resulting in agents being able to learn more robust and effective policies.
    This was especially evident in puzzles where unmasked agents had considerable difficulty, thus showcasing the tangible advantages of implementing action masking for these puzzles.

\paragraph{Limitations.}

    While the \rlb{} framework provides the ability to gain comprehensive insights into the performance of various RL algorithms on logic puzzles, it is crucial to recognize certain limitations when interpreting results. 
    The sparse rewards used in this baseline evaluation add to the complexity of the task. 
    Moreover, all algorithms were evaluated with their default hyper-parameters. 
    Additionally, the constraint of discrete action spaces excludes the application of certain RL algorithms.

    In summary, the different challenges posed by the logic-requiring nature of these puzzles necessitates a good reward system, strong guidance of agents, and an agent design more focused on logical reasoning capabilities.
    It will be interesting to see how alternative architectures such as graph neural networks (GNNs) perform. 
    GNNs are designed to align more closely with the algorithmic solution of many puzzles.
    While the notion that ``reward is enough'' \citep{silver2021reward,vamplew2022scalar} might hold true, our results indicate that not just \textit{any} form of correct reward will suffice, and that advanced architectures might be necessary to learn an optimal solution.

\section{Conclusion}\label{sec:future-work-and-conclusion}
    In this work, we have proposed \rlb{}, a benchmark that bridges the gap between algorithmic reasoning and RL.
    In addition to containing a rich diversity of logic puzzles, \rlb{} also offers an adjustable difficulty progression for each puzzle, making it a useful tool for benchmarking, evaluating and improving RL algorithms.
    Our empirical evaluation shows that while RL algorithms exhibit varying degrees of success, challenges persist, particularly in puzzles with higher complexity or those requiring nuanced logical reasoning.
    We are excited to share \rlb{} with the broader research community and hope that \rlb{} will foster further research for improving the algorithmic reasoning abilities of RL algorithms.

\section*{Broader Impact}
This paper aims to contribute to the advancement of the field of Machine Learning (ML).
Given the current challenges in ML related to algorithmic reasoning, we believe that our newly proposed benchmark will facilitate significant progress in this area, potentially elevating the capabilities of ML systems.
Progress in algorithmic reasoning can contribute to the development of more transparent, explainable, and fair ML systems.
This can further help address issues related to bias and discrimination in automated decision-making processes, promoting fairness and accountability.

\newpage

\bibliography{bibliography}

\newpage
\appendix

\section{\rlb{} Environment Usage Guide}
    \subsection{General Usage}
        \label{app:rlp_usage}
            A Python code example for using the \rlb{} environment is provided in \cref{code:init-and-play-episode}. All puzzles support seeding the initialization, by adding \verb+#{seed}+ after the parameters, where \texttt{\{seed\}} is an \verb+int+. The allowed parameters are displayed in \cref{tab:parameters}. A full custom initialization argument would be as follows: \texttt{\{parameters\}\#\{seed\}}.
        
        \begin{lstlisting}[language=Python, caption={Code example of how to initialize an environment and have an agent complete one episode. The \rlb{} environment is designed to be compatible with the Gymnasium API. The choice of \texttt{Agent} is up to the user, it can be a trained agent or random policy.}, label=code:init-and-play-episode]
import gymnasium as gym
import rlp

# init an agent suitable for Gymnasium environments
agent = Agent.create()

# init the environment
env = gym.make('rlp/Puzzle-v0', puzzle="bridges", 
               render_mode="rgb_array", params="4x4#42")
observation, info = env.reset()

# complete an episode
terminated = False
while not terminated:
    action = agent.choose(env)  # the agent chooses the next action
    observation, reward, terminated, truncated, info = env.step(action)
env.close()
        \end{lstlisting}
    \subsection{Custom Reward}
        A Python code example for implementing a custom reward system is provided in \cref{code:custom-reward-wrapper}.  To this end, the environment's \verb+step()+ function provides the puzzle's internal state inside the \verb+info+ Python dict.
        \begin{lstlisting}[language=Python, caption={Code example of a custom reward implementation using Gymnasium's \texttt{Wrapper} class. A user can use the game state information provided in \texttt{info["puzzle\_state"]} to modify the rewards received by the agent after performing an action.}, label=code:custom-observation-wrapper]
import gymnasium as gym
class PuzzleRewardWrapper(gym.Wrapper):
    def step(self, action):
        obs, reward, terminated, truncated, info = self.env.step(action)
        # Modify the reward by using members of info["puzzle_state"]
        return obs, reward, terminated, truncated, info
        \end{lstlisting}
    
    \subsection{Custom Observation}
        \label{app:custom_observation}
        A Python code example for implementing a custom observation structure that is compatible with an agent using a transformer encoder.
        Here, we provide the example for Netslide, please refer to our \href{https://github.com/ETH-DISCO/rlp}{GitHub} for more examples.
        \begin{lstlisting}[language=Python, caption={Code example of a custom observation implementation using Gymnasium's \texttt{Wrapper} class. A user can use the all elements of rpovided in the \texttt{obs} dict to create a custom observation. In this code example, the resulting observation is suitable for a transformer-based encoder.}, label=code:custom-reward-wrapper, basicstyle=\tiny]
import gymnasium as gym
import numpy as np
class NetslideTransformerWrapper(gym.ObservationWrapper):
def __init__(self, env):
    super(NetslideTransformerWrapper, self).__init__(env)
    self.original_space = env.observation_space
   
    self.max_length = 512
    self.embedding_dim = 16 + 4
    self.observation_space = gym.spaces.Box(
        low=-1, high=1, shape=(self.max_length, self.embedding_dim,), dtype=np.float32
    )
    
    self.observation_space = gym.spaces.Dict(
        {'obs': self.observation_space,
        'len': gym.spaces.Box(low=0, high=self.max_length, shape=(1,),
        dtype=np.int32)}
    )

def observation(self, obs):
    # The original observation is an ordereddict with the keys ['barriers', 'cursor_pos', 'height',
    # 'last_move_col', 'last_move_dir', 'last_move_row', 'move_count', 'movetarget', 'tiles', 'width', 'wrapping']
    # We are only interested in 'barriers', 'tiles', 'cursor_pos', 'height' and 'width'
    barriers = obs['barriers']
    # each element of barriers is an uint16, signifying different elements
    barriers = np.unpackbits(barriers.view(np.uint8)).reshape(-1, 16)
    # add some positional embedding to the barriers
    embedded_barriers = np.concatenate(
        [barriers, self.pos_embedding(np.arange(barriers.shape[0]), obs['width'], obs['height'])], axis=1)
        
    tiles = obs['tiles']
    # each element of tiles is an uint16, signifying different elements
    tiles = np.unpackbits(tiles.view(np.uint8)).reshape(-1, 16)
    # add some positional embedding to the tiles
    embedded_tiles = np.concatenate(
        [tiles, self.pos_embedding(np.arange(tiles.shape[0]), obs['width'], obs['height'])], axis=1)
    cursor_pos = obs['cursor_pos']

    embedded_cursor_pos = np.concatenate(
        [np.ones((1, 16)), self.pos_embedding_cursor(cursor_pos, obs['width'], obs['height'])], axis=1)
        
    embedded_obs = np.concatenate([embedded_barriers, embedded_tiles, embedded_cursor_pos], axis=0)

    current_length = embedded_obs.shape[0]
    # pad with zeros to accomodate different sizes
    if current_length < self.max_length:
        embedded_obs = np.concatenate(
            [embedded_obs, np.zeros((self.max_length - current_length, self.embedding_dim))], axis=0)
    return {'obs': embedded_obs, 'len': np.array([current_length])}

@staticmethod
def pos_embedding(pos, width, height):
    # pos is an array of integers from 0 to width*height
    # width and height are integers
    # return a 2D array with the positional embedding, using sin and cos
    x, y = pos % width, pos // width
    # x and y are integers from 0 to width-1 and height-1
    pos_embed = np.zeros((len(pos), 4))
    pos_embed[:, 0] = np.sin(2 * np.pi * x / width)
    pos_embed[:, 1] = np.cos(2 * np.pi * x / width)
    pos_embed[:, 2] = np.sin(2 * np.pi * y / height)
    pos_embed[:, 3] = np.cos(2 * np.pi * y / height)
    return pos_embed

@staticmethod
def pos_embedding_cursor(pos, width, height):
    # cursor pos goes from -1 to width or height
    x, y = pos
    x += 1
    y += 1
    width += 1
    height += 1
    pos_embed = np.zeros((1, 4))
    pos_embed[0, 0] = np.sin(2 * np.pi * x / width)
    pos_embed[0, 1] = np.cos(2 * np.pi * x / width)
    pos_embed[0, 2] = np.sin(2 * np.pi * y / height)
    pos_embed[0, 3] = np.cos(2 * np.pi * y / height)
    return pos_embed
    
    \end{lstlisting}
    
    \subsection{Generalization Example}
    \label{app:generalization_example}
        In \cref{code:transformer-encoder}, we show how a transformer-based features extractor can be built for Stable Baseline 3's PPO MultiInputPolicy. Together with the observations from \cref{code:custom-observation-wrapper}, this feature extractor can work with variable-length inputs.
        This allows for easy evaluation in environments of different sizes than the environment the agent was originally trained in.
        \begin{lstlisting}[language=Python, caption={Code example of a transformer-based feature extractor written in PyTorch, compatible with Stable Baselines 3's PPO. This encoder design allows for variable-length inputs, enabling generalization to previously unseen puzzle sizes.},
        label=code:transformer-encoder, basicstyle=\tiny]
import gymnasium as gym
import numpy as np
from stable_baselines3.common.torch_layers import BaseFeaturesExtractor
from stable_baselines3 import PPO
import torch
import torch.nn as nn
from torch.nn import TransformerEncoder, TransformerEncoderLayer

class TransformerFeaturesExtractor(BaseFeaturesExtractor):
    def __init__(self, observation_space, data_dim, embedding_dim, nhead, num_layers, dim_feedforward, dropout=0.1):
        super(TransformerFeaturesExtractor, self).__init__(observation_space, embedding_dim)
        self.transformer = Transformer(embedding_dim=embedding_dim,
                                       data_dim=data_dim,
                                       nhead=nhead,
                                       num_layers=num_layers,
                                       dim_feedforward=dim_feedforward,
                                       dropout=dropout)

    def forward(self, observations: gym.spaces.Dict) -> torch.Tensor:
        # Extract the 'obs' key from the dict
        obs = observations['obs']
        length = observations['len']
        # all elements of length should be the same (we can't train on different puzzle sizes at the same time)
        length = int(length[0])
        obs = obs[:, :length]
        # Return the embedding of the cursor token (which is last)
        return self.transformer(obs)[:, -1, :]

class Transformer(nn.Module):
    def __init__(self, embedding_dim, data_dim, nhead, num_layers, dim_feedforward, dropout=0.1):
        super(Transformer, self).__init__()
        self.embedding_dim = embedding_dim
        self.data_dim = data_dim

        self.lin = nn.Linear(data_dim, embedding_dim)

        encoder_layers = TransformerEncoderLayer(
            d_model=self.embedding_dim,
            nhead=nhead,
            dim_feedforward=dim_feedforward,
            dropout=dropout,
            batch_first=True
        )

        self.transformer_encoder = TransformerEncoder(encoder_layers, num_layers)

    def forward(self, x):
        # x is of shape (batch_size, seq_length, embedding_dim)
        x = self.lin(x)
        transformed = self.transformer_encoder(x)
        return transformed

if __name__ == "__main__":
    policy_kwargs = dict(
        features_extractor_class=TransformerFeaturesExtractor,
        features_extractor_kwargs=dict(embedding_dim=args.transformer_embedding_dim,
                                       nhead=args.transformer_nhead,
                                       num_layers=args.transformer_layers,
                                       dim_feedforward=args.transformer_ff_dim,
                                       dropout=args.transformer_dropout,
                                       data_dim=data_dims[args.puzzle])
    )

    model = PPO("MultiInputPolicy",
                env,
                policy_kwargs=policy_kwargs,
                )
    \end{lstlisting}

\section{Environment Features}
\label{app:env_features}
\subsection{Episode Definition}
    An episode is played with the intention of solving a given puzzle.
    The episode begins with a newly generated puzzle and terminates in one of two states.
    To achieve a reward, the puzzle is either solved completely or the agent has failed irreversibly.
    The latter state is unlikely to occur, as only a few games, for example pegs or minesweeper, are able to terminate in a failed state.
    Alternatively, the episode can be terminated early.
    Starting a new episode generates a new puzzle of the same kind, with the same parameters such as size or grid type. 
    However, if the random seed is not fixed, the puzzle is likely to have a different layout from the puzzle in the previous episode.
\subsection{Observation Space}\label{par:observation-space}
    There are two kinds of observations which can be used by the agent.
    The first observation type is a representation of the discrete internal game state of the puzzle, consisting of a combination of arrays and scalars.
    This observation is provided by the underlying code of Tathams's puzzle collection.
    The composition and shape of the internal game state is different for each puzzle, which, in turn, requires the agent architecture to be adapted.

    The second type of observation is a representation of the pixel screen, given as an integer matrix of shape (3$\times$width$\times$height). The environment deals with different aspect ratios by adding padding.
    The advantage of the pixel representation is a consistent representation for all puzzles, similar to the Atari RL Benchmark~\citep{mnih2013atari}. It could even allow for a single agent to be trained on different puzzles.
    On the other hand, it forces the agent to learn to solve the puzzles only based on the visual representation of the puzzles, analogous to human players.
    This might increase difficulty as the agent has to learn the task representation implicitly.
\subsection{Action Space}\label{par:action_space}
    Natively, the puzzles support two types of input, mouse and keyboard.
    Agents in \rlb{} play the puzzles only through keyboard input.
    This is due to our decision to provide the discrete internal game state of the puzzle as an observation, for which mouse input would not be useful.
    
    The action space for each puzzle is restricted to actions that can actively contribute to changing the logical state of a puzzle.
    This excludes ``memory aides'' such as markers that signify the absence of a certain connection in \textit{Bridges} or adding candidate digits in cells in \textit{Sudoku}.
    The action space also includes possibly rule-breaking actions, as long as the game can represent the effect of the action correctly.

    The largest action space has a cardinality of 14, but most puzzles only have five to six valid actions which the agent can choose from.
    Generally, an action is in one of two categories: selector movement or game state change.
    Selector movement is a mechanism that allows the agent to select game objects during play.
    This includes for example grid cells, edges, or screen regions.
    The selector can be moved to the next object by four discrete directional inputs and as such represents an alternative to continuous mouse input.
    A game state change action ideally follows a selector movement action.
    The game state change action will then be applied to the selected object.
    The environment responds by updating the game state, for example by entering a digit or inserting a grid edge at the current selector position.
\subsection{Action Masking}\label{par:action_masking}
    The fixed-size action space allows an agent to execute actions that may not result in any change in game state.
    For example, the action of moving the selector to the right if the selector is already placed at the right border.
    The \rlb{} environment provides an action mask that marks all actions that change the state of the game.
    Such an action mask can be used to improve performance of model-based and even some model-free RL approaches.
    The action masking provided by \rlb{} does not ensure adherence to game rules, rule-breaking actions can most often still be represented as a change in the game state.
\subsection{Reward Structure}
    In the default implementation, the agent only receives a reward for completing an episode.
    Rewards consist of a fixed positive value for successful completion and a fixed negative value otherwise.
    This reward structure encourages an agent to solve a given puzzle in the least amount of steps possible.
    The \rlb{} environment provides the option to define intermediate rewards tailored to specific puzzles, which could help improve training progress. 
    This could be, for example, a negative reward if the agent breaks the rules of the game, or a positive reward if the agent correctly achieves a part of the final solution.
\subsection{Early Episode Termination}
    Most of the puzzles in \rlb{} do not have an upper bound on the number of steps, where the only natural end can be reached via successfully solving the puzzle.
    The \rlb{} environment also provides the option for early episode termination based on state repetitions.
    If an agent reaches the exact same game state multiple times, the episode can be terminated in order to prevent wasteful continuation of episodes that no longer contribute to learning or are bound to fail.

\section{\rlb{} Implementation Details}
\label{app:environment-implementation}

    In the following, a brief overview of \rlb's code implementation is given. The environment is written in both Python and C, in order to interface with Gymnasium~\citep{site:farama-gymnasium} as the RL toolkit and the C source code of the original puzzle collection. The original puzzle collection source code is available under the MIT License.\footnote{The source code and license are available at \url{https://www.chiark.greenend.org.uk/~sgtatham/puzzles/}.} In maintext \cref{fig:environment-landscape}, an overview of the environment and how it fits with external libraries is presented.
    The modular design in both \rlb{} and the Puzzle Collection's original code allows users to build and integrate new puzzles into the environment.

\paragraph{Environment Class} 
    The reinforcement learning environment is implemented in the Python class \verb+PuzzleEnv+ in the \verb+rlp+ package. It is designed to be compatible with the Gymnasium-style API for RL environments to facilitate easy adoption. As such, it provides the two important functions needed for progressing an environment, \verb+reset()+ and \verb+step()+. 

    Upon initializing a \texttt{PuzzleEnv}, a 2D surface displaying the environment is created. This surface and all changes to it are handled by the Pygame~\citep{site:pygame} graphics library. \rlb{} uses various functions provided in the library, such as shape drawing, or partial surface saving and loading. 

    The \verb+reset()+ function changes the environment state to the beginning of a new episode, usually by generating a new puzzle with the given parameters. An agent solving the puzzle is also reset to a new state. \verb+reset()+ also returns two variables, \verb+observation+ and \verb+info+, where \verb+observation+ is a Python \texttt{dict} containing a NumPy 3D array called \verb+pixels+ of size (3 $\times$ surface\_width $\times$ surface\_height).
    This NumPy array contains the RGB pixel data of the Pygame surface, as explained in \cref{par:observation-space}. The \texttt{info} dict contains a \texttt{dict} called \texttt{puzzle\_state}, representing a copy of the current internal data structures containing the logical game state, allowing the user to create custom rewards.

    The \verb+step()+ function increments the time in the environment by one step, while performing an action chosen from the action space. Upon returning, \verb+step()+ provides the user with five variables, listed in \cref{tab:step-function-return-values}.

\begin{table}[h]
    \caption{
        Return values of the environment's \texttt{step()} function. 
        This information can then be used by an RL framework to train an agent.
    }
    \label{tab:step-function-return-values}
    \centering
    \begin{tabular}{ll}
    \toprule
    \textbf{Variable} & \textbf{Description} \\
    \midrule
    \texttt{observation} & 3D NumPy array containing RGB pixel data\\
    \texttt{reward} & The cumulative reward gained throughout all steps of the episode\\
    \texttt{terminated} & A \texttt{bool} stating whether an episode was completed by the agent\\
    \texttt{truncated} & A \texttt{bool} stating whether an episode was ended early, for example by reaching\\&
    the maximum allowed steps for an episode\\
    \texttt{info} & A \texttt{dict} containing a copy of the internal game state\\
    \bottomrule\\
    \end{tabular}
\end{table}
\paragraph{Intermediate Rewards}\label{sec:custom-rewards}
    The environment encourages the use of Gymnasium's \verb+Wrapper+ interface to implement custom reward structures for a given puzzle.
    Such custom reward structures can provide an easier game setting, compared to the sparse reward only provided when finishing a puzzle.

\paragraph{Puzzle Module}
    The \texttt{PuzzleEnv} object creates an instance of the class \texttt{Puzzle}. A \texttt{Puzzle} is essentially the glue between all Pygame surface tasks and the C back-end that contains the puzzle logic. To this end, it initializes a Pygame window, on which shapes and text are drawn. The \texttt{Puzzle} instance also loads the previously compiled shared library containing the C back-end code for the relevant puzzle.

    The \texttt{PuzzleEnv} also converts and forwards keyboard inputs (which are for example given by an RL agent's action) into the format the C back-end understands.

\paragraph{Compiled C Code}
    The C part of the environment sits on top of the highly-optimized original puzzle collection source code as a custom front-end, as detailed in the collection's developer documentation~\citep{site:sgt-dev-documentation}. Similar to other front-end types, it represents the bridge between the graphics library that is used to display the puzzles and the game logic back-end. Specifically, this is done using Python API calls to Pygame's drawing facilities. 

\section{Puzzle Descriptions}
\FloatBarrier
We provide short descriptions of each puzzle from \href{https://www.chiark.greenend.org.uk/~sgtatham/puzzles/}{www.chiark.greenend.org.uk/~sgtatham/puzzles/}. For detailed instructions for each puzzle, please visit the docs available at \href{https://www.chiark.greenend.org.uk/~sgtatham/puzzles/doc/index.html}{www.chiark.greenend.org.uk/~sgtatham/puzzles/doc/index.html}
\input{img/puzzles/puzzle_collection}
\FloatBarrier
    
\section{Puzzle-specific Metadata}\label{app:puzzle-specific}

\subsection{Action Space}\label{sec:action-space}
    We display the action spaces for all supported puzzles in \cref{tab:action-spaces}. The action spaces vary in size and in the types of actions they contain. As a result, an agent must learn the meaning of each action independently for each puzzle.
\begin{table}[htb]
  \caption{The action spaces for each puzzle are listed, along with their cardinalities. The actions are listed with their name in the original Puzzle Collection C code.}
  \label{tab:action-spaces}
  \centering
  \begin{tabular}{lll}
    \toprule
    \textbf{Puzzle}     & \textbf{Cardinality} & \textbf{Action space}      \\
    \midrule
    Black Box &  5 & \verb+UP+, \verb+DOWN+, \verb+LEFT+, \verb+RIGHT+, \verb+SELECT+       \\
    Bridges & 5 & \verb+UP+, \verb+DOWN+, \verb+LEFT+, \verb+RIGHT+, \verb+SELECT+       \\
    Cube & 4 & \verb+UP+, \verb+DOWN+, \verb+LEFT+, \verb+RIGHT+       \\
    Dominosa & 5 & \verb+UP+, \verb+DOWN+, \verb+LEFT+, \verb+RIGHT+, \verb+SELECT+       \\
    Fifteen & 4 & \verb+UP+, \verb+DOWN+, \verb+LEFT+, \verb+RIGHT+       \\
    Filling & 13 & \verb+UP+, \verb+DOWN+, \verb+LEFT+, \verb+RIGHT+, \verb+1+, \verb+2+, \verb+3+, \verb+4+, \verb+5+, \verb+6+, \verb+7+, \verb+8+, \verb+9+       \\
    Flip & 5 & \verb+UP+, \verb+DOWN+, \verb+LEFT+, \verb+RIGHT+, \verb+SELECT+       \\
    Flood & 5 & \verb+UP+, \verb+DOWN+, \verb+LEFT+, \verb+RIGHT+, \verb+SELECT+       \\
    Galaxies & 5 & \verb+UP+, \verb+DOWN+, \verb+LEFT+, \verb+RIGHT+, \verb+SELECT+       \\
    Guess & 5 & \verb+UP+, \verb+DOWN+, \verb+LEFT+, \verb+RIGHT+, \verb+SELECT+       \\
    Inertia & 9 & \verb+1+, \verb+2+, \verb+3+, \verb+4+, \verb+6+, \verb+7+, \verb+8+, \verb+9+, \verb+UNDO+       \\
    Keen & 14 & \verb+UP+, \verb+DOWN+, \verb+LEFT+, \verb+RIGHT+, \verb+SELECT2+, \verb+1+, \verb+2+, \verb+3+, \verb+4+, \verb+5+, \verb+6+, \verb+7+, \verb+8+, \verb+9+       \\
    Light Up & 5 & \verb+UP+, \verb+DOWN+, \verb+LEFT+, \verb+RIGHT+, \verb+SELECT+       \\
    Loopy & 6 & \verb+UP+, \verb+DOWN+, \verb+LEFT+, \verb+RIGHT+, \verb+SELECT+, \verb+SELECT2+     \\
    Magnets & 6 & \verb+UP+, \verb+DOWN+, \verb+LEFT+, \verb+RIGHT+, \verb+SELECT+, \verb+SELECT2+     \\
    Map & 5 & \verb+UP+, \verb+DOWN+, \verb+LEFT+, \verb+RIGHT+, \verb+SELECT+       \\
    Mines & 7 & \verb+UP+, \verb+DOWN+, \verb+LEFT+, \verb+RIGHT+, \verb+SELECT+, \verb+SELECT2+, \verb+UNDO+     \\
    Mosaic & 6 & \verb+UP+, \verb+DOWN+, \verb+LEFT+, \verb+RIGHT+, \verb+SELECT+, \verb+SELECT2+     \\
    Net & 5 & \verb+UP+, \verb+DOWN+, \verb+LEFT+, \verb+RIGHT+, \verb+SELECT+       \\
    Netslide & 5 & \verb+UP+, \verb+DOWN+, \verb+LEFT+, \verb+RIGHT+, \verb+SELECT+       \\
    Palisade & 5 & \verb+UP+, \verb+DOWN+, \verb+LEFT+, \verb+RIGHT+, \verb+CTRL+       \\
    Pattern & 6 & \verb+UP+, \verb+DOWN+, \verb+LEFT+, \verb+RIGHT+, \verb+SELECT+, \verb+SELECT2+     \\
    Pearl & 5 & \verb+UP+, \verb+DOWN+, \verb+LEFT+, \verb+RIGHT+, \verb+SELECT+       \\
    Pegs & 6 & \verb+UP+, \verb+DOWN+, \verb+LEFT+, \verb+RIGHT+, \verb+SELECT+, \verb+UNDO+       \\
    Range & 5 & \verb+UP+, \verb+DOWN+, \verb+LEFT+, \verb+RIGHT+, \verb+SELECT+       \\
    Rectangles & 5 & \verb+UP+, \verb+DOWN+, \verb+LEFT+, \verb+RIGHT+, \verb+SELECT+       \\
    Same Game & 6 & \verb+UP+, \verb+DOWN+, \verb+LEFT+, \verb+RIGHT+, \verb+SELECT+, \verb+UNDO+       \\
    Signpost & 6 & \verb+UP+, \verb+DOWN+, \verb+LEFT+, \verb+RIGHT+, \verb+SELECT+, \verb+SELECT2+     \\
    Singles & 6 & \verb+UP+, \verb+DOWN+, \verb+LEFT+, \verb+RIGHT+, \verb+SELECT+, \verb+SELECT2+     \\
    Sixteen & 6 & \verb+UP+, \verb+DOWN+, \verb+LEFT+, \verb+RIGHT+, \verb+SELECT+, \verb+SELECT2+     \\
    Slant & 6 & \verb+UP+, \verb+DOWN+, \verb+LEFT+, \verb+RIGHT+, \verb+SELECT+, \verb+SELECT2+     \\
    Solo & 13 & \verb+UP+, \verb+DOWN+, \verb+LEFT+, \verb+RIGHT+, \verb+1+, \verb+2+, \verb+3+, \verb+4+, \verb+5+, \verb+6+, \verb+7+, \verb+8+, \verb+9+       \\
    Tents & 6 & \verb+UP+, \verb+DOWN+, \verb+LEFT+, \verb+RIGHT+, \verb+SELECT+, \verb+SELECT2+     \\
    Towers & 14 & \verb+UP+, \verb+DOWN+, \verb+LEFT+, \verb+RIGHT+, \verb+SELECT2+, \verb+1+, \verb+2+, \verb+3+, \verb+4+, \verb+5+, \verb+6+, \verb+7+, \verb+8+, \verb+9+     \\
    Tracks & 5 & \verb+UP+, \verb+DOWN+, \verb+LEFT+, \verb+RIGHT+, \verb+SELECT+\\  
    Twiddle & 6 & \verb+UP+, \verb+DOWN+, \verb+LEFT+, \verb+RIGHT+, \verb+SELECT+, \verb+SELECT2+     \\
    Undead & 8 & \verb+UP+, \verb+DOWN+, \verb+LEFT+, \verb+RIGHT+, \verb+SELECT2+, \verb+1+, \verb+2+, \verb+3+     \\
    Unequal & 13 & \verb+UP+, \verb+DOWN+, \verb+LEFT+, \verb+RIGHT+, \verb+1+, \verb+2+, \verb+3+, \verb+4+, \verb+5+, \verb+6+, \verb+7+, \verb+8+, \verb+9+       \\
    Unruly & 6 & \verb+UP+, \verb+DOWN+, \verb+LEFT+, \verb+RIGHT+, \verb+SELECT+, \verb+SELECT2+     \\
    Untangle & 5 & \verb+UP+, \verb+DOWN+, \verb+LEFT+, \verb+RIGHT+, \verb+SELECT+ \\
    \bottomrule
  \end{tabular}
\end{table}
\FloatBarrier
\newpage
\subsection{Optional Parameters}\label{sec:custom-parameters}
    We display the optional parameters for all supported puzzles in \cref{tab:parameters}.
    If none are supplied upon initialization, a set of default parameters gets used for the puzzle generation process.

{\scriptsize
\begin{longtable}[hc]{lllll}
    \caption{For each puzzle, all optional parameters a user may supply are shown and described. We also give the required data type of variable, where applicable (e.g., \texttt{int} or \texttt{char}). For parameters that accept one of a few choices (such as difficulty), the accepted values and corresponding explanation are given in braces. As as example: a difficulty parameter is listed as \texttt{d\{int\}} with allowed values \{0~=~easy, 1 = medium, 2 = hard\}. In this case, choosing medium difficulty would correspond to \texttt{d1}.} \\
    \toprule  \multicolumn{1}{l}{\label{tab:parameters}\textbf{Puzzle}} & \multicolumn{1}{l}{\textbf{Example}} & \multicolumn{1}{l}{\textbf{Parameter}} & \multicolumn{1}{l}{\textbf{Description}} & \multicolumn{1}{l}{\textbf{Optimal Step Upper Bound}}\\ \midrule 
    \endfirsthead

    \multicolumn{4}{c}%
    {\tablename\ \thetable{} -- continued from previous page} \\
    \toprule \multicolumn{1}{l}{\textbf{Puzzle}} & \multicolumn{1}{l}{\textbf{Example}} & \multicolumn{1}{l}{\textbf{Parameter}} & \multicolumn{1}{l}{\textbf{Description}} & \multicolumn{1}{l}{\textbf{Optimal Step Upper Bound}}\\ \midrule 
    \endhead
    
    \multicolumn{4}{r}{{Continued on next page}} \\ 
    \endfoot
    
    \bottomrule
    \endlastfoot

    Black Box & \verb+w8h8m5M5+ & \verb+w{int}+ & grid width & (w$\cdot$ h + w + h + 1) 
       \\*
                                                      && \verb+h{int}+ & grid height   & $\cdot$ (w + 2) $\cdot$ (h + 2)    \\*
                                                      && \verb+m{int}+ & minimum number of balls       \\*
                                                      && \verb+M{int}+ & maximum number of balls       \\
    \midrule
    Bridges & \verb+7x7i5e2m2d0+ & \verb+{int}x{int}+ &  grid width $\times$ grid height & 3 $\cdot$ w $\cdot$ h $\cdot$ (w + h + 8)
 \\*
                                                      && \verb+i{int}+ & percentage of island squares       \\*
                                                      && \verb+e{int}+ & expansion factor       \\*
                                                      && \verb+m{int}+ & max bridges per direction       \\*
                                                      && \verb+d{int}+ & difficulty \{0 = easy, 1 = medium, 2 = hard\}       \\
    \midrule
    Cube & \verb+c4x4+ &  \verb+{char}+ & type \{c = cube, t = tetrahedron, &  w $\cdot$ h $\cdot$ F\\*
    &&&o = octahedron, i = icosahedron\} & F = number of the body's faces
   \\*
                                                      && \verb+{int}x{int}+ &  grid width $\times$ grid height       \\
    \midrule
    Dominosa & \verb+6db+ &  \verb+{int}+ &  maximum number of dominoes & $\frac{1}{2}\left(\text{w}^2\text{ + 3w + 2}\right) $
 \\*
                                                      && \verb+d{char}+ &  difficulty \{t = trivial, b = basic, h = hard, & $\cdot (\text{4}\sqrt{\text{w}^2\text{ + 3w + 2}}\text{  + 1})$\\*&&& 
                                                                                            e = extreme, a = ambiguous\}       \\
    \midrule
    Fifteen & \verb+4x4+ &  \verb+{int}x{int}+ &  grid width $\times$ grid height &   $(w \cdot h)^4$   \\
    \midrule
    Filling & \verb+13x9+ &   \verb+{int}x{int}+ &  grid width $\times$ grid height &  $(w \cdot h) \cdot (w + h + 1)$  \\
    \midrule
    Flip & \verb+5x5c+ &  \verb+{int}x{int}+ &  grid width $\times$ grid height  &    $(w \cdot h) \cdot (w + h + 1)$ \\*
                                                      && \verb+{char}+ & type \{c = crosses, r = random\}       \\
    \midrule
    Flood & \verb+12x12c6m5+ &  \verb+{int}x{int}+ &  grid width $\times$ grid height & $(w \cdot h) \cdot (w + h + 1)$ \\*
                                                      && \verb+c{int}+ & number of colors       \\*
                                                      && \verb+m{int}+ & extra moves permitted (above the\\*&&& solver's minimum)      \\
    \midrule
    Galaxies & \verb+7x7dn+ &  \verb+{int}x{int}+ &  grid width $\times$ grid height & $(2 \cdot w \cdot h - w - h) $ \\*
                                                      && \verb+d{char}+ &  difficulty \{n = normal, u = unreasonable\} & $\cdot (2 \cdot w + 2 \cdot h + 1)$       \\
    \midrule
    Guess & \verb+c6p4g10Bm+ &  \verb+c{int}+ &  number of colors &    $(p+1) \cdot g \cdot (c + p)$ \\*
                                                      && \verb+p{int}+ & pegs per guess       \\*
                                                      && \verb+g{int}+ & maximum number of guesses       \\*
                                                      && \verb+{char}+ & allow blanks  \{B = no, b = yes\}    \\*
                                                      && \verb+{char}+ & allow duplicates  \{M = no, m = yes\}      \\
    \midrule
    Inertia & \verb+10x8+ &   \verb+{int}x{int}+ &  grid width $\times$ grid height & $0.2 \cdot w^2 \cdot h^2$\\
    \midrule
    Keen & \verb+6dn+ &  \verb+{int}+ & grid size     & $(2 \cdot w + 1) \cdot w^2$   \\*
                                                      && \verb+d{char}+ & difficulty \{e = easy, n = normal, h = hard, \\*&&&
                                                                                           x = extreme, u = unreasonable\}       \\*
                                                      && \verb+{char}+ & (Optional) multiplication only \{m = yes\}       \\
    \midrule
    Light Up & \verb+7x7b20s4d0+ &  \verb+{int}x{int}+ &  grid width $\times$ grid height & $\frac{1}{2} \cdot (w + h + 1)$  \\*
                                                      && \verb+b{int}+ & percentage of black squares  & $\cdot (w \cdot h + 1)$     \\*
                                                      && \verb+s{int}+ & symmetry \{0 = none, 1 = 2-way mirror, \\*&&& 2 = 2-way rotational, 3 = 4-way mirror, \\*&&& 4 = 4-way rotational\}       \\*
                                                      && \verb+d{int}+ & difficulty \{0 = easy, 1 = tricky, 2 = hard\}       \\
    \midrule
    Loopy & \verb+10x10t12dh+ & \verb+{int}x{int}+ &  grid width $\times$ grid height & $(2 \cdot w \cdot h + 1) \cdot 3 \cdot (w\cdot h)^2$\\*
                                                      && \verb+t{int}+ & type \{0 = squares, 1 = triangular, \\*
                                                      &&& 2 = honeycomb, 3 = snub-square, \\*
                                                      &&& 4 = cairo, 5 = great-hexagonal, \\*
                                                      &&& 6 = octagonal, 7 = kites, \\*
                                                      &&& 8 = floret, 9 = dodecagonal, \\*
                                                      &&& 10 = great-dodecagonal, \\*
                                                      &&& 11 = Penrose (kite/dart), \\*
                                                      &&& 12 = Penrose (rhombs),  \\*
                                                      &&& 13 = great-great-dodecagonal, \\*
                                                      &&& 14 = kagome, 15 = compass-dodecagonal, \\*
                                                      &&& 16 = hats\}       \\*
                                                      && \verb+d{char}+ & difficulty \{e = easy, n = normal, \\*&&& t = tricky, h = hard\}       \\*
    \midrule
    Magnets & \verb+6x5dtS+ &  \verb+{int}x{int}+ &  grid width $\times$ grid height & $w \cdot h \cdot (w + h + 2)$\\*
                                                      && \verb+d{char}+ & difficulty \{e = easy, t = tricky       \\*
                                                      && \verb+{char}+ & (Optional) strip clues \{S = yes\}       \\
    \midrule
    Map & \verb+20x15n30dn+ &  \verb+{int}x{int}+ &  grid width $\times$ grid height  & $2 \cdot n \cdot (1 + w + h)$\\*
                                                      && \verb+n{int}+ & number of regions       \\
                                                      && \verb+d{char}+ & difficulty \{e = easy, n = normal, h = hard, \\*&&&u = unreasonable\}       \\
    \midrule
    Mines & \verb+9x9n10+ &  \verb+{int}x{int}+ &  grid width $\times$ grid height & $w \cdot h \cdot (w + h + 1)$\\*
                                                      && \verb+n{int}+ & number of mines       \\
                                                      && \verb+p{char}+ & (Optional) ensure solubility \{a = no\}       \\
    \midrule
    Mosaic & \verb+10x10h0+ &  \verb+{int}x{int}+ &  grid width $\times$ grid height & $w \cdot h \cdot (w + h + 1)$\\*
                                                      && \verb+{str}+ & (Optional) aggressive generation  \{h0 = no\}     \\
    \midrule
    Net & \verb+5x5wb0.5+ & \verb+{int}x{int}+ &  grid width $\times$ grid height & $w \cdot h \cdot (w + h + 3)$\\*
                                                      && \verb+{char}+ & (Optional) walls wrap around  \{w = yes\}     \\
                                                      && \verb+b{float}+ & barrier probability, interval: [0, 1]   \\
                                                      && \verb+{char}+ & (Optional) ensure unique solution  \{a = no\}      \\
    \midrule
    Netslide & \verb+4x4wb1m2+ &  \verb+{int}x{int}+ &  grid width $\times$ grid height & $2 \cdot w \cdot h \cdot (w + h - 1)$ \\*
                                                      && \verb+{char}+ & (Optional) walls wrap around  \{w = yes\}     \\
                                                      && \verb+b{float}+ & barrier probability, interval: [0, 1]   \\
                                                      && \verb+m{int}+ & (Optional) number of shuffling moves      \\
    \midrule
    Palisade & \verb+5x5n5+ &  \verb+{int}x{int}+ &  grid width $\times$ grid height & $(2 \cdot w \cdot h - w - h)$\\*
                                                      && \verb+n{int}+ & region size     & $\cdot (w + h + 3)$\\
    \midrule
    Pattern & \verb+15x15+ & \verb+{int}x{int}+ &  grid width $\times$ grid height &  $w \cdot h (w + h + 1)$   \\
    \midrule
    Pearl & \verb+8x8dtn+ & \verb+{int}x{int}+ &  grid width $\times$ grid height & $w \cdot h \cdot (w + h + 2)$\\*
                                                      && \verb+d{char}+ & difficulty \{e = easy, t = tricky\}  \\*
                                                      && \verb+{char}+ & allow unsoluble \{n = yes\}     \\
    \midrule
    Pegs & \verb+7x7cross+ &  \verb+{int}x{int}+ &  grid width $\times$ grid height &  $w \cdot h \cdot (w + h + 2)$\\*
                                                      && \verb+{str}+ & type \{cross, octagon, random\}  \\
    \midrule
    Range & \verb+9x6+ &  \verb+{int}x{int}+ &  grid width $\times$ grid height & $w \cdot h \cdot (w + h + 1)$\\
    \midrule
    Rectangles & \verb+7x7e4+ &  \verb+{int}x{int}+ &  grid width $\times$ grid height & $2 \cdot w \cdot h \cdot (w + h + 1)$\\*
                                                      && \verb+e{int}+ & expansion factor  \\*
                                                      && \verb+{char}+ & ensure unique solution \{a = no\}     \\
    \midrule
    Same Game & \verb+5x5c3s2+ &  \verb+{int}x{int}+ &  grid width $\times$ grid height & $w \cdot h \cdot (w + h + 2)$ \\*
                                                      && \verb+c{int}+ & number of colors  \\*
                                                      && \verb+s{int}+ & scoring system \{1 = $(n-1)^2$,\\*&&& 2 = $(n-2)^2$\}     \\*
                                                      && \verb+{char}+ & (Optional) ensure solubility \{r = no\}     \\
    \midrule
    Signpost & \verb+4x4c+ &  \verb+{int}x{int}+ &  grid width $\times$ grid height & $2 \cdot w \cdot h \cdot (w + h + 1)$\\*
                                                      && \verb+{char}+ & (Optional) start and end in corners\\*&&& \{c = yes\}  \\
    \midrule
    Singles & \verb+5x5de+ & \verb+{int}x{int}+ &  grid width $\times$ grid height & $w \cdot h \cdot (w + h + 1)$\\*
                                                      && \verb+d{char}+ & difficulty \{e = easy, k = tricky\}  \\
    \midrule
    Sixteen & \verb+5x5m2+ &  \verb+{int}x{int}+ &  grid width $\times$ grid height & $w \cdot h \cdot (w + h + 3)$ \\*
                                                      && \verb+m{int}+ & (Optional) number of shuffling moves  \\
    \midrule
    Slant & \verb+8x8de+ &  \verb+{int}x{int}+ &  grid width $\times$ grid height & $w \cdot h \cdot (w + h + 1)$ \\*
                                                      && \verb+d{char}+ & difficulty \{e = easy, h = hard\}  \\
    \midrule
    Solo & \verb+3x3+ &  \verb+{int}x{int}+ &  rows of sub-blocks $\times$ cols of sub-blocks & $(w \cdot h)^2 * (2 \cdot w \cdot h + 1)$\\*
                                                      && \verb+{char}+ & (Optional) require every digit on each \\*&&& main diagonal \{x = yes\} \\**
                                                      && \verb+{char}+ & (Optional) jigsaw (irregularly shaped \\*&&& sub-blocks)  main diagonal \{j = yes\} \\**
                                                      && \verb+{char}+ & (Optional) killer (digit sums) \{k = yes\} \\**
                                                      && \verb+{str}+ & (Optional) symmetry. If not set, \\*&&&
                                                                                     it is 2-way rotation. \{a = None, \\*&&& 
                                                                                      m2 = 2-way mirror, m4 = 4-way mirror,  \\*&&&
                                                                                      r4 = 4-way rotation, m8 = 8-way mirror, \\*&&&
                                                                                      md2 = 2-way diagonal mirror,  \\*&&&
                                                                                      md4 = 4-way diagonal mirror\} \\*
                                                      && \verb+d{char}+ & difficulty \{t = trivial, b = basic, \\*&&&
                                                                                            i = intermediate, a = advanced, \\*&&&
                                                                                            e = extreme, u = unreasonable\}       \\
    \midrule
    Tents & \verb+8x8de+ &  \verb+{int}x{int}+ &  grid width $\times$ grid height & $\frac{1}{4} \cdot (w + 1) \cdot (h + 1) $\\*
                                                      && \verb+d{char}+ & difficulty \{e = easy, t = tricky\}  & $\cdot (w + h + 1)$\\
    \midrule
    Towers & \verb+5de+ &  \verb+{int}+ &  grid size & $2 \cdot (w + 1) \cdot w^2$\\*
                                                      && \verb+d{char}+ & difficulty \{e = easy, h = hard\\*&&&
                                                      x = extreme, u = unreasonable\}  \\
    \midrule
    Tracks & \verb+8x8dto+ &  \verb+{int}x{int}+ &  grid width $\times$ grid height & $w \cdot h (2 \cdot (w + h) + 1)$ \\*
                                                      && \verb+d{char}+ & difficulty \{e = easy, t = tricky, h = hard\}  \\*
                                                      && \verb+{char}+ & (Optional) disallow consecutive 1 clues\\*&&&
                                                                                         \{o = no\}  \\
    \midrule
    Twiddle & \verb+3x3n2+ &  \verb+{int}x{int}+ &  grid width $\times$ grid height & $(2 \cdot w \cdot h \cdot n^2 + 1) $\\*
                                                      && \verb+n{int}+ & rotating block size & $\cdot (w + h - 2\cdot n + 1)$\\*
                                                      && \verb+{char}+ & (Optional) one number per row \{r = yes\}  \\
                                                      && \verb+{char}+ & (Optional) orientation matters \{o = yes\}  \\
                                                      && \verb+m{int}+ & (Optional) number of shuffling moves \\
    \midrule
    Undead & \verb+4x4dn+ &  \verb+{int}x{int}+ &  grid width $\times$ grid height & $w \cdot h \cdot (w + h + 1)$\\*
                                                      && \verb+d{char}+ & difficulty \{e = easy, n = normal, t = tricky\}  \\
    \midrule
    Unequal & \verb+4adk+ &  \verb+{int}+ &  grid size & $w^2 \cdot (2 \cdot w + 1)$\\*
                                                      && \verb+{char}+ & (Optional) adjacent mode \{a = yes\}  \\*
                                                      && \verb+d{char}+ & difficulty \{t = trivial, e = easy, k = tricky,\\*&&& 
                                                                                           x = extreme, r = recursive\}  \\
    \midrule
    Unruly & \verb+8x8dt+ &  \verb+{int}+ &  grid size & $w \cdot h \cdot (w + h + 1)$\\*
                                                      && \verb+{char}+ & (Optional) unique rows and cols \{u = yes\}  \\*
                                                      && \verb+d{char}+ & difficulty \{t = trivial, e = easy, n = normal\}  \\
    \midrule
    Untangle & \verb+25+ &  \verb+{int}+ &  number of points & $n \cdot (n + \sqrt{3n} \cdot 4 + 2)$\\
\end{longtable}
}
\FloatBarrier
\newpage
\subsection{Baseline Parameters}\label{sec:experiment-parameters}
     In \cref{tab:experiment-parameters}, the parameters used for training the agents used for the comparisons in \cref{sec:baselines} is shown. 

 \begin{table}[h]
  \caption{
  Listed below are the generation parameters supplied to each puzzle instance before training an agent, as well as some puzzle-specific notes.
  We propose the easiest preset difficulty setting as a first challenge for RL algorithms to reach human-level performance.
  }
  \label{tab:experiment-parameters}
  \centering
  \scriptsize
    \begin{tabular}{llll}
    \toprule
    \textbf{Puzzle} & \textbf{Supplied Parameters} & \textbf{Easiest Human Level Preset} & \textbf{Notes} \\
    \midrule
        Black Box & \texttt{w2h2m2M2} & \texttt{w5h5m3M3} & \\
        Bridges & \texttt{3x3} & \texttt{7x7i30e10m2d0} & \\
        Cube & \texttt{c3x3} & \texttt{c4x4} & \\
        Dominosa & \texttt{1dt} & \texttt{3dt} & \\
        Fifteen & \texttt{2x2} & \texttt{4x4} & \\
        Filling & \texttt{2x3} & \texttt{9x7} & \\
        Flip & \texttt{3x3c} & \texttt{3x3c} & \\
        Flood & \texttt{3x3c6m5} & \texttt{12x12c6m5} & \\
        Galaxies & \texttt{3x3de} & \texttt{7x7dn} & \\
        Guess & \texttt{c2p3g10Bm} & \texttt{c6p4g10Bm} & Episodes were terminated and negatively rewarded \\
        &&& after the maximum number of guesses was made\\
        &&& without finding the correct solution. \\
        Inertia & \texttt{4x4} & \texttt{10x8} & \\
        Keen & \texttt{3dem} & \texttt{4de} &  Even the minimum allowed problem size\\
        &&& proved to be infeasible for a random agent\\
        Light Up & \texttt{3x3b20s0d0} & \texttt{7x7b20s4d0} & \\
        Loopy & \texttt{3x3t0de} & \texttt{3x3t0de} & \\
        Magnets & \texttt{3x3deS} & \texttt{6x5de} & \\
        Map & \texttt{3x3n5de} & \texttt{20x15n30de} & \\
        Mines & \texttt{4x4n2} & \texttt{9x9n10} & \\
        Mosaic & \texttt{3x3} & \texttt{3x3} & \\
        Net & \texttt{2x2} & \texttt{5x5} & \\
        Netslide & \texttt{2x3b1} & \texttt{3x3b1} & \\
        Palisade & \texttt{2x3n3} & \texttt{5x5n5} & \\
        Pattern & \texttt{3x2} & \texttt{10x10} & \\
        Pearl & \texttt{5x5de} & \texttt{6x6de} & \\
        Pegs & \texttt{4x4random} & \texttt{5x7cross} & \\
        Range & \texttt{3x3} & \texttt{9x6} & \\
        Rectangles & \texttt{3x2} & \texttt{7x7} & \\
        Same Game & \texttt{2x3c3s2} & \texttt{5x5c3s2} & \\
        Signpost & \texttt{2x3} & \texttt{4x4c} & \\
        Singles & \texttt{2x3de} & \texttt{5x5de} & \\
        Sixteen & \texttt{2x3} & \texttt{3x3} & \\
        Slant & \texttt{2x2de} & \texttt{5x5de} & \\
        Solo & \texttt{2x2} & \texttt{2x2} & \\
        Tents & \texttt{4x4de} & \texttt{8x8de} & \\
        Towers & \texttt{3de} & \texttt{4de} & \\
        Tracks & \texttt{4x4de} & \texttt{8x8de} & \\
        Twiddle & \texttt{2x3n2} & \texttt{3x3n2r} & \\
        Undead & \texttt{3x3de} & \texttt{4x4de} & \\
        Unequal & \texttt{3de} & \texttt{4de} & \\
        Unruly & \texttt{6x6dt} & \texttt{8x8dt} & Even the minimum allowed problem size\\
        &&& proved to be infeasible for a random agent\\
        Untangle & \texttt{4} & \texttt{6} & \\
    \bottomrule\\
    \end{tabular}
\end{table}

\FloatBarrier
\newpage

 \subsection{Detailed Baseline Results}\label{sec:experiment-results}
    We summarize all evaluated algorithms in \cref{table:rl-algorithms}.

            \begin{table*}[h]
        \small
        \caption{Summary of all evaluated RL algorithms.}
        \label{table:rl-algorithms}
            \centering
            \begin{tabular}{lcc}
                \toprule
                Algorithm & Policy Type & Action Masking\\
                \midrule
                Proximal Policy Optimization (PPO) \cite{schulman2017ppo} & On-Policy & No \\
                Recurrent PPO \citep{shengyi2022the37implementation} & On-Policy & No \\
                Advantage Actor Critic (A2C) \citep{mnih2016a3c} & On-Policy & No \\
                Asynchronous Advantage Actor Critic (A3C) \citep{mnih2016a3c} & On-Policy & No \\
                Trust Region Policy Optimization (TRPO) \citep{schulman2015trpo} & On-Policy & No \\
                Deep Q-Network (DQN) \citep{mnih2013atari} & Off-Policy & No \\
                Quantile Regression DQN (QRDQN) \citep{dabney2017qrdqn} & Off-Policy & No \\
                MuZero \citep{schrittwieser2020mastering} & Off-Policy & Yes \\
                DreamerV3 \citep{hafner2023mastering} & Off-Policy & No\\
                \bottomrule
            \end{tabular}
        \end{table*}
     As we limited the agents to a single final reward upon completion, where possible, we chose puzzle parameters that allowed random policies to successfully find a solution. 
     Note that if a random policy fails to find a solution, an RL algorithm without guidance (such as intermediate rewards) will also be affected by this.
     If an agent has never accumulated a reward with the initial (random) policy, it will be unable to improve its performance at all.
     
    The chosen parameters roughly corresponded to the smallest and easiest puzzles, as more complex puzzles were found to be intractable. 
    This fact is highlighted for example in \textit{Solo/Sudoku}, where the reasoning needed to find a valid solution is already rather complex, even for a grid with 2$ \times $2 sub-blocks.
    A few puzzles were still intractable due to the minimum complexity permitted by Tathams's puzzle-specific problem generators, such as with \textit{Unruly}. 
    
    For the RGB pixel observations, the window size chosen for these small problems was set at 128$ \times $128 pixels.

\begin{table}[h]
  \caption{Listed below are the detailed results for all evaluated algorithms. 
  Results show the average number of steps required for all successful episodes and standard deviation with respect to the random seeds.
  In brackets, we show the overall percentage of successful episodes.
  In the summary row, the last number in brackets denotes the total number of puzzles where a solution below the upper bound of optimal steps was found.
  Entries without values mean that no successful policy was found among all random seeds.
  This Table is continued in \cref{tab:phat_table_state_cont}.}
  \label{tab:phat_table_state}
  \centering
  \begin{adjustbox}{angle=90}
  \scriptsize
\begin{tabular}{lllllllllllll}
    \toprule
    \textbf{Puzzle} & \textbf{Supplied Parameters} & \textbf{Optimal} & \textbf{Random} & \textbf{PPO} & \textbf{TRPO} & \textbf{DreamerV3} & \textbf{MuZero} \\
    \midrule
    Blackbox & \texttt{w2h2m2M2} & $ 144$  & $ 2206 $ \hfill $ (99.2\%)$ & $1773 \pm 472 $ \hfill $ (59.5\%)$ & $1744 \pm 454 $ \hfill $ (96.3\%)$ & $\mathbf{32 \pm 5 } $ \hfill $ (100.0\%)$ & $\mathbf{46 \pm 0 } $ \hfill $ (0.1\%)$ \\
    Bridges & \texttt{3x3} & $ 378$  & $ 547 $ \hfill $ (100.0\%)$ & $682 \pm 197 $ \hfill $ (85.1\%)$ & $546 \pm 13 $ \hfill $ (100.0\%)$ & $\mathbf{9 \pm 0 } $ \hfill $ (100.0\%)$ & $397 \pm 181 $ \hfill $ (86.7\%)$ \\
    Cube & \texttt{c3x3} & $ 54$  & $ 4181 $ \hfill $ (66.9\%)$ & $744 \pm 1610 $ \hfill $ (77.5\%)$ & $433 \pm 917 $ \hfill $ (99.8\%)$ & $5068 \pm 657 $ \hfill $ (22.5\%)$ & - \\
    Dominosa & \texttt{1dt} & $ 32$  & $ 1980 $ \hfill $ (99.2\%)$ & $457 \pm 954 $ \hfill $ (70.0\%)$ & $\mathbf{12 \pm 1 } $ \hfill $ (100.0\%)$ & $\mathbf{11 \pm 1 } $ \hfill $ (100.0\%)$ & $3659 \pm 0 $ \hfill $ (0.0\%)$ \\
    Fifteen & \texttt{2x2} & $ 256$  & $ 54 $ \hfill $ (100.0\%)$ & $\mathbf{3 \pm 0 } $ \hfill $ (100.0\%)$ & $\mathbf{3 \pm 0 } $ \hfill $ (100.0\%)$ & $\mathbf{4 \pm 0 } $ \hfill $ (100.0\%)$ & $\mathbf{5 \pm 1 } $ \hfill $ (100.0\%)$ \\
    Filling & \texttt{2x3} & $ 36$  & $ 820 $ \hfill $ (100.0\%)$ & $290 \pm 249 $ \hfill $ (97.5\%)$ & $\mathbf{9 \pm 2 } $ \hfill $ (100.0\%)$ & $443 \pm 56 $ \hfill $ (83.4\%)$ & $1099 \pm 626 $ \hfill $ (15.0\%)$ \\
    Flip & \texttt{3x3c} & $ 63$  & $ 3138 $ \hfill $ (88.9\%)$ & $3008 \pm 837 $ \hfill $ (40.1\%)$ & $2951 \pm 564 $ \hfill $ (90.8\%)$ & $1762 \pm 568 $ \hfill $ (8.0\%)$ & $1207 \pm 1305 $ \hfill $ (3.1\%)$ \\
    Flood & \texttt{3x3c6m5} & $ 63$  & $ 134 $ \hfill $ (97.4\%)$ & $\mathbf{12 \pm 0 } $ \hfill $ (99.9\%)$ & $\mathbf{21 \pm 4 } $ \hfill $ (99.6\%)$ & $\mathbf{14 \pm 1 } $ \hfill $ (100.0\%)$ & $994 \pm 472 $ \hfill $ (14.4\%)$ \\
    Galaxies & \texttt{3x3de} & $ 156$  & $ 4306 $ \hfill $ (33.9\%)$ & $3860 \pm 1778 $ \hfill $ (8.3\%)$ & $4755 \pm 527 $ \hfill $ (24.8\%)$ & $3367 \pm 1585 $ \hfill $ (11.0\%)$ & $6046 \pm 2722 $ \hfill $ (8.2\%)$ \\
    Guess & \texttt{c2p3g10Bm} & $ 200$  & $ 358 $ \hfill $ (73.4\%)$ & - & $316 \pm 52 $ \hfill $ (72.0\%)$ & $268 \pm 226 $ \hfill $ (77.0\%)$ & $\mathbf{24 \pm 0 } $ \hfill $ (0.8\%)$ \\
    Inertia & \texttt{4x4} & $ 51$  & $ 13 $ \hfill $ (6.5\%)$ & $\mathbf{22 \pm 9 } $ \hfill $ (6.3\%)$ & $635 \pm 1373 $ \hfill $ (5.7\%)$ & $926 \pm 217 $ \hfill $ (5.7\%)$ & $104 \pm 73 $ \hfill $ (3.1\%)$ \\
    Keen & \texttt{3dem} & $ 63$  & $ 3152 $ \hfill $ (0.5\%)$ & $3817 \pm 0 $ \hfill $ (0.2\%)$ & $5887 \pm 1526 $ \hfill $ (0.4\%)$ & $4350 \pm 1163 $ \hfill $ (1.3\%)$ & - \\
    Lightup & \texttt{3x3b20s0d0} & $ 35$  & $ 2237 $ \hfill $ (98.1\%)$ & $1522 \pm 1115 $ \hfill $ (82.7\%)$ & $2127 \pm 168 $ \hfill $ (95.8\%)$ & $438 \pm 247 $ \hfill $ (72.0\%)$ & $1178 \pm 1109 $ \hfill $ (2.1\%)$ \\
    Loopy & \texttt{3x3t0de} & $ 4617$  & - & - & - & - & - \\
    Magnets & \texttt{3x3deS} & $ 72$  & $ 1895 $ \hfill $ (99.1\%)$ & $1366 \pm 1090 $ \hfill $ (90.2\%)$ & $1912 \pm 60 $ \hfill $ (99.1\%)$ & $574 \pm 56 $ \hfill $ (78.5\%)$ & $1491 \pm 0 $ \hfill $ (0.7\%)$ \\
    Map & \texttt{3x3n5de} & $ 70$  & $ 903 $ \hfill $ (99.9\%)$ & $1172 \pm 297 $ \hfill $ (75.7\%)$ & $950 \pm 34 $ \hfill $ (99.9\%)$ & $1680 \pm 197 $ \hfill $ (64.9\%)$ & $467 \pm 328 $ \hfill $ (0.9\%)$ \\
    Mines & \texttt{4x4n2} & $ 144$  & $ 87 $ \hfill $ (18.1\%)$ & $2478 \pm 2424 $ \hfill $ (9.9\%)$ & $\mathbf{123 \pm 66 } $ \hfill $ (18.8\%)$ & $272 \pm 246 $ \hfill $ (50.1\%)$ & $\mathbf{19 \pm 22 } $ \hfill $ (4.6\%)$ \\
    Mosaic & \texttt{3x3} & $ 63$  & $ 4996 $ \hfill $ (9.8\%)$ & $4928 \pm 438 $ \hfill $ (2.5\%)$ & $5233 \pm 615 $ \hfill $ (5.0\%)$ & $4469 \pm 387 $ \hfill $ (15.9\%)$ & $5586 \pm 0 $ \hfill $ (0.2\%)$ \\
    Net & \texttt{2x2} & $ 28$  & $ 1279 $ \hfill $ (100.0\%)$ & $\mathbf{9 \pm 0 } $ \hfill $ (100.0\%)$ & $\mathbf{9 \pm 0 } $ \hfill $ (100.0\%)$ & $\mathbf{10 \pm 0 } $ \hfill $ (100.0\%)$ & $339 \pm 448 $ \hfill $ (8.2\%)$ \\
    Netslide & \texttt{2x3b1} & $ 48$  & $ 766 $ \hfill $ (100.0\%)$ & $1612 \pm 1229 $ \hfill $ (41.6\%)$ & $635 \pm 145 $ \hfill $ (100.0\%)$ & $\mathbf{12 \pm 0 } $ \hfill $ (100.0\%)$ & $683 \pm 810 $ \hfill $ (25.0\%)$ \\
    Netslide & \texttt{3x3b1} & $ 90$  & $ 4671 $ \hfill $ (11.0\%)$ & $4671 \pm 498 $ \hfill $ (9.2\%)$ & $4008 \pm 1214 $ \hfill $ (8.9\%)$ & $3586 \pm 677 $ \hfill $ (22.4\%)$ & $3721 \pm 1461 $ \hfill $ (13.2\%)$ \\
    Palisade & \texttt{2x3n3} & $ 56$  & $ 1428 $ \hfill $ (100.0\%)$ & $939 \pm 604 $ \hfill $ (87.0\%)$ & $1377 \pm 35 $ \hfill $ (99.9\%)$ & $\mathbf{39 \pm 56 } $ \hfill $ (100.0\%)$ & $86 \pm 0 $ \hfill $ (0.0\%)$ \\
    Pattern & \texttt{3x2} & $ 36$  & $ 3247 $ \hfill $ (92.9\%)$ & $1542 \pm 1262 $ \hfill $ (71.9\%)$ & $2908 \pm 355 $ \hfill $ (90.2\%)$ & $820 \pm 516 $ \hfill $ (58.0\%)$ & $4063 \pm 1696 $ \hfill $ (1.9\%)$ \\
    Pearl & \texttt{5x5de} & $ 300$  & - & - & - & - & - \\
    Pegs & \texttt{4x4Random} & $ 160$  & - & - & - & - & - \\
    Range & \texttt{3x3} & $ 63$  & $ 535 $ \hfill $ (100.0\%)$ & $780 \pm 305 $ \hfill $ (65.8\%)$ & $661 \pm 198 $ \hfill $ (99.9\%)$ & $888 \pm 238 $ \hfill $ (55.6\%)$ & $91 \pm 76 $ \hfill $ (5.1\%)$ \\
    Rect & \texttt{3x2} & $ 72$  & $ 723 $ \hfill $ (100.0\%)$ & $\mathbf{27 \pm 44 } $ \hfill $ (99.8\%)$ & $\mathbf{9 \pm 4 } $ \hfill $ (100.0\%)$ & $\mathbf{8 \pm 1 } $ \hfill $ (100.0\%)$ & - \\
    Samegame & \texttt{2x3c3s2} & $ 42$  & $ 76 $ \hfill $ (100.0\%)$ & $123 \pm 197 $ \hfill $ (98.8\%)$ & $\mathbf{7 \pm 0 } $ \hfill $ (100.0\%)$ & $\mathbf{7 \pm 0 } $ \hfill $ (100.0\%)$ & $1444 \pm 541 $ \hfill $ (28.7\%)$ \\
    Samegame & \texttt{5x5c3s2} & $ 300$  & $ 571 $ \hfill $ (32.1\%)$ & $1003 \pm 827 $ \hfill $ (30.5\%)$ & $672 \pm 160 $ \hfill $ (30.8\%)$ & $527 \pm 162 $ \hfill $ (30.2\%)$ & $\mathbf{184 \pm 107 } $ \hfill $ (4.9\%)$ \\
    Signpost & \texttt{2x3} & $ 72$  & $ 776 $ \hfill $ (96.1\%)$ & $838 \pm 53 $ \hfill $ (97.2\%)$ & $799 \pm 13 $ \hfill $ (97.0\%)$ & $859 \pm 304 $ \hfill $ (91.3\%)$ & $4883 \pm 1285 $ \hfill $ (5.9\%)$ \\
    Singles & \texttt{2x3de} & $ 36$  & $ 353 $ \hfill $ (100.0\%)$ & $\mathbf{7 \pm 3 } $ \hfill $ (100.0\%)$ & $\mathbf{7 \pm 4 } $ \hfill $ (100.0\%)$ & $\mathbf{11 \pm 8 } $ \hfill $ (99.9\%)$ & $733 \pm 551 $ \hfill $ (28.4\%)$ \\
    Sixteen & \texttt{2x3} & $ 48$  & $ 2908 $ \hfill $ (94.1\%)$ & $2371 \pm 1226 $ \hfill $ (55.7\%)$ & $2968 \pm 181 $ \hfill $ (92.8\%)$ & $\mathbf{17 \pm 1 } $ \hfill $ (100.0\%)$ & $3281 \pm 472 $ \hfill $ (68.7\%)$ \\
    Slant & \texttt{2x2de} & $ 20$  & $ 447 $ \hfill $ (100.0\%)$ & $333 \pm 190 $ \hfill $ (80.4\%)$ & $21 \pm 2 $ \hfill $ (99.9\%)$ & $596 \pm 163 $ \hfill $ (100.0\%)$ & $1005 \pm 665 $ \hfill $ (7.4\%)$ \\
    Solo & \texttt{2x2} & $ 144$  & - & - & - & - & - \\
    Tents & \texttt{4x4de} & $ 56$  & $ 4442 $ \hfill $ (44.3\%)$ & $4781 \pm 86 $ \hfill $ (10.3\%)$ & $4828 \pm 752 $ \hfill $ (31.0\%)$ & $3137 \pm 581 $ \hfill $ (12.1\%)$ & $4556 \pm 3259 $ \hfill $ (0.6\%)$ \\
    Towers & \texttt{3de} & $ 72$  & $ 4876 $ \hfill $ (1.0\%)$ & - & $3789 \pm 1288 $ \hfill $ (0.5\%)$ & $3746 \pm 1861 $ \hfill $ (0.5\%)$ & - \\
    Tracks & \texttt{4x4de} & $ 272$  & $ 5213 $ \hfill $ (0.5\%)$ & $4129 \pm nan $ \hfill $ (0.1\%)$ & $5499 \pm 2268 $ \hfill $ (0.3\%)$ & $4483 \pm 1513 $ \hfill $ (0.3\%)$ & - \\
    Twiddle & \texttt{2x3n2} & $ 98$  & $ 851 $ \hfill $ (100.0\%)$ & $\mathbf{8 \pm 1 } $ \hfill $ (99.9\%)$ & $\mathbf{11 \pm 7 } $ \hfill $ (100.0\%)$ & $\mathbf{8 \pm 0 } $ \hfill $ (100.0\%)$ & $761 \pm 860 $ \hfill $ (37.6\%)$ \\
    Undead & \texttt{3x3de} & $ 63$  & $ 4390 $ \hfill $ (40.1\%)$ & $4542 \pm 292 $ \hfill $ (5.7\%)$ & $4179 \pm 299 $ \hfill $ (31.0\%)$ & $4088 \pm 297 $ \hfill $ (35.8\%)$ & $3677 \pm 342 $ \hfill $ (9.0\%)$ \\
    Unequal & \texttt{3de} & $ 63$  & $ 4540 $ \hfill $ (6.7\%)$ & - & $5105 \pm 193 $ \hfill $ (3.6\%)$ & $2468 \pm 2025 $ \hfill $ (4.8\%)$ & $4944 \pm 368 $ \hfill $ (7.2\%)$ \\
    Unruly & \texttt{6x6dt} & $ 468$  & - & - & - & - & - \\
    Untangle & \texttt{4} & $ 150$  & $ 141 $ \hfill $ (100.0\%)$ & $\mathbf{13 \pm 1 } $ \hfill $ (100.0\%)$ & $\mathbf{11 \pm 0 } $ \hfill $ (100.0\%)$ & $\mathbf{6 \pm 0 } $ \hfill $ (100.0\%)$ & $499 \pm 636 $ \hfill $ (26.5\%)$ \\
    Untangle & \texttt{6} & $ 79$  & $ 2165 $ \hfill $ (96.9\%)$ & $2295 \pm 66 $ \hfill $ (96.2\%)$ & $2228 \pm 126 $ \hfill $ (96.5\%)$ & $1683 \pm 74 $ \hfill $ (82.0\%)$ & $2380 \pm 0 $ \hfill $ (11.2\%)$ \\
    \midrule
    Summary & - & $ 217 $ & $ 1984 $ \hfill $(71.2\%)$ & $ 1604 \pm 801 $ \hfill $(61.6\%) (8)$ & $ 1773 \pm 639 $ \hfill $(70.8\%) (11)$ & $ 1334 \pm 654 $ \hfill $(62.7\%) (14)$ & $ 1808 \pm 983 $ \hfill $(16.0\%) (5)$ \\
    \bottomrule
\end{tabular}
  \end{adjustbox}
\end{table}

\begin{table}[h]
  \caption{
  Continuation from \cref{tab:phat_table_state}.
  Listed below are the detailed results for all evaluated algorithms. 
  Results show the average number of steps required for all successful episodes and standard deviation with respect to the random seeds.
  In brackets, we show the overall percentage of successful episodes.
  In the summary row, the last number in brackets denotes the total number of puzzles where a solution below the upper bound of optimal steps was found.
  Entries without values mean that no successful policy was found among all random seeds.
  }
  \label{tab:phat_table_state_cont}
  \centering
  \begin{adjustbox}{angle=90}
  \scriptsize
\begin{tabular}{lllllllllllll}
    \toprule
    \textbf{Puzzle} & \textbf{Supplied Parameters} & \textbf{Optimal} & \textbf{Random} & \textbf{A2C} & \textbf{RecurrentPPO} & \textbf{DQN} & \textbf{QRDQN} \\
    \midrule
    Blackbox & \texttt{w2h2m2M2} & $ 144$  & $ 2206 $ \hfill $ (99.2\%)$ & $2524 \pm 1193 $ \hfill $ (85.2\%)$ & $2009 \pm 427 $ \hfill $ (98.7\%)$ & $2063 \pm 70 $ \hfill $ (99.0\%)$ & $2984 \pm 1584 $ \hfill $ (76.8\%)$ \\
    Bridges & \texttt{3x3} & $ 378$  & $ 547 $ \hfill $ (100.0\%)$ & $540 \pm 69 $ \hfill $ (100.0\%)$ & $653 \pm 165 $ \hfill $ (100.0\%)$ & $549 \pm 20 $ \hfill $ (100.0\%)$ & $1504 \pm 2037 $ \hfill $ (83.4\%)$ \\
    Cube & \texttt{c3x3} & $ 54$  & $ 4181 $ \hfill $ (66.9\%)$ & $4516 \pm 954 $ \hfill $ (17.5\%)$ & $4943 \pm 620 $ \hfill $ (16.2\%)$ & $4407 \pm 414 $ \hfill $ (43.4\%)$ & $4241 \pm 283 $ \hfill $ (26.4\%)$ \\
    Dominosa & \texttt{1dt} & $ 32$  & $ 1980 $ \hfill $ (99.2\%)$ & $6408 \pm nan $ \hfill $ (0.2\%)$ & $3009 \pm 988 $ \hfill $ (80.6\%)$ & $\mathbf{15 \pm 6 } $ \hfill $ (100.0\%)$ & $4457 \pm 2183 $ \hfill $ (50.0\%)$ \\
    Fifteen & \texttt{2x2} & $ 256$  & $ 54 $ \hfill $ (100.0\%)$ & $\mathbf{4 \pm 1 } $ \hfill $ (100.0\%)$ & $\mathbf{3 \pm 0 } $ \hfill $ (100.0\%)$ & $\mathbf{3 \pm 0 } $ \hfill $ (100.0\%)$ & $\mathbf{3 \pm 0 } $ \hfill $ (100.0\%)$ \\
    Filling & \texttt{2x3} & $ 36$  & $ 820 $ \hfill $ (100.0\%)$ & $777 \pm 310 $ \hfill $ (99.3\%)$ & $764 \pm 106 $ \hfill $ (100.0\%)$ & $761 \pm 109 $ \hfill $ (99.7\%)$ & $2828 \pm 2769 $ \hfill $ (63.2\%)$ \\
    Flip & \texttt{3x3c} & $ 63$  & $ 3138 $ \hfill $ (88.9\%)$ & $4345 \pm 1928 $ \hfill $ (29.4\%)$ & $3356 \pm 1412 $ \hfill $ (46.9\%)$ & $3493 \pm 129 $ \hfill $ (87.1\%)$ & $3741 \pm 353 $ \hfill $ (56.8\%)$ \\
    Flood & \texttt{3x3c6m5} & $ 63$  & $ 134 $ \hfill $ (97.4\%)$ & $406 \pm 623 $ \hfill $ (93.4\%)$ & $120 \pm 17 $ \hfill $ (97.7\%)$ & $128 \pm 12 $ \hfill $ (90.8\%)$ & $1954 \pm 2309 $ \hfill $ (65.2\%)$ \\
    Galaxies & \texttt{3x3de} & $ 156$  & $ 4306 $ \hfill $ (33.9\%)$ & $4586 \pm 980 $ \hfill $ (10.8\%)$ & $3939 \pm 1438 $ \hfill $ (0.4\%)$ & $4657 \pm 147 $ \hfill $ (26.1\%)$ & - \\
    Guess & \texttt{c2p3g10Bm} & $ 200$  & $ 358 $ \hfill $ (73.4\%)$ & - & $323 \pm 52 $ \hfill $ (44.6\%)$ & $550 \pm 248 $ \hfill $ (71.9\%)$ & $3260 \pm 2614 $ \hfill $ (34.4\%)$ \\
    Inertia & \texttt{4x4} & $ 51$  & $ 13 $ \hfill $ (6.5\%)$ & $105 \pm 197 $ \hfill $ (6.1\%)$ & $1198 \pm 1482 $ \hfill $ (5.6\%)$ & $179 \pm 156 $ \hfill $ (7.1\%)$ & $1330 \pm 296 $ \hfill $ (5.8\%)$ \\
    Keen & \texttt{3dem} & $ 63$  & $ 3152 $ \hfill $ (0.5\%)$ & - & - & $6774 \pm 1046 $ \hfill $ (0.4\%)$ & - \\
    Lightup & \texttt{3x3b20s0d0} & $ 35$  & $ 2237 $ \hfill $ (98.1\%)$ & $3034 \pm 793 $ \hfill $ (62.7\%)$ & $3493 \pm 929 $ \hfill $ (66.5\%)$ & $2429 \pm 214 $ \hfill $ (97.5\%)$ & $3440 \pm 945 $ \hfill $ (57.8\%)$ \\
    Loopy & \texttt{3x3t0de} & $ 4617$  & - & - & - & - & - \\
    Magnets & \texttt{3x3deS} & $ 72$  & $ 1895 $ \hfill $ (99.1\%)$ & $3057 \pm 1114 $ \hfill $ (47.9\%)$ & $1874 \pm 222 $ \hfill $ (99.2\%)$ & $2112 \pm 331 $ \hfill $ (98.1\%)$ & $5182 \pm 3878 $ \hfill $ (33.8\%)$ \\
    Map & \texttt{3x3n5de} & $ 70$  & $ 903 $ \hfill $ (99.9\%)$ & $2552 \pm 1223 $ \hfill $ (52.5\%)$ & $2608 \pm 1808 $ \hfill $ (59.4\%)$ & $949 \pm 30 $ \hfill $ (99.9\%)$ & $1753 \pm 769 $ \hfill $ (78.1\%)$ \\
    Mines & \texttt{4x4n2} & $ 144$  & $ 87 $ \hfill $ (18.1\%)$ & $\mathbf{120 \pm 41 } $ \hfill $ (14.7\%)$ & $1189 \pm 1341 $ \hfill $ (12.1\%)$ & $207 \pm 146 $ \hfill $ (17.6\%)$ & $1576 \pm 1051 $ \hfill $ (13.2\%)$ \\
    Mosaic & \texttt{3x3} & $ 63$  & $ 4996 $ \hfill $ (9.8\%)$ & $4937 \pm 424 $ \hfill $ (8.4\%)$ & $4907 \pm 219 $ \hfill $ (8.3\%)$ & $5279 \pm 564 $ \hfill $ (7.0\%)$ & $9490 \pm 155 $ \hfill $ (0.0\%)$ \\
    Net & \texttt{2x2} & $ 28$  & $ 1279 $ \hfill $ (100.0\%)$ & $149 \pm 288 $ \hfill $ (100.0\%)$ & $1232 \pm 92 $ \hfill $ (100.0\%)$ & $\mathbf{9 \pm 0 } $ \hfill $ (100.0\%)$ & $1793 \pm 1663 $ \hfill $ (81.3\%)$ \\
    Netslide & \texttt{2x3b1} & $ 48$  & $ 766 $ \hfill $ (100.0\%)$ & $976 \pm 584 $ \hfill $ (100.0\%)$ & $2079 \pm 1989 $ \hfill $ (64.7\%)$ & $779 \pm 37 $ \hfill $ (100.0\%)$ & $1023 \pm 206 $ \hfill $ (80.9\%)$ \\
    Netslide & \texttt{3x3b1} & $ 90$  & $ 4671 $ \hfill $ (11.0\%)$ & $4324 \pm 657 $ \hfill $ (8.1\%)$ & $2737 \pm 1457 $ \hfill $ (1.7\%)$ & $4099 \pm 846 $ \hfill $ (5.1\%)$ & $2025 \pm 1475 $ \hfill $ (0.4\%)$ \\
    Palisade & \texttt{2x3n3} & $ 56$  & $ 1428 $ \hfill $ (100.0\%)$ & $1666 \pm 198 $ \hfill $ (99.4\%)$ & $1981 \pm 1053 $ \hfill $ (92.5\%)$ & $1445 \pm 96 $ \hfill $ (99.9\%)$ & $1519 \pm 142 $ \hfill $ (99.8\%)$ \\
    Pattern & \texttt{3x2} & $ 36$  & $ 3247 $ \hfill $ (92.9\%)$ & $3445 \pm 635 $ \hfill $ (82.9\%)$ & $3733 \pm 513 $ \hfill $ (79.7\%)$ & $2809 \pm 733 $ \hfill $ (89.7\%)$ & $3406 \pm 384 $ \hfill $ (51.1\%)$ \\
    Pearl & \texttt{5x5de} & $ 300$  & - & - & - & - & - \\
    Pegs & \texttt{4x4Random} & $ 160$  & - & - & - & - & - \\
    Range & \texttt{3x3} & $ 63$  & $ 535 $ \hfill $ (100.0\%)$ & $1438 \pm 782 $ \hfill $ (81.4\%)$ & $730 \pm 172 $ \hfill $ (99.9\%)$ & $594 \pm 28 $ \hfill $ (100.0\%)$ & - \\
    Rect & \texttt{3x2} & $ 72$  & $ 723 $ \hfill $ (100.0\%)$ & $3470 \pm 2521 $ \hfill $ (17.6\%)$ & $916 \pm 420 $ \hfill $ (99.6\%)$ & $511 \pm 193 $ \hfill $ (97.4\%)$ & $1560 \pm 1553 $ \hfill $ (81.8\%)$ \\
    Samegame & \texttt{2x3c3s2} & $ 42$  & $ 76 $ \hfill $ (100.0\%)$ & $\mathbf{8 \pm 1 } $ \hfill $ (100.0\%)$ & $1777 \pm 1643 $ \hfill $ (43.5\%)$ & $\mathbf{8 \pm 0 } $ \hfill $ (100.0\%)$ & $\mathbf{14 \pm 9 } $ \hfill $ (100.0\%)$ \\
    Samegame & \texttt{5x5c3s2} & $ 300$  & $ 571 $ \hfill $ (32.1\%)$ & $609 \pm 155 $ \hfill $ (29.9\%)$ & $1321 \pm 1170 $ \hfill $ (30.3\%)$ & $850 \pm 546 $ \hfill $ (29.2\%)$ & $5577 \pm 1211 $ \hfill $ (12.8\%)$ \\
    Signpost & \texttt{2x3} & $ 72$  & $ 776 $ \hfill $ (96.1\%)$ & $2259 \pm 1394 $ \hfill $ (85.9\%)$ & $1000 \pm 266 $ \hfill $ (77.9\%)$ & $793 \pm 17 $ \hfill $ (97.0\%)$ & $2298 \pm 2845 $ \hfill $ (78.0\%)$ \\
    Singles & \texttt{2x3de} & $ 36$  & $ 353 $ \hfill $ (100.0\%)$ & $372 \pm 47 $ \hfill $ (100.0\%)$ & $331 \pm 66 $ \hfill $ (100.0\%)$ & $361 \pm 47 $ \hfill $ (99.1\%)$ & $392 \pm 29 $ \hfill $ (100.0\%)$ \\
    Sixteen & \texttt{2x3} & $ 48$  & $ 2908 $ \hfill $ (94.1\%)$ & $3903 \pm 479 $ \hfill $ (71.7\%)$ & $3409 \pm 574 $ \hfill $ (67.6\%)$ & $2970 \pm 107 $ \hfill $ (93.2\%)$ & $4550 \pm 848 $ \hfill $ (21.9\%)$ \\
    Slant & \texttt{2x2de} & $ 20$  & $ 447 $ \hfill $ (100.0\%)$ & $984 \pm 470 $ \hfill $ (99.8\%)$ & $465 \pm 34 $ \hfill $ (100.0\%)$ & $496 \pm 97 $ \hfill $ (100.0\%)$ & $1398 \pm 2097 $ \hfill $ (87.1\%)$ \\
    Solo & \texttt{2x2} & $ 144$  & - & - & - & - & - \\
    Tents & \texttt{4x4de} & $ 56$  & $ 4442 $ \hfill $ (44.3\%)$ & $6157 \pm 1961 $ \hfill $ (2.1\%)$ & $4980 \pm 397 $ \hfill $ (12.8\%)$ & $4515 \pm 59 $ \hfill $ (38.1\%)$ & $5295 \pm 688 $ \hfill $ (7.8\%)$ \\
    Towers & \texttt{3de} & $ 72$  & $ 4876 $ \hfill $ (1.0\%)$ & $9850 \pm nan $ \hfill $ (0.0\%)$ & $8549 \pm nan $ \hfill $ (0.0\%)$ & $5836 \pm 776 $ \hfill $ (0.5\%)$ & - \\
    Tracks & \texttt{4x4de} & $ 272$  & $ 5213 $ \hfill $ (0.5\%)$ & $4501 \pm nan $ \hfill $ (0.0\%)$ & - & $5809 \pm 661 $ \hfill $ (0.3\%)$ & - \\
    Twiddle & \texttt{2x3n2} & $ 98$  & $ 851 $ \hfill $ (100.0\%)$ & $1248 \pm 430 $ \hfill $ (99.6\%)$ & $827 \pm 71 $ \hfill $ (100.0\%)$ & $\mathbf{83 \pm 149 } $ \hfill $ (100.0\%)$ & $3170 \pm 1479 $ \hfill $ (33.4\%)$ \\
    Undead & \texttt{3x3de} & $ 63$  & $ 4390 $ \hfill $ (40.1\%)$ & $5818 \pm 154 $ \hfill $ (0.9\%)$ & $5060 \pm 2381 $ \hfill $ (0.5\%)$ & - & - \\
    Unequal & \texttt{3de} & $ 63$  & $ 4540 $ \hfill $ (6.7\%)$ & $5067 \pm 1600 $ \hfill $ (1.0\%)$ & $5929 \pm 1741 $ \hfill $ (1.1\%)$ & $5057 \pm 582 $ \hfill $ (5.6\%)$ & - \\
    Unruly & \texttt{6x6dt} & $ 468$  & - & - & - & - & - \\
    Untangle & \texttt{4} & $ 150$  & $ 141 $ \hfill $ (100.0\%)$ & $1270 \pm 1745 $ \hfill $ (90.4\%)$ & $\mathbf{135 \pm 18 } $ \hfill $ (100.0\%)$ & $170 \pm 29 $ \hfill $ (100.0\%)$ & $871 \pm 837 $ \hfill $ (99.0\%)$ \\
    Untangle & \texttt{6} & $ 79$  & $ 2165 $ \hfill $ (96.9\%)$ & $3324 \pm 1165 $ \hfill $ (72.5\%)$ & $2739 \pm 588 $ \hfill $ (91.7\%)$ & $2219 \pm 84 $ \hfill $ (95.9\%)$ & - \\
    \midrule
    Summary & - & $ 217 $ & $ 1984 $ \hfill $(71.2\%)$ & $ 2743 \pm 954 $ \hfill $(54.8\%) (3)$ & $ 2342 \pm 989 $ \hfill $(61.1\%) (2)$ & $ 1999 \pm 365 $ \hfill $(70.2\%) (5)$ & $ 2754 \pm 1579 $ \hfill $(56.0\%) (2)$ \\
    \bottomrule
\end{tabular}
  \end{adjustbox}
\end{table}

\begin{table}[h]
  \caption{We list the detailed results for all the experiments of action masking and input representation. 
  Results show the average number of steps required for all successful episodes and standard deviation with respect to the random seeds.
  In brackets, we show the overall percentage of successful episodes.
  In the summary row, the last number in brackets denotes the total number of puzzles where a solution below the upper bound of optimal steps was found.
  Entries without values mean that no successful policy was found among all random seeds.
  }
  \label{tab:phat_table_masked}
  \centering
  \begin{adjustbox}{angle=90}
  \scriptsize
\begin{tabular}{lllllllllllll}
    \toprule
    \textbf{Puzzle} & \textbf{Supplied Parameters} & \textbf{Optimal} & \textbf{Random} & \textbf{PPO (Internal State)} & \textbf{PPO (RGB Pixels)} & \textbf{MaskablePPO (Internal State)} & \textbf{MaskablePPO (RGB Pixels)} \\
    \midrule
    Blackbox & \texttt{w2h2m2M2} & $ 144$  & $ 2206 $ \hfill $ (99.2\%)$ & $1773 \pm 472 $ \hfill $ (59.5\%)$ & $1509 \pm 792 $ \hfill $ (97.9\%)$ & $\mathbf{9 \pm 0 } $ \hfill $ (99.7\%)$ & $\mathbf{30 \pm 1 } $ \hfill $ (99.2\%)$ \\
    Bridges & \texttt{3x3} & $ 378$  & $ 547 $ \hfill $ (100.0\%)$ & $682 \pm 197 $ \hfill $ (85.1\%)$ & $\mathbf{89 \pm 176 } $ \hfill $ (99.1\%)$ & $\mathbf{25 \pm 0 } $ \hfill $ (99.4\%)$ & $\mathbf{9 \pm 0 } $ \hfill $ (99.6\%)$ \\
    Cube & \texttt{c3x3} & $ 54$  & $ 4181 $ \hfill $ (66.9\%)$ & $744 \pm 1610 $ \hfill $ (77.5\%)$ & $3977 \pm 442 $ \hfill $ (67.7\%)$ & $\mathbf{16 \pm 1 } $ \hfill $ (81.2\%)$ & $410 \pm 157 $ \hfill $ (75.1\%)$ \\
    Dominosa & \texttt{1dt} & $ 32$  & $ 1980 $ \hfill $ (99.2\%)$ & $457 \pm 954 $ \hfill $ (70.0\%)$ & $539 \pm 581 $ \hfill $ (100.0\%)$ & $\mathbf{12 \pm 0 } $ \hfill $ (100.0\%)$ & $\mathbf{19 \pm 2 } $ \hfill $ (100.0\%)$ \\
    Fifteen & \texttt{2x2} & $ 256$  & $ 54 $ \hfill $ (100.0\%)$ & $\mathbf{3 \pm 0 } $ \hfill $ (100.0\%)$ & $\mathbf{37 \pm 26 } $ \hfill $ (100.0\%)$ & $\mathbf{4 \pm 0 } $ \hfill $ (100.0\%)$ & $\mathbf{3 \pm 0 } $ \hfill $ (100.0\%)$ \\
    Filling & \texttt{2x3} & $ 36$  & $ 820 $ \hfill $ (100.0\%)$ & $290 \pm 249 $ \hfill $ (97.5\%)$ & $373 \pm 175 $ \hfill $ (99.9\%)$ & $\mathbf{7 \pm 0 } $ \hfill $ (100.0\%)$ & $\mathbf{34 \pm 3 } $ \hfill $ (99.9\%)$ \\
    Flip & \texttt{3x3c} & $ 63$  & $ 3138 $ \hfill $ (88.9\%)$ & $3008 \pm 837 $ \hfill $ (40.1\%)$ & $3616 \pm 395 $ \hfill $ (78.3\%)$ & $2174 \pm 1423 $ \hfill $ (70.3\%)$ & $319 \pm 128 $ \hfill $ (81.3\%)$ \\
    Flood & \texttt{3x3c6m5} & $ 63$  & $ 134 $ \hfill $ (97.4\%)$ & $\mathbf{12 \pm 0 } $ \hfill $ (99.9\%)$ & $\mathbf{28 \pm 12 } $ \hfill $ (99.7\%)$ & $\mathbf{12 \pm 0 } $ \hfill $ (99.9\%)$ & $\mathbf{14 \pm 0 } $ \hfill $ (99.9\%)$ \\
    Galaxies & \texttt{3x3de} & $ 156$  & $ 4306 $ \hfill $ (33.9\%)$ & $3860 \pm 1778 $ \hfill $ (8.3\%)$ & $4439 \pm 224 $ \hfill $ (29.1\%)$ & $3640 \pm 928 $ \hfill $ (40.2\%)$ & $3372 \pm 430 $ \hfill $ (40.5\%)$ \\
    Guess & \texttt{c2p3g10Bm} & $ 200$  & $ 358 $ \hfill $ (73.4\%)$ & - & $344 \pm 35 $ \hfill $ (72.0\%)$ & $\mathbf{145 \pm 19 } $ \hfill $ (75.4\%)$ & - \\
    Inertia & \texttt{4x4} & $ 51$  & $ 13 $ \hfill $ (6.5\%)$ & $\mathbf{22 \pm 9 } $ \hfill $ (6.3\%)$ & $237 \pm 10 $ \hfill $ (99.7\%)$ & $\mathbf{41 \pm 19 } $ \hfill $ (79.0\%)$ & $169 \pm 233 $ \hfill $ (69.8\%)$ \\
    Keen & \texttt{3dem} & $ 63$  & $ 3152 $ \hfill $ (0.5\%)$ & $3817 \pm 0 $ \hfill $ (0.2\%)$ & - & - & - \\
    Lightup & \texttt{3x3b20s0d0} & $ 35$  & $ 2237 $ \hfill $ (98.1\%)$ & $1522 \pm 1115 $ \hfill $ (82.7\%)$ & $2401 \pm 148 $ \hfill $ (97.5\%)$ & $\mathbf{25 \pm 8 } $ \hfill $ (99.1\%)$ & $1608 \pm 1144 $ \hfill $ (90.1\%)$ \\
    Loopy & \texttt{3x3t0de} & $ 4617$  & - & - & - & - & - \\
    Magnets & \texttt{3x3deS} & $ 72$  & $ 1895 $ \hfill $ (99.1\%)$ & $1366 \pm 1090 $ \hfill $ (90.2\%)$ & $1794 \pm 109 $ \hfill $ (98.7\%)$ & $222 \pm 33 $ \hfill $ (98.8\%)$ & $425 \pm 68 $ \hfill $ (99.2\%)$ \\
    Map & \texttt{3x3n5de} & $ 70$  & $ 903 $ \hfill $ (99.9\%)$ & $1172 \pm 297 $ \hfill $ (75.7\%)$ & $958 \pm 33 $ \hfill $ (99.9\%)$ & $321 \pm 33 $ \hfill $ (99.9\%)$ & $467 \pm 69 $ \hfill $ (99.1\%)$ \\
    Mines & \texttt{4x4n2} & $ 144$  & $ 87 $ \hfill $ (18.1\%)$ & $2478 \pm 2424 $ \hfill $ (9.9\%)$ & $2406 \pm 296 $ \hfill $ (44.7\%)$ & $412 \pm 268 $ \hfill $ (43.3\%)$ & $653 \pm 396 $ \hfill $ (43.1\%)$ \\
    Mosaic & \texttt{3x3} & $ 63$  & $ 4996 $ \hfill $ (9.8\%)$ & $4928 \pm 438 $ \hfill $ (2.5\%)$ & $5673 \pm 1547 $ \hfill $ (6.7\%)$ & $3381 \pm 906 $ \hfill $ (29.4\%)$ & $3158 \pm 247 $ \hfill $ (28.5\%)$ \\
    Net & \texttt{2x2} & $ 28$  & $ 1279 $ \hfill $ (100.0\%)$ & $\mathbf{9 \pm 0 } $ \hfill $ (100.0\%)$ & $180 \pm 44 $ \hfill $ (100.0\%)$ & $\mathbf{9 \pm 0 } $ \hfill $ (100.0\%)$ & - \\
    Netslide & \texttt{2x3b1} & $ 48$  & $ 766 $ \hfill $ (100.0\%)$ & $1612 \pm 1229 $ \hfill $ (41.6\%)$ & $\mathbf{35 \pm 18 } $ \hfill $ (100.0\%)$ & $\mathbf{13 \pm 0 } $ \hfill $ (100.0\%)$ & $96 \pm 7 $ \hfill $ (100.0\%)$ \\
    Netslide & \texttt{3x3b1} & $ 90$  & $ 4671 $ \hfill $ (11.0\%)$ & $4671 \pm 498 $ \hfill $ (9.2\%)$ & - & - & - \\
    Palisade & \texttt{2x3n3} & $ 56$  & $ 1428 $ \hfill $ (100.0\%)$ & $939 \pm 604 $ \hfill $ (87.0\%)$ & $1412 \pm 23 $ \hfill $ (99.9\%)$ & $90 \pm 55 $ \hfill $ (99.9\%)$ & $347 \pm 26 $ \hfill $ (99.8\%)$ \\
    Pattern & \texttt{3x2} & $ 36$  & $ 3247 $ \hfill $ (92.9\%)$ & $1542 \pm 1262 $ \hfill $ (71.9\%)$ & $2983 \pm 173 $ \hfill $ (92.5\%)$ & $\mathbf{14 \pm 0 } $ \hfill $ (96.9\%)$ & $1201 \pm 1021 $ \hfill $ (88.7\%)$ \\
    Pearl & \texttt{5x5de} & $ 300$  & - & - & - & - & - \\
    Pegs & \texttt{4x4Random} & $ 160$  & - & - & - & $1730 \pm 579 $ \hfill $ (34.9\%)$ & $1482 \pm 687 $ \hfill $ (37.3\%)$ \\
    Range & \texttt{3x3} & $ 63$  & $ 535 $ \hfill $ (100.0\%)$ & $780 \pm 305 $ \hfill $ (65.8\%)$ & $613 \pm 25 $ \hfill $ (100.0\%)$ & $\mathbf{50 \pm 69 } $ \hfill $ (100.0\%)$ & $209 \pm 26 $ \hfill $ (100.0\%)$ \\
    Rect & \texttt{3x2} & $ 72$  & $ 723 $ \hfill $ (100.0\%)$ & $\mathbf{27 \pm 44 } $ \hfill $ (99.8\%)$ & $300 \pm 387 $ \hfill $ (100.0\%)$ & $\mathbf{8 \pm 0 } $ \hfill $ (100.0\%)$ & $\mathbf{38 \pm 9 } $ \hfill $ (100.0\%)$ \\
    Samegame & \texttt{2x3c3s2} & $ 42$  & $ 76 $ \hfill $ (100.0\%)$ & $123 \pm 197 $ \hfill $ (98.8\%)$ & $\mathbf{11 \pm 8 } $ \hfill $ (100.0\%)$ & $\mathbf{8 \pm 0 } $ \hfill $ (100.0\%)$ & $\mathbf{9 \pm 0 } $ \hfill $ (100.0\%)$ \\
    Samegame & \texttt{5x5c3s2} & $ 300$  & $ 571 $ \hfill $ (32.1\%)$ & $1003 \pm 827 $ \hfill $ (30.5\%)$ & - & - & - \\
    Signpost & \texttt{2x3} & $ 72$  & $ 776 $ \hfill $ (96.1\%)$ & $838 \pm 53 $ \hfill $ (97.2\%)$ & $779 \pm 50 $ \hfill $ (97.0\%)$ & $567 \pm 149 $ \hfill $ (97.7\%)$ & $454 \pm 50 $ \hfill $ (97.5\%)$ \\
    Singles & \texttt{2x3de} & $ 36$  & $ 353 $ \hfill $ (100.0\%)$ & $\mathbf{7 \pm 3 } $ \hfill $ (100.0\%)$ & $306 \pm 57 $ \hfill $ (100.0\%)$ & $\mathbf{5 \pm 1 } $ \hfill $ (100.0\%)$ & $218 \pm 17 $ \hfill $ (100.0\%)$ \\
    Sixteen & \texttt{2x3} & $ 48$  & $ 2908 $ \hfill $ (94.1\%)$ & $2371 \pm 1226 $ \hfill $ (55.7\%)$ & $3211 \pm 450 $ \hfill $ (89.6\%)$ & $\mathbf{19 \pm 2 } $ \hfill $ (94.3\%)$ & $3650 \pm 190 $ \hfill $ (68.5\%)$ \\
    Slant & \texttt{2x2de} & $ 20$  & $ 447 $ \hfill $ (100.0\%)$ & $333 \pm 190 $ \hfill $ (80.4\%)$ & $325 \pm 119 $ \hfill $ (100.0\%)$ & $\mathbf{12 \pm 0 } $ \hfill $ (100.0\%)$ & $89 \pm 21 $ \hfill $ (100.0\%)$ \\
    Solo & \texttt{2x2} & $ 144$  & - & - & - & - & - \\
    Tents & \texttt{4x4de} & $ 56$  & $ 4442 $ \hfill $ (44.3\%)$ & $4781 \pm 86 $ \hfill $ (10.3\%)$ & $4493 \pm 155 $ \hfill $ (37.5\%)$ & $3485 \pm 63 $ \hfill $ (39.9\%)$ & $3485 \pm 456 $ \hfill $ (45.0\%)$ \\
    Towers & \texttt{3de} & $ 72$  & $ 4876 $ \hfill $ (1.0\%)$ & - & - & - & - \\
    Tracks & \texttt{4x4de} & $ 272$  & $ 5213 $ \hfill $ (0.5\%)$ & $4129 \pm nan $ \hfill $ (0.1\%)$ & $4217 \pm nan $ \hfill $ (1.6\%)$ & $5461 \pm 976 $ \hfill $ (0.3\%)$ & $5019 \pm 2297 $ \hfill $ (0.4\%)$ \\
    Twiddle & \texttt{2x3n2} & $ 98$  & $ 851 $ \hfill $ (100.0\%)$ & $\mathbf{8 \pm 1 } $ \hfill $ (99.9\%)$ & $348 \pm 466 $ \hfill $ (100.0\%)$ & $\mathbf{7 \pm 0 } $ \hfill $ (100.0\%)$ & $\mathbf{12 \pm 1 } $ \hfill $ (100.0\%)$ \\
    Undead & \texttt{3x3de} & $ 63$  & $ 4390 $ \hfill $ (40.1\%)$ & $4542 \pm 292 $ \hfill $ (5.7\%)$ & $4129 \pm 139 $ \hfill $ (40.0\%)$ & $3415 \pm 379 $ \hfill $ (42.8\%)$ & $3482 \pm 406 $ \hfill $ (46.1\%)$ \\
    Unequal & \texttt{3de} & $ 63$  & $ 4540 $ \hfill $ (6.7\%)$ & - & - & $2322 \pm 988 $ \hfill $ (38.7\%)$ & $3021 \pm 1368 $ \hfill $ (26.5\%)$ \\
    Unruly & \texttt{6x6dt} & $ 468$  & - & - & - & - & - \\
    Untangle & \texttt{4} & $ 150$  & $ 141 $ \hfill $ (100.0\%)$ & $\mathbf{13 \pm 1 } $ \hfill $ (100.0\%)$ & $\mathbf{35 \pm 58 } $ \hfill $ (100.0\%)$ & $\mathbf{12 \pm 0 } $ \hfill $ (100.0\%)$ & $\mathbf{7 \pm 0 } $ \hfill $ (100.0\%)$ \\
    Untangle & \texttt{6} & $ 79$  & $ 2165 $ \hfill $ (96.9\%)$ & $2295 \pm 66 $ \hfill $ (96.2\%)$ & - & - & - \\
    \midrule
    Summary & - & $ 217 $ & $ 1984 $ \hfill $(71.2\%)$ & $ 1604 \pm 801 $ \hfill $(61.6\%) (8)$ & $ 1619 \pm 380 $ \hfill $(82.8\%) (6)$ & $ 814 \pm 428 $ \hfill $(81.2\%) (21)$ & $ 1047 \pm 583 $ \hfill $(79.2\%) (10)$ \\
    \bottomrule
\end{tabular}
  \end{adjustbox}
\end{table}

\FloatBarrier
\newpage

\subsection{Episode Length and Early Termination Parameters}
\label{app:ep_length_params}
In \cref{tab:ep_length_params}, the puzzles and parameters used for training the agents for the ablation in \cref{experiments:ep_length} are shown in combination with the results.
Due to limited computational budget, we included only a subset of all puzzles at the easy human difficulty preset for DreamerV3.
Namely, we have selected all puzzles where a random policy was able to complete at least a single episode successfully within 10,000 steps in 1000 evaluations.
It contains a subset of the more challenging puzzles, as can be seen by the performance of many algorithms in \cref{tab:phat_table_state}.
For some puzzles, e.g. Netslide, Samegame, Sixteen and Untangle, terminating episodes early brings a benefit in final evaluation performance when using a large maximal episode length during training.
For the smaller maximal episode length, the difference is not always as pronounced.

 \begin{table}[h]
  \caption{
  Listed below are the puzzles and their corresponding supplied parameters. 
  For each setting, we report average success episode length with standard deviation with respect to the random seed, all averaged over all selected puzzles.
  In brackets, the percentage of successful episodes is reported.
  }
  \label{tab:ep_length_params}
  \centering
  \scriptsize
\begin{tabular}{lllllllllllll}
    \toprule
    \textbf{Puzzle} & \textbf{Supplied Parameters} & \textbf{\# Steps} & \textbf{ET} & \textbf{DreamerV3} \\
    \midrule
    \multirow{4}{*}{Bridges} & \multirow{4}{*}{\texttt{7x7i30e10m2d0}} & \multirow{2}{*}{$1e4$} & 10& $4183.0 \pm 2140.5$ \hfill (0.2\%) \\
    &&  & - & - \\
    &&   \multirow{2}{*}{$1e5$} & 10& $4017.9 \pm 1390.1$ \hfill (0.3\%) \\
    &&  & - & $4396.2 \pm 2517.2$ \hfill (0.3\%) \\
    \midrule
    \multirow{4}{*}{Cube} & \multirow{4}{*}{\texttt{c4x4}} & \multirow{2}{*}{$1e4$} & 10& $21.9 \pm 1.4$ \hfill (100.0\%) \\
    &&  & - & $21.4 \pm 0.9$ \hfill (100.0\%) \\
    &&   \multirow{2}{*}{$1e5$} & 10& $22.6 \pm 2.0$ \hfill (100.0\%) \\
    &&  & - & $21.3 \pm 1.2$ \hfill (100.0\%) \\
    \midrule
    \multirow{4}{*}{Flood} & \multirow{4}{*}{\texttt{12x12c6m5}} & \multirow{2}{*}{$1e4$} & 10& - \\
    &&  & - & - \\
    &&   \multirow{2}{*}{$1e5$} & 10& - \\
    &&  & - & - \\
    \midrule
    \multirow{4}{*}{Guess} & \multirow{4}{*}{\texttt{c6p4g10Bm}} & \multirow{2}{*}{$1e4$} & 10& - \\
    &&  & - & $1060.4 \pm 851.3$ \hfill (0.6\%) \\
    &&   \multirow{2}{*}{$1e5$} & 10& $2405.5 \pm 2476.4$ \hfill (0.5\%) \\
    &&  & - & $3165.2 \pm 1386.8$ \hfill (0.6\%) \\
    \midrule
    \multirow{4}{*}{Netslide} & \multirow{4}{*}{\texttt{3x3b1}} & \multirow{2}{*}{$1e4$} & 10& $3820.3 \pm 681.0$ \hfill (18.4\%) \\
    &&  & - & $3181.3 \pm 485.5$ \hfill (21.1\%) \\
    &&   \multirow{2}{*}{$1e5$} & 10& $3624.9 \pm 746.5$ \hfill (23.0\%) \\
    &&  & - & $4050.6 \pm 505.5$ \hfill (10.6\%) \\
    \midrule
    \multirow{4}{*}{Samegame} & \multirow{4}{*}{\texttt{5x5c3s2}} & \multirow{2}{*}{$1e4$} & 10& $53.8 \pm 7.5$ \hfill (38.3\%) \\
    &&  & - & $717.4 \pm 309.0$ \hfill (29.1\%) \\
    &&   \multirow{2}{*}{$1e5$} & 10& $47.3 \pm 6.6$ \hfill (36.7\%) \\
    &&  & - & $1542.9 \pm 824.0$ \hfill (26.4\%) \\
    \midrule
    \multirow{4}{*}{Signpost} & \multirow{4}{*}{\texttt{4x4c}} & \multirow{2}{*}{$1e4$} & 10& $6848.9 \pm 677.7$ \hfill (1.1\%) \\
    &&  & - & $6861.8 \pm 301.8$ \hfill (1.5\%) \\
    &&   \multirow{2}{*}{$1e5$} & 10& $6983.7 \pm 392.4$ \hfill (1.6\%) \\
    &&  & - & - \\
    \midrule
    \multirow{4}{*}{Sixteen} & \multirow{4}{*}{\texttt{3x3}} & \multirow{2}{*}{$1e4$} & 10& $4770.5 \pm 890.5$ \hfill (2.9\%) \\
    &&  & - & $4480.5 \pm 2259.3$ \hfill (25.5\%) \\
    &&   \multirow{2}{*}{$1e5$} & 10& $3193.3 \pm 2262.0$ \hfill (57.0\%) \\
    &&  & - & $3517.1 \pm 1846.7$ \hfill (23.5\%) \\
    \midrule
    \multirow{4}{*}{Undead} & \multirow{4}{*}{\texttt{4x4de}} & \multirow{2}{*}{$1e4$} & 10& $5378.0 \pm 1552.7$ \hfill (0.5\%) \\
    &&  & - & $5324.4 \pm 557.9$ \hfill (0.6\%) \\
    &&   \multirow{2}{*}{$1e5$} & 10& $5666.2 \pm 553.3$ \hfill (0.5\%) \\
    &&  & - & $5771.3 \pm 2323.6$ \hfill (0.4\%) \\
    \midrule
    \multirow{4}{*}{Untangle} & \multirow{4}{*}{\texttt{6}} & \multirow{2}{*}{$1e4$} & 10& $474.7 \pm 117.6$ \hfill (99.1\%) \\
    &&  & - & $1491.9 \pm 193.8$ \hfill (89.3\%) \\
    &&   \multirow{2}{*}{$1e5$} & 10& $597.0 \pm 305.5$ \hfill (96.3\%) \\
    &&  & - & $1338.4 \pm 283.6$ \hfill (88.7\%) \\
                \bottomrule
            \end{tabular}
\end{table}

\end{document}

%% file: img/puzzles/puzzle_collection.tex
\begin{figure}[h!]
    \centering
    \begin{minipage}[b]{0.2\textwidth}
        \includegraphics[width=\textwidth]{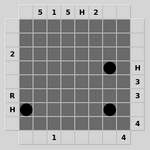}
    \end{minipage}
    \begin{minipage}[b]{0.75\textwidth}
        \vspace*{\fill}
        \captionsetup{singlelinecheck=off}
        \raggedright
        \raggedright
        \caption{\textbf{Black Box}: Find the hidden balls in the box by bouncing laser beams off them.}
        \vspace*{\fill}
    \end{minipage}
\end{figure}
\begin{figure}[h!]
    \centering
    \begin{minipage}[b]{0.2\textwidth}
        \includegraphics[width=\textwidth]{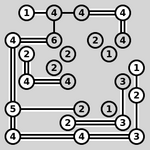}
    \end{minipage}
    \begin{minipage}[b]{0.75\textwidth}
        \vspace*{\fill}
        \captionsetup{singlelinecheck=off}
        \raggedright
        \raggedright
        \caption{\textbf{Bridges}: Connect all the islands with a network of bridges.}
        \vspace*{\fill}
    \end{minipage}
\end{figure}
\begin{figure}[h!]
    \centering
    \begin{minipage}[b]{0.2\textwidth}
        \includegraphics[width=\textwidth]{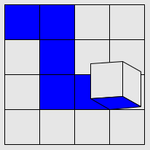}
    \end{minipage}
    \begin{minipage}[b]{0.75\textwidth}
        \vspace*{\fill}
        \captionsetup{singlelinecheck=off}
        \raggedright
        \raggedright
        \caption{\textbf{Cube}: Pick up all the blue squares by rolling the cube over them.}
        \vspace*{\fill}
    \end{minipage}
\end{figure}
\begin{figure}[h!]
    \centering
    \begin{minipage}[b]{0.2\textwidth}
        \includegraphics[width=\textwidth]{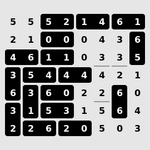}
    \end{minipage}
    \begin{minipage}[b]{0.75\textwidth}
        \vspace*{\fill}
        \captionsetup{singlelinecheck=off}
        \raggedright
        \raggedright
        \caption{\textbf{Dominosa}: Tile the rectangle with a full set of dominoes.}
        \vspace*{\fill}
    \end{minipage}
\end{figure}
\begin{figure}[h!]
    \centering
    \begin{minipage}[b]{0.2\textwidth}
        \includegraphics[width=\textwidth]{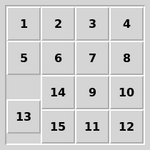}
    \end{minipage}
    \begin{minipage}[b]{0.75\textwidth}
        \vspace*{\fill}
        \captionsetup{singlelinecheck=off}
        \raggedright
        \raggedright
        \caption{\textbf{Fifteen}: Slide the tiles around to arrange them into order.}
        \vspace*{\fill}
    \end{minipage}
\end{figure}
\begin{figure}[h!]
    \centering
    \begin{minipage}[b]{0.2\textwidth}
        \includegraphics[width=\textwidth]{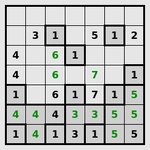}
    \end{minipage}
    \begin{minipage}[b]{0.75\textwidth}
        \vspace*{\fill}
        \captionsetup{singlelinecheck=off}
        \raggedright
        \raggedright
        \caption{\textbf{Filling}: Mark every square with the area of its containing region.}
        \vspace*{\fill}
    \end{minipage}
\end{figure}
\begin{figure}[h!]
    \centering
    \begin{minipage}[b]{0.2\textwidth}
        \includegraphics[width=\textwidth]{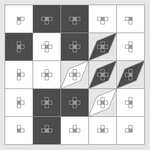}
    \end{minipage}
    \begin{minipage}[b]{0.75\textwidth}
        \vspace*{\fill}
        \captionsetup{singlelinecheck=off}
        \raggedright
        \raggedright
        \caption{\textbf{Flip}: Flip groups of squares to light them all up at once.}
        \vspace*{\fill}
    \end{minipage}
\end{figure}
\begin{figure}[h!]
    \centering
    \begin{minipage}[b]{0.2\textwidth}
        \includegraphics[width=\textwidth]{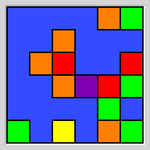}
    \end{minipage}
    \begin{minipage}[b]{0.75\textwidth}
        \vspace*{\fill}
        \captionsetup{singlelinecheck=off}
        \raggedright
        \raggedright
        \caption{\textbf{Flood}: Turn the grid the same colour in as few flood fills as possible.}
        \vspace*{\fill}
    \end{minipage}
\end{figure}
\begin{figure}[h!]
    \centering
    \begin{minipage}[b]{0.2\textwidth}
        \includegraphics[width=\textwidth]{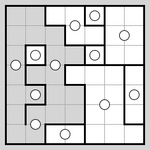}
    \end{minipage}
    \begin{minipage}[b]{0.75\textwidth}
        \vspace*{\fill}
        \captionsetup{singlelinecheck=off}
        \raggedright
        \raggedright
        \caption{\textbf{Galaxies}: Divide the grid into rotationally symmetric regions each centred on a dot.}
        \vspace*{\fill}
    \end{minipage}
\end{figure}
\begin{figure}[h!]
    \centering
    \begin{minipage}[b]{0.2\textwidth}
        \includegraphics[width=\textwidth]{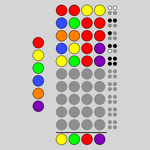}
    \end{minipage}
    \begin{minipage}[b]{0.75\textwidth}
        \vspace*{\fill}
        \captionsetup{singlelinecheck=off}
        \raggedright
        \raggedright
        \caption{\textbf{Guess}: Guess the hidden combination of colours.}
        \vspace*{\fill}
    \end{minipage}
\end{figure}
\begin{figure}[h!]
    \centering
    \begin{minipage}[b]{0.2\textwidth}
        \includegraphics[width=\textwidth]{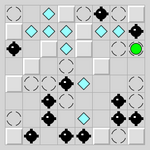}
    \end{minipage}
    \begin{minipage}[b]{0.75\textwidth}
        \vspace*{\fill}
        \captionsetup{singlelinecheck=off}
        \raggedright
        \raggedright
        \caption{\textbf{Inertia}: Collect all the gems without running into any of the mines.}
        \vspace*{\fill}
    \end{minipage}
\end{figure}
\begin{figure}[h!]
    \centering
    \begin{minipage}[b]{0.2\textwidth}
        \includegraphics[width=\textwidth]{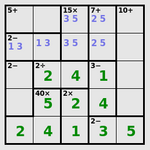}
    \end{minipage}
    \begin{minipage}[b]{0.75\textwidth}
        \vspace*{\fill}
        \captionsetup{singlelinecheck=off}
        \raggedright
        \raggedright
        \caption{\textbf{Keen}: Complete the latin square in accordance with the arithmetic clues.}
        \vspace*{\fill}
    \end{minipage}
\end{figure}
\begin{figure}[h!]
    \centering
    \begin{minipage}[b]{0.2\textwidth}
        \includegraphics[width=\textwidth]{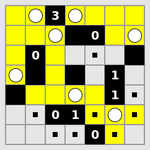}
    \end{minipage}
    \begin{minipage}[b]{0.75\textwidth}
        \vspace*{\fill}
        \captionsetup{singlelinecheck=off}
        \raggedright
        \raggedright
        \caption{\textbf{Light Up}: Place bulbs to light up all the squares.}
        \vspace*{\fill}
    \end{minipage}
\end{figure}
\begin{figure}[h!]
    \centering
    \begin{minipage}[b]{0.2\textwidth}
        \includegraphics[width=\textwidth]{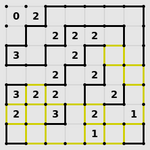}
    \end{minipage}
    \begin{minipage}[b]{0.75\textwidth}
        \vspace*{\fill}
        \captionsetup{singlelinecheck=off}
        \raggedright
        \raggedright
        \caption{\textbf{Loopy}: Draw a single closed loop, given clues about number of adjacent edges.}
        \vspace*{\fill}
    \end{minipage}
\end{figure}
\begin{figure}[h!]
    \centering
    \begin{minipage}[b]{0.2\textwidth}
        \includegraphics[width=\textwidth]{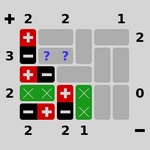}
    \end{minipage}
    \begin{minipage}[b]{0.75\textwidth}
        \vspace*{\fill}
        \captionsetup{singlelinecheck=off}
        \raggedright
        \raggedright
        \caption{\textbf{Magnets}: Place magnets to satisfy the clues and avoid like poles touching.}
        \vspace*{\fill}
    \end{minipage}
\end{figure}
\begin{figure}[h!]
    \centering
    \begin{minipage}[b]{0.2\textwidth}
        \includegraphics[width=\textwidth]{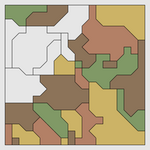}
    \end{minipage}
    \begin{minipage}[b]{0.75\textwidth}
        \vspace*{\fill}
        \captionsetup{singlelinecheck=off}
        \raggedright
        \raggedright
        \caption{\textbf{Map}: Colour the map so that adjacent regions are never the same colour.}
        \vspace*{\fill}
    \end{minipage}
\end{figure}
\begin{figure}[h!]
    \centering
    \begin{minipage}[b]{0.2\textwidth}
        \includegraphics[width=\textwidth]{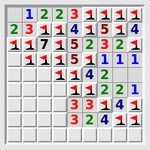}
    \end{minipage}
    \begin{minipage}[b]{0.75\textwidth}
        \vspace*{\fill}
        \captionsetup{singlelinecheck=off}
        \raggedright
        \raggedright
        \caption{\textbf{Mines}: Find all the mines without treading on any of them.}
        \vspace*{\fill}
    \end{minipage}
\end{figure}
\begin{figure}[h!]
    \centering
    \begin{minipage}[b]{0.2\textwidth}
        \includegraphics[width=\textwidth]{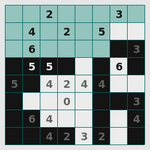}
    \end{minipage}
    \begin{minipage}[b]{0.75\textwidth}
        \vspace*{\fill}
        \captionsetup{singlelinecheck=off}
        \raggedright
        \raggedright
        \caption{\textbf{Mosaic}: Fill in the grid given clues about number of nearby black squares.}
        \vspace*{\fill}
    \end{minipage}
\end{figure}
\begin{figure}[h!]
    \centering
    \begin{minipage}[b]{0.2\textwidth}
        \includegraphics[width=\textwidth]{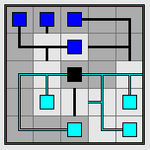}
    \end{minipage}
    \begin{minipage}[b]{0.75\textwidth}
        \vspace*{\fill}
        \captionsetup{singlelinecheck=off}
        \raggedright
        \raggedright
        \caption{\textbf{Net}: Rotate each tile to reassemble the network.}
        \vspace*{\fill}
    \end{minipage}
\end{figure}
\begin{figure}[h!]
    \centering
    \begin{minipage}[b]{0.2\textwidth}
        \includegraphics[width=\textwidth]{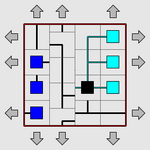}
    \end{minipage}
    \begin{minipage}[b]{0.75\textwidth}
        \vspace*{\fill}
        \captionsetup{singlelinecheck=off}
        \raggedright
        \raggedright
        \caption{\textbf{Netslide}: Slide a row at a time to reassemble the network.}
        \vspace*{\fill}
    \end{minipage}
\end{figure}
\begin{figure}[h!]
    \centering
    \begin{minipage}[b]{0.2\textwidth}
        \includegraphics[width=\textwidth]{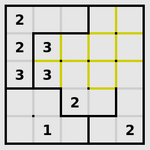}
    \end{minipage}
    \begin{minipage}[b]{0.75\textwidth}
        \vspace*{\fill}
        \captionsetup{singlelinecheck=off}
        \raggedright
        \raggedright
        \caption{\textbf{Palisade}: Divide the grid into equal-sized areas in accordance with the clues.}
        \vspace*{\fill}
    \end{minipage}
\end{figure}
\begin{figure}[h!]
    \centering
    \begin{minipage}[b]{0.2\textwidth}
        \includegraphics[width=\textwidth]{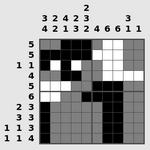}
    \end{minipage}
    \begin{minipage}[b]{0.75\textwidth}
        \vspace*{\fill}
        \captionsetup{singlelinecheck=off}
        \raggedright
        \raggedright
        \caption{\textbf{Pattern}: Fill in the pattern in the grid, given only the lengths of runs of black squares.}
        \vspace*{\fill}
    \end{minipage}
\end{figure}
\begin{figure}[h!]
    \centering
    \begin{minipage}[b]{0.2\textwidth}
        \includegraphics[width=\textwidth]{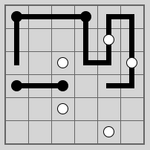}
    \end{minipage}
    \begin{minipage}[b]{0.75\textwidth}
        \vspace*{\fill}
        \captionsetup{singlelinecheck=off}
        \raggedright
        \raggedright
        \caption{\textbf{Pearl}: Draw a single closed loop, given clues about corner and straight squares.}
        \vspace*{\fill}
    \end{minipage}
\end{figure}
\begin{figure}[h!]
    \centering
    \begin{minipage}[b]{0.2\textwidth}
        \includegraphics[width=\textwidth]{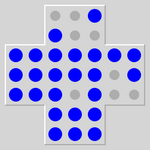}
    \end{minipage}
    \begin{minipage}[b]{0.75\textwidth}
        \vspace*{\fill}
        \captionsetup{singlelinecheck=off}
        \raggedright
        \raggedright
        \caption{\textbf{Pegs}: Jump pegs over each other to remove all but one.}
        \vspace*{\fill}
    \end{minipage}
\end{figure}
\begin{figure}[h!]
    \centering
    \begin{minipage}[b]{0.2\textwidth}
        \includegraphics[width=\textwidth]{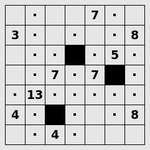}
    \end{minipage}
    \begin{minipage}[b]{0.75\textwidth}
        \vspace*{\fill}
        \captionsetup{singlelinecheck=off}
        \raggedright
        \raggedright
        \caption{\textbf{Range}: Place black squares to limit the visible distance from each numbered cell.}
        \vspace*{\fill}
    \end{minipage}
\end{figure}
\begin{figure}[h!]
    \centering
    \begin{minipage}[b]{0.2\textwidth}
        \includegraphics[width=\textwidth]{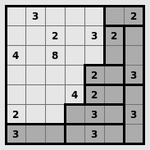}
    \end{minipage}
    \begin{minipage}[b]{0.75\textwidth}
        \vspace*{\fill}
        \captionsetup{singlelinecheck=off}
        \raggedright
        \raggedright
        \caption{\textbf{Rectangles}: Divide the grid into rectangles with areas equal to the numbers.}
        \vspace*{\fill}
    \end{minipage}
\end{figure}
\begin{figure}[h!]
    \centering
    \begin{minipage}[b]{0.2\textwidth}
        \includegraphics[width=\textwidth]{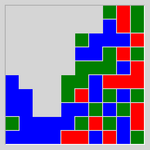}
    \end{minipage}
    \begin{minipage}[b]{0.75\textwidth}
        \vspace*{\fill}
        \captionsetup{singlelinecheck=off}
        \raggedright
        \raggedright
        \caption{\textbf{Same Game}: Clear the grid by removing touching groups of the same colour squares.}
        \vspace*{\fill}
    \end{minipage}
\end{figure}
\begin{figure}[h!]
    \centering
    \begin{minipage}[b]{0.2\textwidth}
        \includegraphics[width=\textwidth]{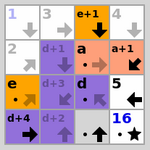}
    \end{minipage}
    \begin{minipage}[b]{0.75\textwidth}
        \vspace*{\fill}
        \captionsetup{singlelinecheck=off}
        \raggedright
        \raggedright
        \caption{\textbf{Signpost}: Connect the squares into a path following the arrows.}
        \vspace*{\fill}
    \end{minipage}
\end{figure}
\begin{figure}[h!]
    \centering
    \begin{minipage}[b]{0.2\textwidth}
        \includegraphics[width=\textwidth]{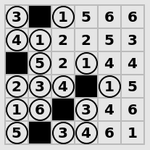}
    \end{minipage}
    \begin{minipage}[b]{0.75\textwidth}
        \vspace*{\fill}
        \captionsetup{singlelinecheck=off}
        \raggedright
        \raggedright
        \caption{\textbf{Singles}: Black out the right set of duplicate numbers.}
        \vspace*{\fill}
    \end{minipage}
\end{figure}
\begin{figure}[h!]
    \centering
    \begin{minipage}[b]{0.2\textwidth}
        \includegraphics[width=\textwidth]{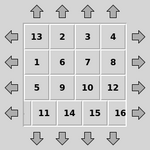}
    \end{minipage}
    \begin{minipage}[b]{0.75\textwidth}
        \vspace*{\fill}
        \captionsetup{singlelinecheck=off}
        \raggedright
        \raggedright
        \caption{\textbf{Sixteen}: Slide a row at a time to arrange the tiles into order.}
        \vspace*{\fill}
    \end{minipage}
\end{figure}
\begin{figure}[h!]
    \centering
    \begin{minipage}[b]{0.2\textwidth}
        \includegraphics[width=\textwidth]{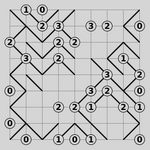}
    \end{minipage}
    \begin{minipage}[b]{0.75\textwidth}
        \vspace*{\fill}
        \captionsetup{singlelinecheck=off}
        \raggedright
        \raggedright
        \caption{\textbf{Slant}: Draw a maze of slanting lines that matches the clues.}
        \vspace*{\fill}
    \end{minipage}
\end{figure}
\begin{figure}[h!]
    \centering
    \begin{minipage}[b]{0.2\textwidth}
        \includegraphics[width=\textwidth]{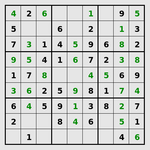}
    \end{minipage}
    \begin{minipage}[b]{0.75\textwidth}
        \vspace*{\fill}
        \captionsetup{singlelinecheck=off}
        \raggedright
        \raggedright
        \caption{\textbf{Solo}: Fill in the grid so that each row, column and square block contains one of every digit.}
        \vspace*{\fill}
    \end{minipage}
\end{figure}
\begin{figure}[h!]
    \centering
    \begin{minipage}[b]{0.2\textwidth}
        \includegraphics[width=\textwidth]{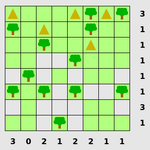}
    \end{minipage}
    \begin{minipage}[b]{0.75\textwidth}
        \vspace*{\fill}
        \captionsetup{singlelinecheck=off}
        \raggedright
        \raggedright
        \caption{\textbf{Tents}: Place a tent next to each tree.}
        \vspace*{\fill}
    \end{minipage}
\end{figure}
\begin{figure}[h!]
    \centering
    \begin{minipage}[b]{0.2\textwidth}
        \includegraphics[width=\textwidth]{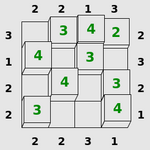}
    \end{minipage}
    \begin{minipage}[b]{0.75\textwidth}
        \vspace*{\fill}
        \captionsetup{singlelinecheck=off}
        \raggedright
        \raggedright
        \caption{\textbf{Towers}: Complete the latin square of towers in accordance with the clues.}
        \vspace*{\fill}
    \end{minipage}
\end{figure}
\begin{figure}[h!]
    \centering
    \begin{minipage}[b]{0.2\textwidth}
        \includegraphics[width=\textwidth]{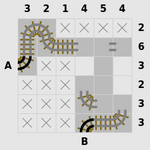}
    \end{minipage}
    \begin{minipage}[b]{0.75\textwidth}
        \vspace*{\fill}
        \captionsetup{singlelinecheck=off}
        \raggedright
        \raggedright
        \caption{\textbf{Tracks}: Fill in the railway track according to the clues.}
        \vspace*{\fill}
    \end{minipage}
\end{figure}
\begin{figure}[h!]
    \centering
    \begin{minipage}[b]{0.2\textwidth}
        \includegraphics[width=\textwidth]{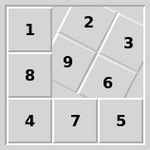}
    \end{minipage}
    \begin{minipage}[b]{0.75\textwidth}
        \vspace*{\fill}
        \captionsetup{singlelinecheck=off}
        \raggedright
        \raggedright
        \caption{\textbf{Twiddle}: Rotate the tiles around themselves to arrange them into order.}
        \vspace*{\fill}
    \end{minipage}
\end{figure}
\begin{figure}[h!]
    \centering
    \begin{minipage}[b]{0.2\textwidth}
        \includegraphics[width=\textwidth]{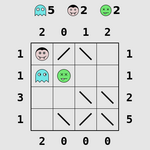}
    \end{minipage}
    \begin{minipage}[b]{0.75\textwidth}
        \vspace*{\fill}
        \captionsetup{singlelinecheck=off}
        \raggedright
        \raggedright
        \caption{\textbf{Undead}: Place ghosts, vampires and zombies so that the right numbers of them can be seen in mirrors.}
        \vspace*{\fill}
    \end{minipage}
\end{figure}
\begin{figure}[h!]
    \centering
    \begin{minipage}[b]{0.2\textwidth}
        \includegraphics[width=\textwidth]{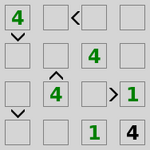}
    \end{minipage}
    \begin{minipage}[b]{0.75\textwidth}
        \vspace*{\fill}
        \captionsetup{singlelinecheck=off}
        \raggedright
        \raggedright
        \caption{\textbf{Unequal}: Complete the latin square in accordance with the > signs.}
        \vspace*{\fill}
    \end{minipage}
\end{figure}
\begin{figure}[h!]
    \centering
    \begin{minipage}[b]{0.2\textwidth}
        \includegraphics[width=\textwidth]{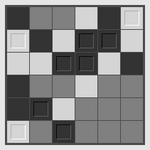}
    \end{minipage}
    \begin{minipage}[b]{0.75\textwidth}
        \vspace*{\fill}
        \captionsetup{singlelinecheck=off}
        \raggedright
        \raggedright
        \caption{\textbf{Unruly}: Fill in the black and white grid to avoid runs of three.}
        \vspace*{\fill}
    \end{minipage}
\end{figure}
\begin{figure}[h!]
    \centering
    \begin{minipage}[b]{0.2\textwidth}
        \includegraphics[width=\textwidth]{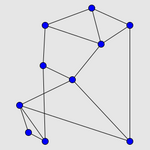}
    \end{minipage}
    \begin{minipage}[b]{0.75\textwidth}
        \vspace*{\fill}
        \captionsetup{singlelinecheck=off}
        \raggedright
        \raggedright
        \caption{\textbf{Untangle}: Reposition the points so that the lines do not cross.}
        \vspace*{\fill}
    \end{minipage}
\end{figure}

%% file: neurips_data_2024.bbl
\newcommand{\SortNoop}[1]{}
\begin{thebibliography}{67}
\providecommand{\natexlab}[1]{#1}
\providecommand{\url}[1]{\texttt{#1}}
\expandafter\ifx\csname urlstyle\endcsname\relax
  \providecommand{\doi}[1]{doi: #1}\else
  \providecommand{\doi}{doi: \begingroup \urlstyle{rm}\Url}\fi

\bibitem[Serafini and Garcez(2016)]{serafini2016logic}
Luciano Serafini and Artur~d'Avila Garcez.
\newblock Logic tensor networks: Deep learning and logical reasoning from data and knowledge.
\newblock \emph{arXiv preprint arXiv:1606.04422}, 2016.

\bibitem[Dai et~al.(2019)Dai, Xu, Yu, and Zhou]{dai2019bridging}
Wang-Zhou Dai, Qiuling Xu, Yang Yu, and Zhi-Hua Zhou.
\newblock Bridging machine learning and logical reasoning by abductive learning.
\newblock \emph{Advances in Neural Information Processing Systems}, 32, 2019.

\bibitem[Li et~al.(2020)Li, Gimeno, Kohli, and Vinyals]{li2020strong}
Yujia Li, Felix Gimeno, Pushmeet Kohli, and Oriol Vinyals.
\newblock Strong generalization and efficiency in neural programs.
\newblock \emph{arXiv preprint arXiv:2007.03629}, 2020.

\bibitem[Veli{\v{c}}kovi{\'c} and Blundell(2021)]{velivckovic2021neural}
Petar Veli{\v{c}}kovi{\'c} and Charles Blundell.
\newblock Neural algorithmic reasoning.
\newblock \emph{Patterns}, 2\penalty0 (7), 2021.

\bibitem[Masry et~al.(2022)Masry, Long, Tan, Joty, and Hoque]{masry2022chartqa}
Ahmed Masry, Do~Long, Jia~Qing Tan, Shafiq Joty, and Enamul Hoque.
\newblock Chartqa: A benchmark for question answering about charts with visual and logical reasoning.
\newblock In \emph{Findings of the Association for Computational Linguistics: ACL 2022}, pages 2263--2279, 2022.

\bibitem[Jiao et~al.(2022)Jiao, Guo, Song, and Nie]{jiao2022merit}
Fangkai Jiao, Yangyang Guo, Xuemeng Song, and Liqiang Nie.
\newblock Merit: Meta-path guided contrastive learning for logical reasoning.
\newblock In \emph{Findings of the Association for Computational Linguistics: ACL 2022}, pages 3496--3509, 2022.

\bibitem[Bardin et~al.(2023)Bardin, Jha, and Ganesh]{bardin2023machine}
S{\'e}bastien Bardin, Somesh Jha, and Vijay Ganesh.
\newblock Machine learning and logical reasoning: The new frontier (dagstuhl seminar 22291).
\newblock In \emph{Dagstuhl Reports}, volume~12. Schloss Dagstuhl-Leibniz-Zentrum f{\"u}r Informatik, 2023.

\bibitem[Li et~al.(2021)Li, Yu, Wu, and Paulson]{li2021isarstep}
Wenda Li, Lei Yu, Yuhuai Wu, and Lawrence~C Paulson.
\newblock Isarstep: a benchmark for high-level mathematical reasoning.
\newblock In \emph{International Conference on Learning Representations}, 2021.

\bibitem[Veli{\v{c}}kovi{\'c} et~al.(2022)Veli{\v{c}}kovi{\'c}, Badia, Budden, Pascanu, Banino, Dashevskiy, Hadsell, and Blundell]{velikovic2022algorithmicreasoning}
Petar Veli{\v{c}}kovi{\'c}, Adri{\`a}~Puigdom{\`e}nech Badia, David Budden, Razvan Pascanu, Andrea Banino, Misha Dashevskiy, Raia Hadsell, and Charles Blundell.
\newblock The {CLRS} algorithmic reasoning benchmark.
\newblock In Kamalika Chaudhuri, Stefanie Jegelka, Le~Song, Csaba Szepesvari, Gang Niu, and Sivan Sabato, editors, \emph{Proceedings of the 39th International Conference on Machine Learning}, volume 162 of \emph{Proceedings of Machine Learning Research}, pages 22084--22102. PMLR, 17--23 Jul 2022.
\newblock URL \url{https://proceedings.mlr.press/v162/velickovic22a.html}.

\bibitem[Srivastava et~al.(2022)Srivastava, Rastogi, Rao, Shoeb, Abid, Fisch, Brown, Santoro, Gupta, Garriga-Alonso, et~al.]{srivastava2022beyond}
Aarohi Srivastava, Abhinav Rastogi, Abhishek Rao, Abu Awal~Md Shoeb, Abubakar Abid, Adam Fisch, Adam~R Brown, Adam Santoro, Aditya Gupta, Adri{\`a} Garriga-Alonso, et~al.
\newblock Beyond the imitation game: Quantifying and extrapolating the capabilities of language models.
\newblock \emph{arXiv preprint arXiv:2206.04615}, 2022.

\bibitem[Mnih et~al.(2013)Mnih, Kavukcuoglu, Silver, Graves, Antonoglou, Wierstra, and Riedmiller]{mnih2013atari}
Volodymyr Mnih, Koray Kavukcuoglu, David Silver, Alex Graves, Ioannis Antonoglou, Daan Wierstra, and Martin~A. Riedmiller.
\newblock Playing {A}tari with {D}eep {R}einforcement {L}earning.
\newblock \emph{CoRR}, abs/1312.5602, 2013.
\newblock URL \url{http://arxiv.org/abs/1312.5602}.

\bibitem[Tang et~al.(2017)Tang, Houthooft, Foote, Stooke, Xi~Chen, Duan, Schulman, DeTurck, and Abbeel]{tang2017exploration}
Haoran Tang, Rein Houthooft, Davis Foote, Adam Stooke, OpenAI Xi~Chen, Yan Duan, John Schulman, Filip DeTurck, and Pieter Abbeel.
\newblock \# exploration: A study of count-based exploration for deep reinforcement learning.
\newblock \emph{Advances in neural information processing systems}, 30, 2017.

\bibitem[Silver et~al.(2018)Silver, Hubert, Schrittwieser, Antonoglou, Lai, Guez, Lanctot, Sifre, Kumaran, Graepel, et~al.]{silver2018general}
David Silver, Thomas Hubert, Julian Schrittwieser, Ioannis Antonoglou, Matthew Lai, Arthur Guez, Marc Lanctot, Laurent Sifre, Dharshan Kumaran, Thore Graepel, et~al.
\newblock A general reinforcement learning algorithm that masters chess, shogi, and go through self-play.
\newblock \emph{Science}, 362\penalty0 (6419):\penalty0 1140--1144, 2018.

\bibitem[Badia et~al.(2020)Badia, Piot, Kapturowski, Sprechmann, Vitvitskyi, Guo, and Blundell]{badia2020agent57}
Adri{\`a}~Puigdom{\`e}nech Badia, Bilal Piot, Steven Kapturowski, Pablo Sprechmann, Alex Vitvitskyi, Zhaohan~Daniel Guo, and Charles Blundell.
\newblock Agent57: Outperforming the atari human benchmark.
\newblock In \emph{International conference on machine learning}, pages 507--517. PMLR, 2020.

\bibitem[Wurman et~al.(2022)Wurman, Barrett, Kawamoto, MacGlashan, Subramanian, Walsh, Capobianco, Devlic, Eckert, Fuchs, et~al.]{wurman2022outracing}
Peter~R Wurman, Samuel Barrett, Kenta Kawamoto, James MacGlashan, Kaushik Subramanian, Thomas~J Walsh, Roberto Capobianco, Alisa Devlic, Franziska Eckert, Florian Fuchs, et~al.
\newblock Outracing champion gran turismo drivers with deep reinforcement learning.
\newblock \emph{Nature}, 602\penalty0 (7896):\penalty0 223--228, 2022.

\bibitem[Kalashnikov et~al.(2018)Kalashnikov, Irpan, Pastor, Ibarz, Herzog, Jang, Quillen, Holly, Kalakrishnan, Vanhoucke, et~al.]{kalashnikov2018scalable}
Dmitry Kalashnikov, Alex Irpan, Peter Pastor, Julian Ibarz, Alexander Herzog, Eric Jang, Deirdre Quillen, Ethan Holly, Mrinal Kalakrishnan, Vincent Vanhoucke, et~al.
\newblock Scalable deep reinforcement learning for vision-based robotic manipulation.
\newblock In \emph{Conference on Robot Learning}, pages 651--673. PMLR, 2018.

\bibitem[Kiran et~al.(2021)Kiran, Sobh, Talpaert, Mannion, Al~Sallab, Yogamani, and P{\'e}rez]{kiran2021deep}
B~Ravi Kiran, Ibrahim Sobh, Victor Talpaert, Patrick Mannion, Ahmad~A Al~Sallab, Senthil Yogamani, and Patrick P{\'e}rez.
\newblock Deep reinforcement learning for autonomous driving: A survey.
\newblock \emph{IEEE Transactions on Intelligent Transportation Systems}, 23\penalty0 (6):\penalty0 4909--4926, 2021.

\bibitem[Rudin et~al.(2022)Rudin, Hoeller, Reist, and Hutter]{rudin2022learning}
Nikita Rudin, David Hoeller, Philipp Reist, and Marco Hutter.
\newblock Learning to walk in minutes using massively parallel deep reinforcement learning.
\newblock In \emph{Conference on Robot Learning}, pages 91--100. PMLR, 2022.

\bibitem[Rana et~al.(2023)Rana, Xu, Tidd, Milford, and S{\"u}nderhauf]{rana2023residual}
Krishan Rana, Ming Xu, Brendan Tidd, Michael Milford, and Niko S{\"u}nderhauf.
\newblock Residual skill policies: Learning an adaptable skill-based action space for reinforcement learning for robotics.
\newblock In \emph{Conference on Robot Learning}, pages 2095--2104. PMLR, 2023.

\bibitem[Wang and Hong(2020)]{wang2020reinforcement}
Zhe Wang and Tianzhen Hong.
\newblock Reinforcement learning for building controls: The opportunities and challenges.
\newblock \emph{Applied Energy}, 269:\penalty0 115036, 2020.

\bibitem[Wu et~al.(2022)Wu, Lei, He, Zhang, and Ji]{wu2022deep}
Di~Wu, Yin Lei, Maoen He, Chunjiong Zhang, and Li~Ji.
\newblock Deep reinforcement learning-based path control and optimization for unmanned ships.
\newblock \emph{Wireless Communications and Mobile Computing}, 2022:\penalty0 1--8, 2022.

\bibitem[Brunke et~al.(2022)Brunke, Greeff, Hall, Yuan, Zhou, Panerati, and Schoellig]{brunke2022safe}
Lukas Brunke, Melissa Greeff, Adam~W Hall, Zhaocong Yuan, Siqi Zhou, Jacopo Panerati, and Angela~P Schoellig.
\newblock Safe learning in robotics: From learning-based control to safe reinforcement learning.
\newblock \emph{Annual Review of Control, Robotics, and Autonomous Systems}, 5:\penalty0 411--444, 2022.

\bibitem[Todorov et~al.(2012)Todorov, Erez, and Tassa]{todorov2012mujoco}
Emanuel Todorov, Tom Erez, and Yuval Tassa.
\newblock Mujoco: A physics engine for model-based control.
\newblock In \emph{2012 IEEE/RSJ international conference on intelligent robots and systems}, pages 5026--5033. IEEE, 2012.

\bibitem[Bellemare et~al.(2013)Bellemare, Naddaf, Veness, and Bowling]{bellemare2013arcade}
Marc~G Bellemare, Yavar Naddaf, Joel Veness, and Michael Bowling.
\newblock The arcade learning environment: An evaluation platform for general agents.
\newblock \emph{Journal of Artificial Intelligence Research}, 47:\penalty0 253--279, 2013.

\bibitem[Brockman et~al.(2016)Brockman, Cheung, Pettersson, Schneider, Schulman, Tang, and Zaremba]{brockman2016openai}
Greg Brockman, Vicki Cheung, Ludwig Pettersson, Jonas Schneider, John Schulman, Jie Tang, and Wojciech Zaremba.
\newblock Openai gym.
\newblock \emph{arXiv preprint arXiv:1606.01540}, 2016.

\bibitem[Duan et~al.(2016)Duan, Chen, Houthooft, Schulman, and Abbeel]{duan2016benchmarking}
Yan Duan, Xi~Chen, Rein Houthooft, John Schulman, and Pieter Abbeel.
\newblock Benchmarking deep reinforcement learning for continuous control.
\newblock In \emph{International conference on machine learning}, pages 1329--1338. PMLR, 2016.

\bibitem[Tassa et~al.(2018)Tassa, Doron, Muldal, Erez, Li, Casas, Budden, Abdolmaleki, Merel, Lefrancq, et~al.]{tassa2018deepmind}
Yuval Tassa, Yotam Doron, Alistair Muldal, Tom Erez, Yazhe Li, Diego de~Las Casas, David Budden, Abbas Abdolmaleki, Josh Merel, Andrew Lefrancq, et~al.
\newblock Deepmind control suite.
\newblock \emph{arXiv preprint arXiv:1801.00690}, 2018.

\bibitem[C\^ot\'e et~al.(2018)C\^ot\'e, K\'ad\'ar, Yuan, Kybartas, Barnes, Fine, Moore, Tao, Hausknecht, Asri, Adada, Tay, and Trischler]{cote2018textworld}
Marc-Alexandre C\^ot\'e, \'Akos K\'ad\'ar, Xingdi Yuan, Ben Kybartas, Tavian Barnes, Emery Fine, James Moore, Ruo~Yu Tao, Matthew Hausknecht, Layla~El Asri, Mahmoud Adada, Wendy Tay, and Adam Trischler.
\newblock Textworld: A learning environment for text-based games.
\newblock \emph{CoRR}, abs/1806.11532, 2018.

\bibitem[Lanctot et~al.(2019)Lanctot, Lockhart, Lespiau, Zambaldi, Upadhyay, P\'{e}rolat, Srinivasan, Timbers, Tuyls, Omidshafiei, Hennes, Morrill, Muller, Ewalds, Faulkner, Kram\'{a}r, Vylder, Saeta, Bradbury, Ding, Borgeaud, Lai, Schrittwieser, Anthony, Hughes, Danihelka, and Ryan-Davis]{Lanctot2019OpenSpiel}
Marc Lanctot, Edward Lockhart, Jean-Baptiste Lespiau, Vinicius Zambaldi, Satyaki Upadhyay, Julien P\'{e}rolat, Sriram Srinivasan, Finbarr Timbers, Karl Tuyls, Shayegan Omidshafiei, Daniel Hennes, Dustin Morrill, Paul Muller, Timo Ewalds, Ryan Faulkner, J\'{a}nos Kram\'{a}r, Bart~De Vylder, Brennan Saeta, James Bradbury, David Ding, Sebastian Borgeaud, Matthew Lai, Julian Schrittwieser, Thomas Anthony, Edward Hughes, Ivo Danihelka, and Jonah Ryan-Davis.
\newblock {OpenSpiel}: A framework for reinforcement learning in games.
\newblock \emph{CoRR}, abs/1908.09453, 2019.
\newblock URL \url{http://arxiv.org/abs/1908.09453}.

\bibitem[Jiang and Luo(2019)]{jiang2019neural}
Zhengyao Jiang and Shan Luo.
\newblock Neural logic reinforcement learning.
\newblock In \emph{International conference on machine learning}, pages 3110--3119. PMLR, 2019.

\bibitem[Fawzi et~al.(2022)Fawzi, Balog, Huang, Hubert, Romera-Paredes, Barekatain, Novikov, R~Ruiz, Schrittwieser, Swirszcz, et~al.]{fawzi2022discovering}
Alhussein Fawzi, Matej Balog, Aja Huang, Thomas Hubert, Bernardino Romera-Paredes, Mohammadamin Barekatain, Alexander Novikov, Francisco~J R~Ruiz, Julian Schrittwieser, Grzegorz Swirszcz, et~al.
\newblock Discovering faster matrix multiplication algorithms with reinforcement learning.
\newblock \emph{Nature}, 610\penalty0 (7930):\penalty0 47--53, 2022.

\bibitem[Mankowitz et~al.(2023)Mankowitz, Michi, Zhernov, Gelmi, Selvi, Paduraru, Leurent, Iqbal, Lespiau, Ahern, et~al.]{mankowitz2023faster}
Daniel~J Mankowitz, Andrea Michi, Anton Zhernov, Marco Gelmi, Marco Selvi, Cosmin Paduraru, Edouard Leurent, Shariq Iqbal, Jean-Baptiste Lespiau, Alex Ahern, et~al.
\newblock Faster sorting algorithms discovered using deep reinforcement learning.
\newblock \emph{Nature}, 618\penalty0 (7964):\penalty0 257--263, 2023.

\bibitem[Lai(2015)]{lai2015giraffe}
Matthew Lai.
\newblock Giraffe: Using deep reinforcement learning to play chess.
\newblock \emph{arXiv preprint arXiv:1509.01549}, 2015.

\bibitem[Silver et~al.(2016)Silver, Huang, Maddison, Guez, Sifre, van~den Driessche, Schrittwieser, Antonoglou, Panneershelvam, Lanctot, Dieleman, Grewe, Nham, Kalchbrenner, Sutskever, Lillicrap, Leach, Kavukcuoglu, Graepel, and Hassabis]{silver2016alphago}
David Silver, Aja Huang, Chris~J. Maddison, Arthur Guez, Laurent Sifre, George van~den Driessche, Julian Schrittwieser, Ioannis Antonoglou, Veda Panneershelvam, Marc Lanctot, Sander Dieleman, Dominik Grewe, John Nham, Nal Kalchbrenner, Ilya Sutskever, Timothy Lillicrap, Madeleine Leach, Koray Kavukcuoglu, Thore Graepel, and Demis Hassabis.
\newblock Mastering the game of go with deep neural networks and tree search.
\newblock \emph{Nature}, 529:\penalty0 484--489, 2016.
\newblock URL \url{https://doi.org/10.1038/nature16961}.

\bibitem[Tatham(2004{\natexlab{a}})]{site:sgt-puzzles}
Simon Tatham.
\newblock Simon tatham's portable puzzle collection, 2004{\natexlab{a}}.
\newblock URL \url{https://www.chiark.greenend.org.uk/~sgtatham/puzzles/}.
\newblock Accessed: 2023-05-16.

\bibitem[Foundation(2022)]{site:farama-gymnasium}
Farama Foundation.
\newblock Gymnasium website, 2022.
\newblock URL \url{https://gymnasium.farama.org/}.
\newblock Accessed: 2023-05-12.

\bibitem[Wang et~al.(2022)Wang, Chen, Li, and Wang]{wang2022rethinking}
Chao Wang, Chen Chen, Dong Li, and Bin Wang.
\newblock Rethinking reinforcement learning based logic synthesis.
\newblock \emph{arXiv preprint arXiv:2205.07614}, 2022.

\bibitem[Dasgupta et~al.(2019)Dasgupta, Wang, Chiappa, Mitrovic, Ortega, Raposo, Hughes, Battaglia, Botvinick, and Kurth-Nelson]{dasgupta2019causal}
Ishita Dasgupta, Jane Wang, Silvia Chiappa, Jovana Mitrovic, Pedro Ortega, David Raposo, Edward Hughes, Peter Battaglia, Matthew Botvinick, and Zeb Kurth-Nelson.
\newblock Causal reasoning from meta-reinforcement learning.
\newblock \emph{arXiv preprint arXiv:1901.08162}, 2019.

\bibitem[Eppe et~al.(2022)Eppe, Gumbsch, Kerzel, Nguyen, Butz, and Wermter]{eppe2022intelligent}
Manfred Eppe, Christian Gumbsch, Matthias Kerzel, Phuong~DH Nguyen, Martin~V Butz, and Stefan Wermter.
\newblock Intelligent problem-solving as integrated hierarchical reinforcement learning.
\newblock \emph{Nature Machine Intelligence}, 4\penalty0 (1):\penalty0 11--20, 2022.

\bibitem[Deac et~al.(2021)Deac, Veli{\v{c}}kovi{\'c}, Milinkovic, Bacon, Tang, and Nikolic]{deac2021neural}
Andreea-Ioana Deac, Petar Veli{\v{c}}kovi{\'c}, Ognjen Milinkovic, Pierre-Luc Bacon, Jian Tang, and Mladen Nikolic.
\newblock Neural algorithmic reasoners are implicit planners.
\newblock \emph{Advances in Neural Information Processing Systems}, 34:\penalty0 15529--15542, 2021.

\bibitem[He et~al.(2022)He, Veli{\v{c}}kovi{\'c}, Li{\`o}, and Deac]{he2022continuous}
Yu~He, Petar Veli{\v{c}}kovi{\'c}, Pietro Li{\`o}, and Andreea Deac.
\newblock Continuous neural algorithmic planners.
\newblock In \emph{Learning on Graphs Conference}, pages 54--1. PMLR, 2022.

\bibitem[Silver et~al.(2017)Silver, Hubert, Schrittwieser, Antonoglou, Lai, Guez, Lanctot, Sifre, Kumaran, Graepel, et~al.]{silver2017mastering}
David Silver, Thomas Hubert, Julian Schrittwieser, Ioannis Antonoglou, Matthew Lai, Arthur Guez, Marc Lanctot, Laurent Sifre, Dharshan Kumaran, Thore Graepel, et~al.
\newblock Mastering chess and shogi by self-play with a general reinforcement learning algorithm.
\newblock \emph{arXiv preprint arXiv:1712.01815}, 2017.

\bibitem[Dahl(2001)]{dahl2001reinforcement}
Fredrik~A Dahl.
\newblock A reinforcement learning algorithm applied to simplified two-player texas hold’em poker.
\newblock In \emph{European Conference on Machine Learning}, pages 85--96. Springer, 2001.

\bibitem[Heinrich and Silver(2016)]{heinrich2016deep}
Johannes Heinrich and David Silver.
\newblock Deep reinforcement learning from self-play in imperfect-information games.
\newblock \emph{arXiv preprint arXiv:1603.01121}, 2016.

\bibitem[Steinberger(2019)]{steinberger2019pokerrl}
Eric Steinberger.
\newblock Pokerrl.
\newblock \url{https://github.com/TinkeringCode/PokerRL}, 2019.

\bibitem[Zhao et~al.(2022)Zhao, Yan, Li, Li, and Xing]{zhao2022alphaholdem}
Enmin Zhao, Renye Yan, Jinqiu Li, Kai Li, and Junliang Xing.
\newblock Alphaholdem: High-performance artificial intelligence for heads-up no-limit poker via end-to-end reinforcement learning.
\newblock In \emph{Proceedings of the AAAI Conference on Artificial Intelligence}, volume~36, pages 4689--4697, 2022.

\bibitem[Ghory(2004)]{ghory2004reinforcement}
Imran Ghory.
\newblock Reinforcement learning in board games.
\newblock 2004.

\bibitem[Szita(2012)]{szita2012reinforcement}
Istv{\'a}n Szita.
\newblock Reinforcement learning in games.
\newblock In \emph{Reinforcement Learning: State-of-the-art}, pages 539--577. Springer, 2012.

\bibitem[Xenou et~al.(2019)Xenou, Chalkiadakis, and Afantenos]{xenou2019deep}
Konstantia Xenou, Georgios Chalkiadakis, and Stergos Afantenos.
\newblock Deep reinforcement learning in strategic board game environments.
\newblock In \emph{Multi-Agent Systems: 16th European Conference, EUMAS 2018, Bergen, Norway, December 6--7, 2018, Revised Selected Papers 16}, pages 233--248. Springer, 2019.

\bibitem[Perolat et~al.(2022)Perolat, De~Vylder, Hennes, Tarassov, Strub, de~Boer, Muller, Connor, Burch, Anthony, et~al.]{perolat2022mastering}
Julien Perolat, Bart De~Vylder, Daniel Hennes, Eugene Tarassov, Florian Strub, Vincent de~Boer, Paul Muller, Jerome~T Connor, Neil Burch, Thomas Anthony, et~al.
\newblock Mastering the game of stratego with model-free multiagent reinforcement learning.
\newblock \emph{Science}, 378\penalty0 (6623):\penalty0 990--996, 2022.

\bibitem[Cormen et~al.(2022)Cormen, Leiserson, Rivest, and Stein]{cormen2022clrs}
Thomas~H. Cormen, Charles~Eric Leiserson, Ronald~L. Rivest, and Clifford Stein.
\newblock \emph{Introduction to {A}lgorithms}.
\newblock The MIT Press, 4th edition, 2022.

\bibitem[Raffin et~al.(2021)Raffin, Hill, Gleave, Kanervisto, Ernestus, and Dormann]{raffin2021baselines}
Antonin Raffin, Ashley Hill, Adam Gleave, Anssi Kanervisto, Maximilian Ernestus, and Noah Dormann.
\newblock Stable-baselines3: Reliable reinforcement learning implementations.
\newblock \emph{Journal of Machine Learning Research}, 22\penalty0 (268):\penalty0 1--8, 2021.
\newblock URL \url{http://jmlr.org/papers/v22/20-1364.html}.

\bibitem[Werner~Duvaud(2019)]{muzero-general}
Aurèle~Hainaut Werner~Duvaud.
\newblock Muzero general: Open reimplementation of muzero.
\newblock \url{https://github.com/werner-duvaud/muzero-general}, 2019.

\bibitem[Hafner et~al.(2023{\natexlab{a}})Hafner, Pasukonis, Ba, and Lillicrap]{dreamerv3-code}
Danijar Hafner, Jurgis Pasukonis, Jimmy Ba, and Timothy Lillicrap.
\newblock Mastering diverse domains through world models.
\newblock \url{https://github.com/danijar/dreamerv3}, 2023{\natexlab{a}}.

\bibitem[Haarnoja et~al.(2018)Haarnoja, Zhou, Abbeel, and Levine]{haarnoja2018soft}
Tuomas Haarnoja, Aurick Zhou, Pieter Abbeel, and Sergey Levine.
\newblock Soft actor-critic: Off-policy maximum entropy deep reinforcement learning with a stochastic actor.
\newblock In \emph{International conference on machine learning}, pages 1861--1870. PMLR, 2018.

\bibitem[Fujimoto et~al.(2018)Fujimoto, Hoof, and Meger]{fujimoto2018addressing}
Scott Fujimoto, Herke Hoof, and David Meger.
\newblock Addressing function approximation error in actor-critic methods.
\newblock In \emph{International conference on machine learning}, pages 1587--1596. PMLR, 2018.

\bibitem[Silver et~al.(2021)Silver, Singh, Precup, and Sutton]{silver2021reward}
David Silver, Satinder Singh, Doina Precup, and Richard~S Sutton.
\newblock Reward is enough.
\newblock \emph{Artificial Intelligence}, 299:\penalty0 103535, 2021.

\bibitem[Vamplew et~al.(2022)Vamplew, Smith, K{\"a}llstr{\"o}m, Ramos, R{\u{a}}dulescu, Roijers, Hayes, Heintz, Mannion, Libin, et~al.]{vamplew2022scalar}
Peter Vamplew, Benjamin~J Smith, Johan K{\"a}llstr{\"o}m, Gabriel Ramos, Roxana R{\u{a}}dulescu, Diederik~M Roijers, Conor~F Hayes, Fredrik Heintz, Patrick Mannion, Pieter~JK Libin, et~al.
\newblock Scalar reward is not enough: A response to silver, singh, precup and sutton (2021).
\newblock \emph{Autonomous Agents and Multi-Agent Systems}, 36\penalty0 (2):\penalty0 41, 2022.

\bibitem[Community(2000)]{site:pygame}
Pygame Community.
\newblock Pygame github repository, 2000.
\newblock URL \url{https://github.com/pygame/pygame/}.
\newblock Accessed: 2023-05-12.

\bibitem[Tatham(2004{\natexlab{b}})]{site:sgt-dev-documentation}
Simon Tatham.
\newblock Developer documentation for simon tatham's puzzle collection, 2004{\natexlab{b}}.
\newblock URL \url{https://www.chiark.greenend.org.uk/~sgtatham/puzzles/devel/}.
\newblock Accessed: 2023-05-23.

\bibitem[Schulman et~al.(2017)Schulman, Wolski, Dhariwal, Radford, and Klimov]{schulman2017ppo}
John Schulman, Filip Wolski, Prafulla Dhariwal, Alec Radford, and Oleg Klimov.
\newblock Proximal policy optimization algorithms, 2017.
\newblock URL \url{http://arxiv.org/abs/1707.06347}.

\bibitem[Huang et~al.(2022)Huang, Dossa, Raffin, Kanervisto, and Wang]{shengyi2022the37implementation}
Shengyi Huang, Rousslan Fernand~Julien Dossa, Antonin Raffin, Anssi Kanervisto, and Weixun Wang.
\newblock The 37 implementation details of proximal policy optimization.
\newblock In \emph{ICLR Blog Track}, 2022.
\newblock URL \url{https://iclr-blog-track.github.io/2022/03/25/ppo-implementation-details/}.
\newblock https://iclr-blog-track.github.io/2022/03/25/ppo-implementation-details/.

\bibitem[Mnih et~al.(2016)Mnih, Badia, Mirza, Graves, Lillicrap, Harley, Silver, and Kavukcuoglu]{mnih2016a3c}
Volodymyr Mnih, Adri{\`{a}}~Puigdom{\`{e}}nech Badia, Mehdi Mirza, Alex Graves, Timothy~P. Lillicrap, Tim Harley, David Silver, and Koray Kavukcuoglu.
\newblock Asynchronous methods for deep reinforcement learning.
\newblock \emph{CoRR}, abs/1602.01783, 2016.
\newblock URL \url{http://arxiv.org/abs/1602.01783}.

\bibitem[Schulman et~al.(2015)Schulman, Levine, Abbeel, Jordan, and Moritz]{schulman2015trpo}
John Schulman, Sergey Levine, Pieter Abbeel, Michael Jordan, and Philipp Moritz.
\newblock Trust region policy optimization.
\newblock In Francis Bach and David Blei, editors, \emph{Proceedings of the 32nd International Conference on Machine Learning}, volume~37 of \emph{Proceedings of Machine Learning Research}, pages 1889--1897, Lille, France, 07--09 Jul 2015. PMLR.
\newblock URL \url{https://proceedings.mlr.press/v37/schulman15.html}.

\bibitem[Dabney et~al.(2017)Dabney, Rowland, Bellemare, and Munos]{dabney2017qrdqn}
Will Dabney, Mark Rowland, Marc~G. Bellemare, and R{\'{e}}mi Munos.
\newblock Distributional reinforcement learning with quantile regression.
\newblock \emph{CoRR}, abs/1710.10044, 2017.
\newblock URL \url{http://arxiv.org/abs/1710.10044}.

\bibitem[Schrittwieser et~al.(2020)Schrittwieser, Antonoglou, Hubert, Simonyan, Sifre, Schmitt, Guez, Lockhart, Hassabis, Graepel, et~al.]{schrittwieser2020mastering}
Julian Schrittwieser, Ioannis Antonoglou, Thomas Hubert, Karen Simonyan, Laurent Sifre, Simon Schmitt, Arthur Guez, Edward Lockhart, Demis Hassabis, Thore Graepel, et~al.
\newblock Mastering atari, go, chess and shogi by planning with a learned model.
\newblock \emph{Nature}, 588\penalty0 (7839):\penalty0 604--609, 2020.

\bibitem[Hafner et~al.(2023{\natexlab{b}})Hafner, Pasukonis, Ba, and Lillicrap]{hafner2023mastering}
Danijar Hafner, Jurgis Pasukonis, Jimmy Ba, and Timothy Lillicrap.
\newblock Mastering diverse domains through world models.
\newblock \emph{arXiv preprint arXiv:2301.04104}, 2023{\natexlab{b}}.

\end{thebibliography}
